\documentclass{article}

\usepackage{graphicx}
\usepackage{booktabs} 

\usepackage{hyperref}

\usepackage[accepted]{icml2024}

\usepackage{amsmath}
\usepackage{amssymb}
\usepackage{mathtools}
\usepackage{amsthm}
\usepackage[export]{adjustbox}

\usepackage{breakurl}
\usepackage{color}
\usepackage{caption}
\usepackage{multirow}
\usepackage{framed}
\usepackage[subrefformat=parens]{subcaption}

\usepackage[capitalize,noabbrev]{cleveref}

\theoremstyle{plain}

\theoremstyle{definition}

\theoremstyle{remark}

\usepackage[textsize=tiny]{todonotes}

\icmltitlerunning{Erasing Concepts from Text-to-Image Diffusion Models with Few-shot Unlearning}

\begin{document}

\twocolumn[
\icmltitle{Erasing Concepts from Text-to-Image Diffusion Models \\ with Few-shot Unlearning}

\icmlsetsymbol{equal}{*}

\begin{icmlauthorlist}
\icmlauthor{Masane Fuchi}{meiji}
\icmlauthor{Tomohiro Takagi}{meiji} 
\end{icmlauthorlist}
\icmlaffiliation{meiji}{Department of Computer Science, Meiji University, Japan}
\icmlcorrespondingauthor{Masane Fuchi}{ce235031@meiji.ac.jp}
\icmlkeywords{}

\vskip 0.1in

{
\begin{center}
    \centering
    \captionsetup{type=figure}
    \includegraphics[width=0.98\textwidth]{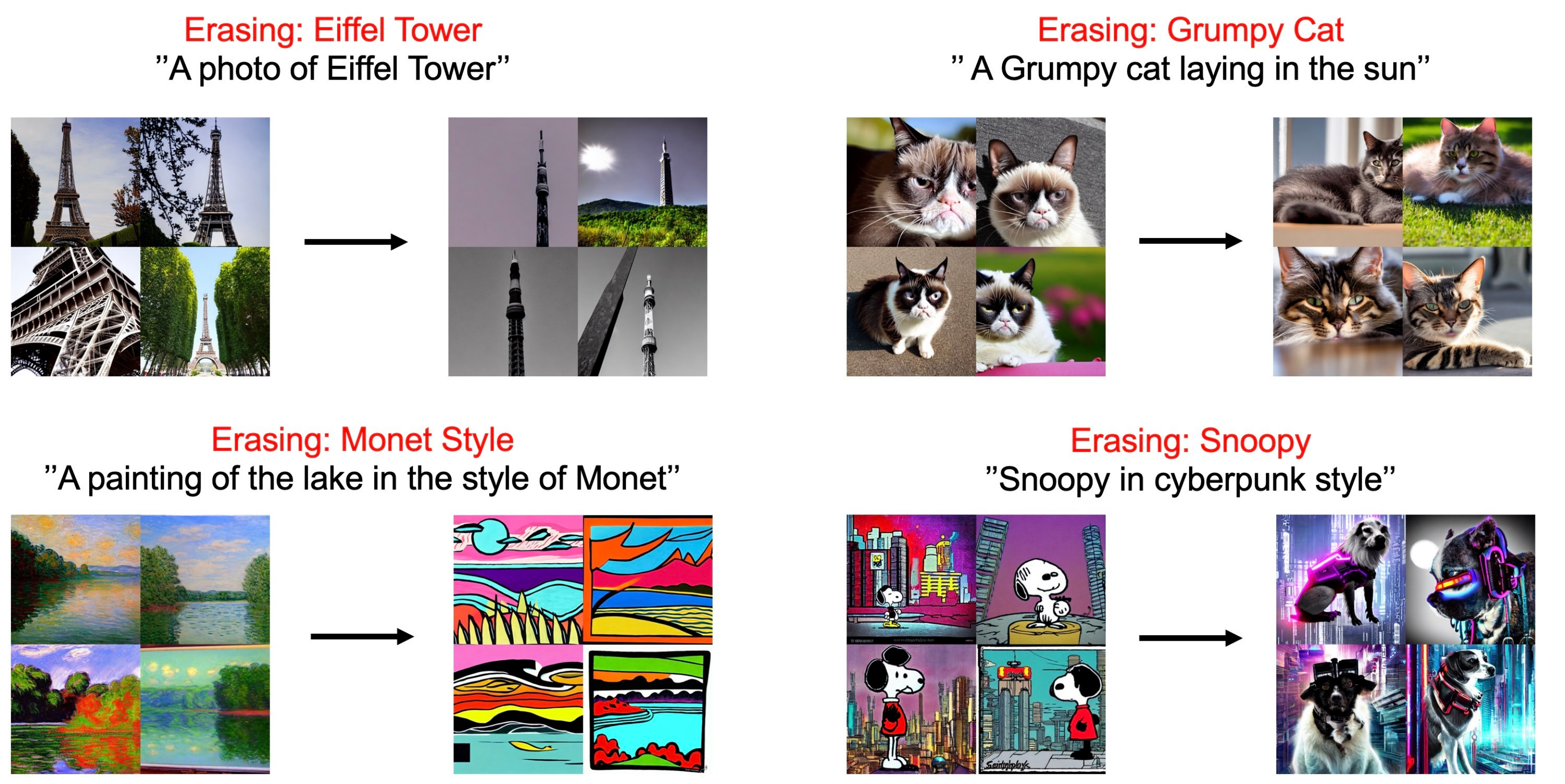}
    \captionof{figure}{Overview of our results. Given simple text, our method is able to erase concept within 10 s per concept with few images. Unlike current methods, our method involves updating the text encoder. Generated images after erasing with our method are mapped to similar concept without providing anchor concepts. For example, ``Snoopy'' is mapped to dog which is its motif and ``Grumpy Cat'' is mapped to cat which is the super-category.}
\end{center}
}
]

\printAffiliationsAndNotice{}  

\begin{abstract}
Generating images from text has become easier because of the scaling of diffusion models and advancements in the field of vision and language. These models are trained using vast amounts of data from the Internet. Hence, they often contain undesirable content such as copyrighted material. As it is challenging to remove such data and retrain the models, methods for erasing specific concepts from pre-trained models have been investigated. We propose a novel concept-erasure method that updates the text encoder using few-shot unlearning in which a few real images are used. The discussion regarding the generated images after erasing a concept has been lacking. While there are methods for specifying the transition destination for concepts, the validity of the specified concepts is unclear. Our method implicitly achieves this by transitioning to the latent concepts inherent in the model or the images. Our method can erase a concept within 10 s, making concept erasure more accessible than ever before. Implicitly transitioning to related concepts leads to more natural concept erasure. We applied the proposed method to various concepts and confirmed that concept erasure can be achieved tens to hundreds of times faster than with current methods. By varying the parameters to be updated, we obtained results suggesting that, like previous research, knowledge is primarily accumulated in the feed-forward networks of the text encoder. Our code is available at \url{https://github.com/fmp453/few-shot-erasing}
\end{abstract}

\section{Introduction}
Diffusion models~\cite{pmlr-v37-sohl-dickstein15, NEURIPS2020_4c5bcfec} have surpassed the previous state-of-the-art generative adversarial networks (GANs)~\cite{NIPS2014_5ca3e9b1, brock2018large, Karras_2020_CVPR} in performance, because of their stable learning and broad representation capabilities~\cite{dhariwal2021diffusion}. With classifier-free guidance~\cite{ho2021classifierfree}, it has also become possible to generate high-quality images on the basis of natural language instructions~\cite{saharia2022photorealistic, ramesh2022hierarchical, balaji2023ediffi, xue2023raphael, podell2024sdxl, dai2023emu}. 

In large-scale image-generative models, since various data are collected to improve generation quality, it is possible to generate undesirable images such as copyright contents. While filtering training data can help mitigate the generation of such undesirable images~\cite{openai2023dalle-3}, it is generally costly and challenging. This approach is also ineffective against pre-trained models. Previous studies have successfully removed specific concepts from text-to-image generative models by updating the weights of U-Net~\cite{10.1007/978-3-319-24574-4_28}, an image generative module, or its conditioned cross attention~\cite{Gandikota_2023_ICCV, Kumari_2023_ICCV, Gandikota_2024_WACV, Zhang_2024_CVPR, zhao2024separable}. However, updating parameters of U-Net can lead to a decrease in generation quality in unconditional cases.

We propose a method for erasing specific concepts from the text-to-image diffusion models without altering the parameters of the U-Net. Specifically, we focus on the text encoder and aim to achieve this by altering the quality of text conditioning. We use several images of the target concept to make slight changes to the parameters of the text encoder to remove the concept. Our method is inspired by textual inversion~\cite{gal2023an}, but since we only make minor parameter adjustments, it operates very quickly. \cref{tab:comparison-existing-methods} compares our proposed method with current methods. It is evident that our proposed method can erase concepts more quickly compared to existing methods. Additionally, our proposed method naturally maps to surrounding concepts, erasing the need for concept induction by an anchor concept\footnote{We describe anchor concepts in detail in \cref{appendix:anchor-concepts}.}.

\begin{table*}[t]
\caption{Comparison of proposed method with current methods when erasing ``Van Gogh style''. $\times$ means that anchor concept is necessary.}
\label{tab:comparison-existing-methods}
\centering
\begin{small}
\begin{tabular}{lcccc}
\toprule
Method & Target parameters & Runtime & Anchor Concepts & \#U-Net \\
\midrule
ESD-x~\cite{Gandikota_2023_ICCV} & Cross-Attention & 1 hour & $\surd$ & 2 \\
UCE~\cite{Gandikota_2024_WACV} & Attention Weight & 10 min & $\times$ & 1 \\
SPM~\cite{Lyu_2024_CVPR} & Adapters in U-Net & 2.5 hours & $\surd$ & 1 \\
Ours & Text Encoder & \textbf{7$\sim$8 sec} & $\surd$ & 1 \\
\bottomrule
\end{tabular}
\end{small}
\end{table*}

Our contributions are as follows:
\begin{itemize}
\item We achieve a speedup of 60$\sim$900 times compared with current methods of updating the traditional U-Net, enabling concept erasure within 10 s.
\item Concept erasure is achieved by providing several images related to the concept to be erased. 
\item While current methods often lack discussion on the generated examples after concept erasure, our proposed method ensures semantic similarity in the resulting concepts. We highlighted this issue in the quantitative evaluation of previous research.
\end{itemize}

\subsection{Intuitive Motivation}
We provide an explanation of the intuitive motivation behind our study. On the basis of this motivation, we introduce our proposed method supported by various justifications.

Text-to-image diffusion models are implemented using models trained on a large amount of data from the web. They are highly versatile foundation models capable of executing various downstream tasks~\cite{bommasani2022opportunities}. Therefore, adapting them to individual domains requires additional fine-tuning. The smallest unit of the domain adaptation is personalization, and numerous methods have been proposed to achieve this~\cite{gal2023an, Ruiz_2023_CVPR, Wei_2023_ICCV, chen2023subjectdriven, Pang_2024_CVPR, 10.1145/3581783.3612599, zhang2024survey}. Inspired by these studies, we have one fundamental question.

\begin{framed}
\it If it is possible for these methods to give specific knowledge, could a similar technique be used to make them forget?
\end{framed}

We address this question by taking textual inversion~\cite{gal2023an} as a reference and partially changing the method. In the subsequent sections, we explore our proposed method and the answer to this question focusing on the modifications made.

\section{Our Method}
Our goal is to prevent the generation of specific concepts by updating the text encoder. The parameters of the U-Net responsible for image generation remain unchanged, ensuring that the model's generation capability (image fidelity) is preserved. We first discuss why we believe the text encoder should be updated (\cref{subsec:why-text-encoder}). We then outline our proposed method (\cref{subsec:updating-method}).

\subsection{Why Text Encoder?}
\label{subsec:why-text-encoder}
In this subsection, we explain why our method targets the updating of the text encoder.

In the field of text-to-image, it has been demonstrated that the quality of the text encoder, represented by its size, correlates with the quality of text-image alignment~\cite{saharia2022photorealistic}. When comparing the use of CLIP~\cite{pmlr-v139-radford21a} and T5~\cite{10.5555/3455716.3455856} as the text encoder, it has been indicated that the T5, which is trained solely on text data outperforms human evaluation compared with using CLIP. While quantitative evaluations show similar results between CLIP and T5, it is known that quantitative metrics (such as Fréchet inception distance (FID)~\cite{NIPS2017_8a1d6947} and CLIP Score~\cite{hessel-etal-2021-clipscore}) do not always align with human evaluations in assessing the quality of generated images~\cite{Otani_2023_CVPR}.  Therefore, achieving superior results in human evaluations support the hypothesis that the performance to some extent depends on the quality of the text encoder. DALLE-3~\cite{openai2023dalle-3}, which is trained by using high-quality and detailed image captions generated by GPT-4~\cite{openai2023gpt4}, achieves extremely high-quality image generation. From these findings, for models possessing sufficient image-generation capabilities (image fidelity, that is, lower FID), we believe that given information by the text affects the text-image alignment. This is also demonstrated in TextCrafter~\cite{Li_2024_CVPR}. CLIP's final output is a multi-dimensional vector. Although somewhat apparent from the concept of latent variables, visualizing this vector using techniques such as t-SNE~\cite{JMLR:v9:vandermaaten08a} allows for meaningful clustering~\cite{weiss2022neural}. Therefore, we believe that slight variations in the output of CLIP can be used to move towards similar concepts, and achieving this could be possible by slightly adjusting the parameters.

In many cases, including community models, the text encoder is often fixed during training. Considering the limited variety of text encoders, erasing specific concepts from one text encoder might be more beneficial than the strategy used with U-Net~\cite{Lyu_2024_CVPR} when transferring it for use in other models.

On the basis of the above, we conclude that updating the text encoder is more appropriate than updating the U-Net. Reframing concept erasure as ``significantly decreasing text-image alignment for specific concepts'' further highlights the natural focus on the text encoder.

\subsection{Erasing Method}
\label{subsec:updating-method}

\begin{figure*}[!htbp]
\centering
\includegraphics[width=\linewidth]{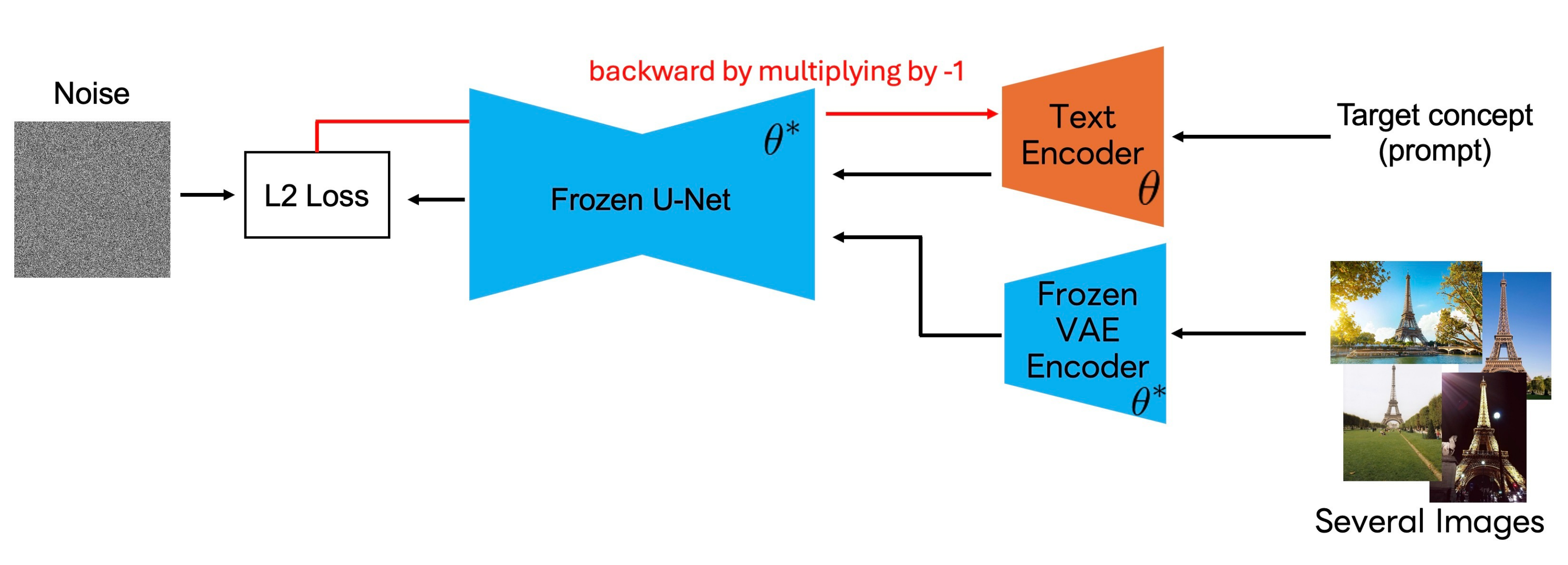}
\caption{Overview of our proposed method. $\theta^*$ denotes that parameters are fixed.}
\label{fig:proposed-method}
\end{figure*}

In this subsection, we consider preventing the generation of specific concepts by updating the text encoder. It is crucial to preserve as many concepts as possible other than the target concept. Therefore, as mentioned in \cref{subsec:why-text-encoder}, we consider making slight variations to the parameters of the text encoder.

\begin{equation}
c_{\theta}\leftarrow c_{\theta}+\Delta c,
\end{equation}

where $c_{\theta}$ denotes the text encoder. The issue lies in how to compute $\Delta c$. Our method is very simple, merely setting the loss in the reverse direction according to Jang et al.~\yrcite{jang-etal-2023-knowledge}. \cref{fig:proposed-method} illustrates an overview of our proposed method. While experiments in a similar setup have been conducted by Kumari et al.~\yrcite{Kumari_2023_ICCV}, they updated the parameters of the U-Net, which differs from our approach. This loss setting is very similar to textual inversion~\cite{gal2023an}. We use stable diffusion~\cite{Rombach_2022_CVPR} due to the restriction of our computational resources. The stable diffusion loss is given by

\begin{equation}
L_{SD}=\mathbb{E}_{x, \epsilon\in\mathcal{N}(0, 1), t, y}\left[\|\varepsilon-\varepsilon_{\theta}(x, t, c_{\theta}(y))\|_2^2\right]
\end{equation}

where $t$ is the timestep, $x$ is the denoised image to time $t$, $\varepsilon$ is the unscaled noise sample, and $\varepsilon_{\theta}$ is the denoising network. Our method involves training to maximize this loss. However, to prevent drastic changes to the text encoder, we terminate training within an appropriate range (managed with the number of epochs) rather than maximizing it completely.

\begin{equation}
-\mathbb{E}_{x, \epsilon\in\mathcal{N}(0, 1), t, y}\left[\|\varepsilon-\varepsilon_{\theta}(x, t, c_{\theta}(y^*))\|_2^2\right]
\end{equation}

We update only the text encoder, so the parameters of $\varepsilon_{\theta}$ are fixed. The notation $y^*$ denotes the caption that includes the concept to erase. We use CLIP ImageNet Template~\cite{pmlr-v139-radford21a} for this caption as well as textual inversion. By making slight variations to the text encoder, we assume that the impact on other concepts is minimal and is further mapped to approximate concepts. Since only minor changes are made, training is completed in a short period.

In this formulation, we investigated two approaches: one using a few-shot method~\cite{10.1145/3386252} with four pre-prepared images for $x$ and the other using completely random noise (zero-shot). 

Intuitively, considering text-image alignment, the few-shot approach is expected to perform better. We aimed to confirm this through experiment. Considering practical aspects, it is straightforward to prepare actual images of the concepts to be erased, making it possible to confirm the effectiveness of the method even with the few-shot approach.

\subsection{Update Parameters}
There are studies suggesting that the knowledge gained during learning is stored in feed-forward networks (i.e. MLPs)~\cite{NEURIPS2022_6f1d43d5, dai-etal-2022-knowledge, geva-etal-2021-transformer}. There is also a study that shows that the final self-attention layer is also effective in knowledge editing~\cite{NEURIPS2022_6f1d43d5}. Following these studies, we update all MLPs and the final self-attention layer. The list of updated parameters is shown in \cref{table:updated-parameters} of \cref{appendix:updated-parameters}.

\section{Experiments}
\label{sec:exp}
We confirm the effectiveness of our proposed method throughout the experiments. Before presenting the results, we discuss the baselines (\cref{subsec:baselines}) and experimental settings (\cref{subsec:exp-setting}).

\subsection{Baselines}
\label{subsec:baselines}
We used ESD~\cite{Gandikota_2023_ICCV} (we specifically used ESD-x-1, which updates the parameters related to cross-attention), Unified Concept Editing (UCE)~\cite{Gandikota_2024_WACV}, and Semi-Permeable Membrane (SPM)~\cite{Lyu_2024_CVPR} which are open sourced and high effectiveness as the baselines. Due to the privacy issue with of LAION-5B~\cite{schuhmann2022laionb}\footnote{\url{https://laion.ai/notes/laion-maintanence/}}, we cannot conduct with Ablating Concept~\cite{Kumari_2023_ICCV}. When elements other than the pre-trained model depend on external materials, such as in this case, we may not be able to reproduce the results completely. The above baselines exclusively use the text-to-image diffusion models, including VAE, U-Net, text encoder, and tokenizer. For further details, please refer to \cref{appendix-subsec:baselines}.

\subsection{Experimental Setup}
\label{subsec:exp-setting}
\paragraph{Training.} We apply our proposed method to Stable Diffusion 1.5\footnote{\url{https://huggingface.co/runwayml/stable-diffusion-v1-5}} (as referred to original SD), the text encoder of which is OpenAI CLIP vit-large-patch14\footnote{\url{https://huggingface.co/openai/clip-vit-large-patch14}}. We use four images in the few-shot setting. The text encoder is optimized using Adam~\cite{kingma2017adam}. The hyperparameters we used in our experiments are listed in \cref{tab:hyperparameters}.

\begin{table}[htbp]
\caption{Hyperparameters of our method}
\label{tab:hyperparameters}
\centering
\begin{small}
\begin{tabular}{lc}
\toprule
Hyperparameter & Value \\
\midrule
Batch Size & 2 \\
\multirow{2}{*}{Training Epochs} & 4 (proper noun) \\
& 5 (common noun) \\
Learning rate & $1\times10^{-5}$\\
Adam $(\beta_1, \beta_2)$ & (0.9, 0.98) \\
Weight decay & $1\times10^{-8}$ \\
\bottomrule
\end{tabular}
\end{small}
\end{table}

\paragraph{Generating.} We used PNDM Scheduler~\cite{liu2022pseudo} with 7.5 guidance scale and 100 inference steps.

\subsection{Qualitative Results}

\begin{figure*}[!htbp]
\centering
\includegraphics[width=\linewidth]{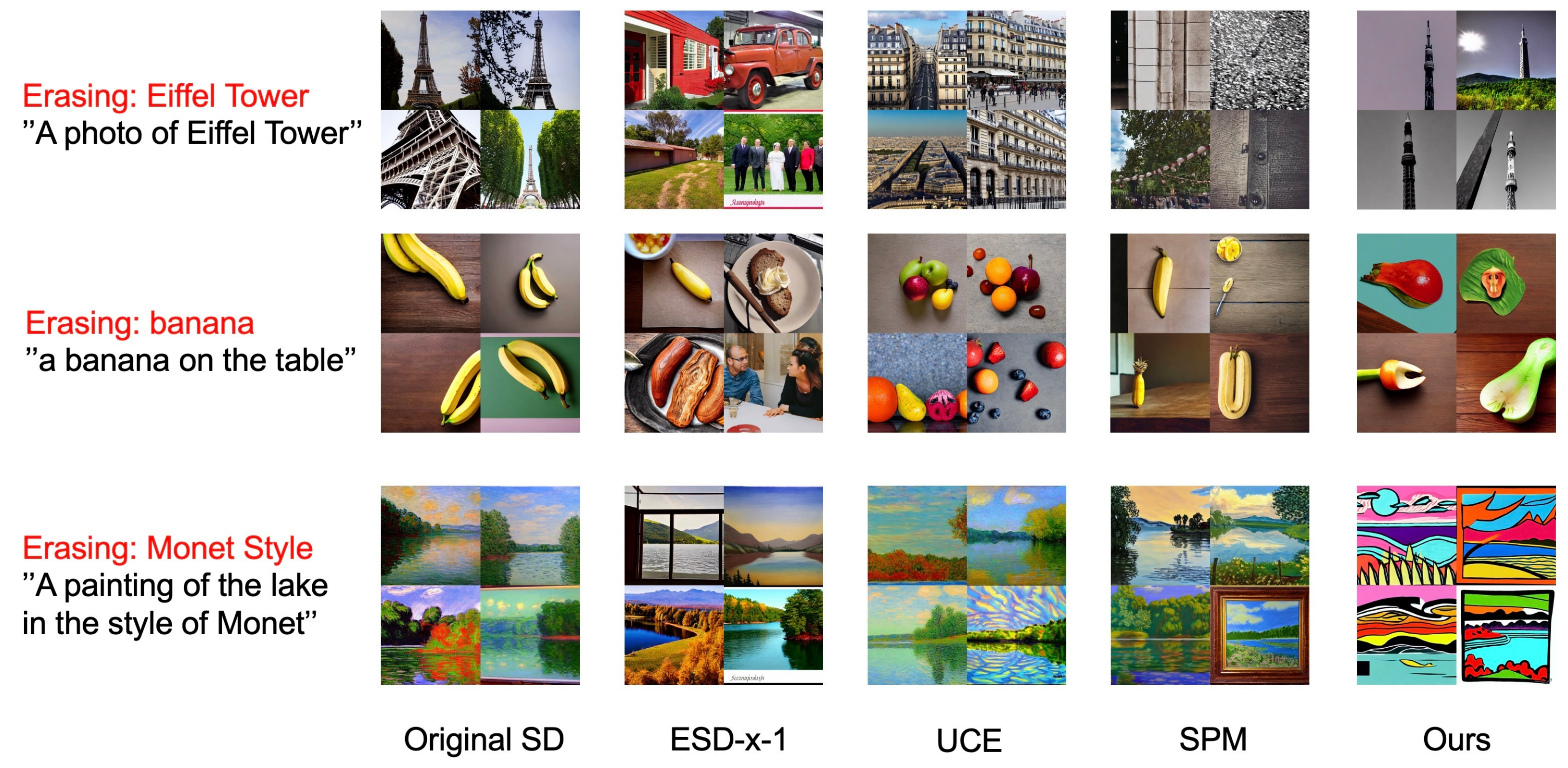}
\caption{Comparison of images generated with each method}
\label{fig:comparison}
\end{figure*}

\paragraph{Erasing Single Concept}
We conducted experiments focusing on proper nouns. We also conducted experiments with common nouns to ensure general performance. We focus on the results using ``Eiffel Tower'' as the proper noun and ``banana'' as the common noun. More results are presented in \cref{supl:additional-results}.

In the first row of \cref{fig:comparison}, the results of erasing ``Eiffel Tower'' are shown. With ESD and SPM, while the concept of the Eiffel Tower disappears, the generated images are completely unrelated. This phenomenon was also observed by Lu et al.~\yrcite{Lu_2024_CVPR}, and it is necessary to consider whether the generated images of the erased concepts are appropriate. With UCE, as the anchor concept was ``Paris'', the results depict the cityscape of Paris. In contrast, our proposed method not only prevented the generation of the Eiffel Tower but also retained only the elements of the tower, indicating that the embedding is mapped to a similar concept.

In the second row of \cref{fig:comparison}, the results of erasing ``banana'' are shown. Overall, similar results when erasing ``Eiffel Tower'' can be observed. SPM produced results that are difficult to interpret as successfully erasing the banana. As common nouns are ubiquitous words, they are likely heavily represented in the training data of original stable diffusion, making it challenging to erase such concepts. With UCE, as the anchor concept is ``Fruit'', various types of fruits were generated. Despite not explicitly specifying such categories, our proposed method generated something resembling a category of fruits to which bananas belong.

In the third row of \cref{fig:comparison}, the results of erasing ``Monet style'' are shown. ESD generated photorealistic images, while these images are not satisfied with the given prompt, ``A painting''. Other methods reflect the given prompt sufficiently. UCE is similar to original SD because the anchor concept is ``impressionism'' that includes ``Monet style''. It is suspicious the ``Monet style'' is erased. Our proposed method generated ``A painting'' while it is not ``Monet style''.

\begin{figure}[!htbp]
\centering
\begin{tabular}{ccc}
Cat & Tokyo Tower & Arc de Triomphe \\
\midrule \\
\begin{minipage}[b]{0.3\linewidth}
\centering
\includegraphics[width=0.9\linewidth]{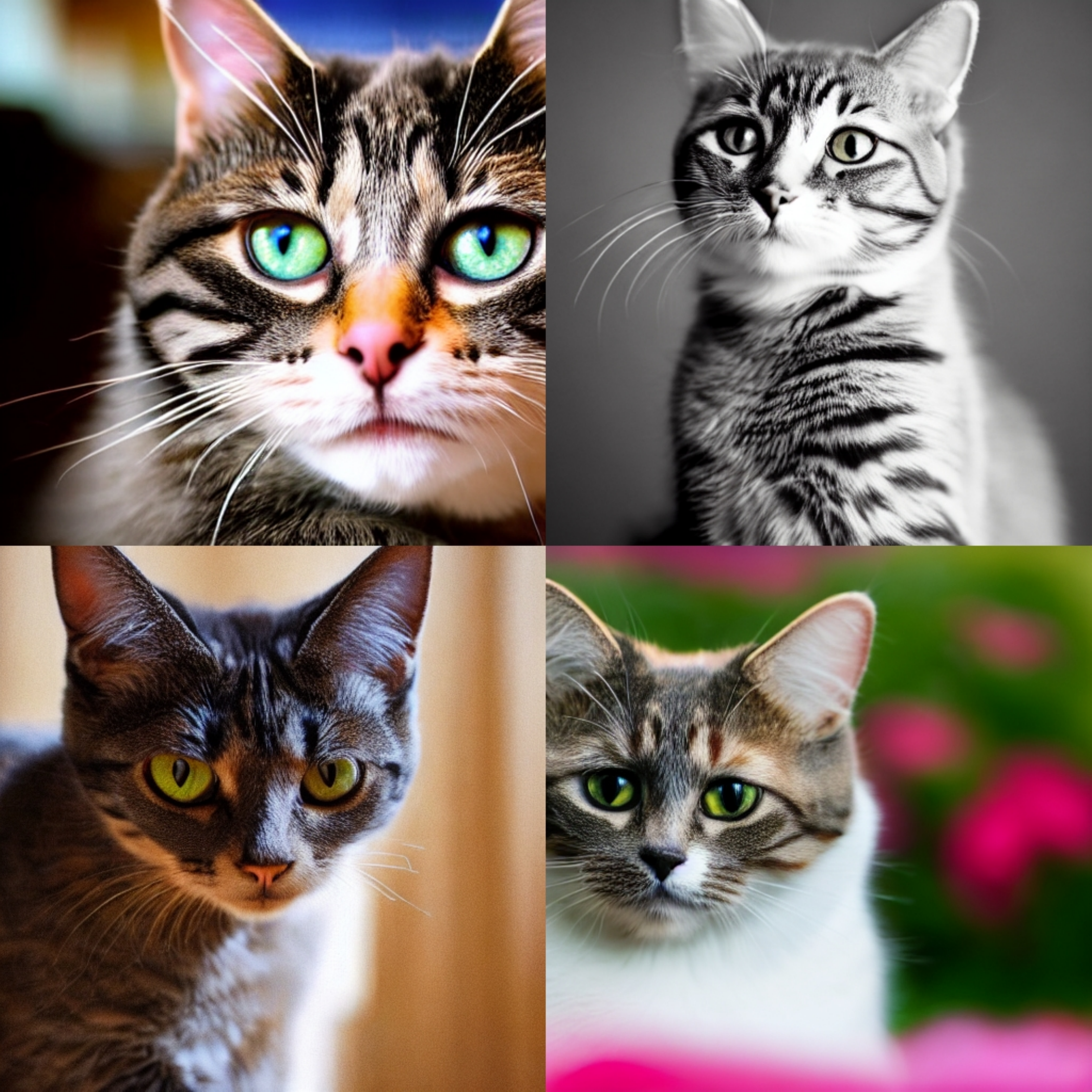}
\end{minipage} &
\begin{minipage}[b]{0.3\linewidth}
\centering
\includegraphics[width=0.9\linewidth]{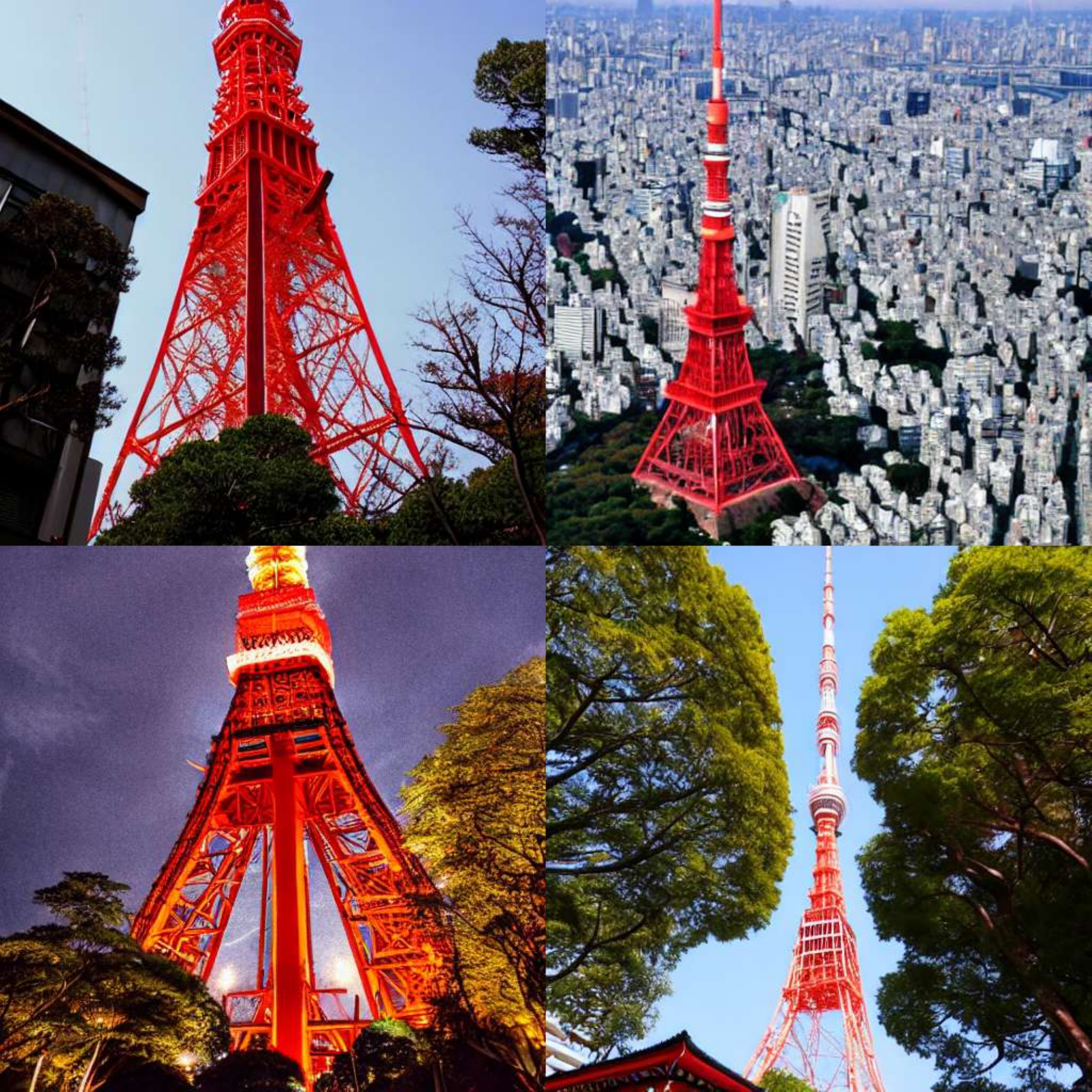}
\end{minipage} &
\begin{minipage}[b]{0.3\linewidth}
\centering
\includegraphics[width=0.9\linewidth]{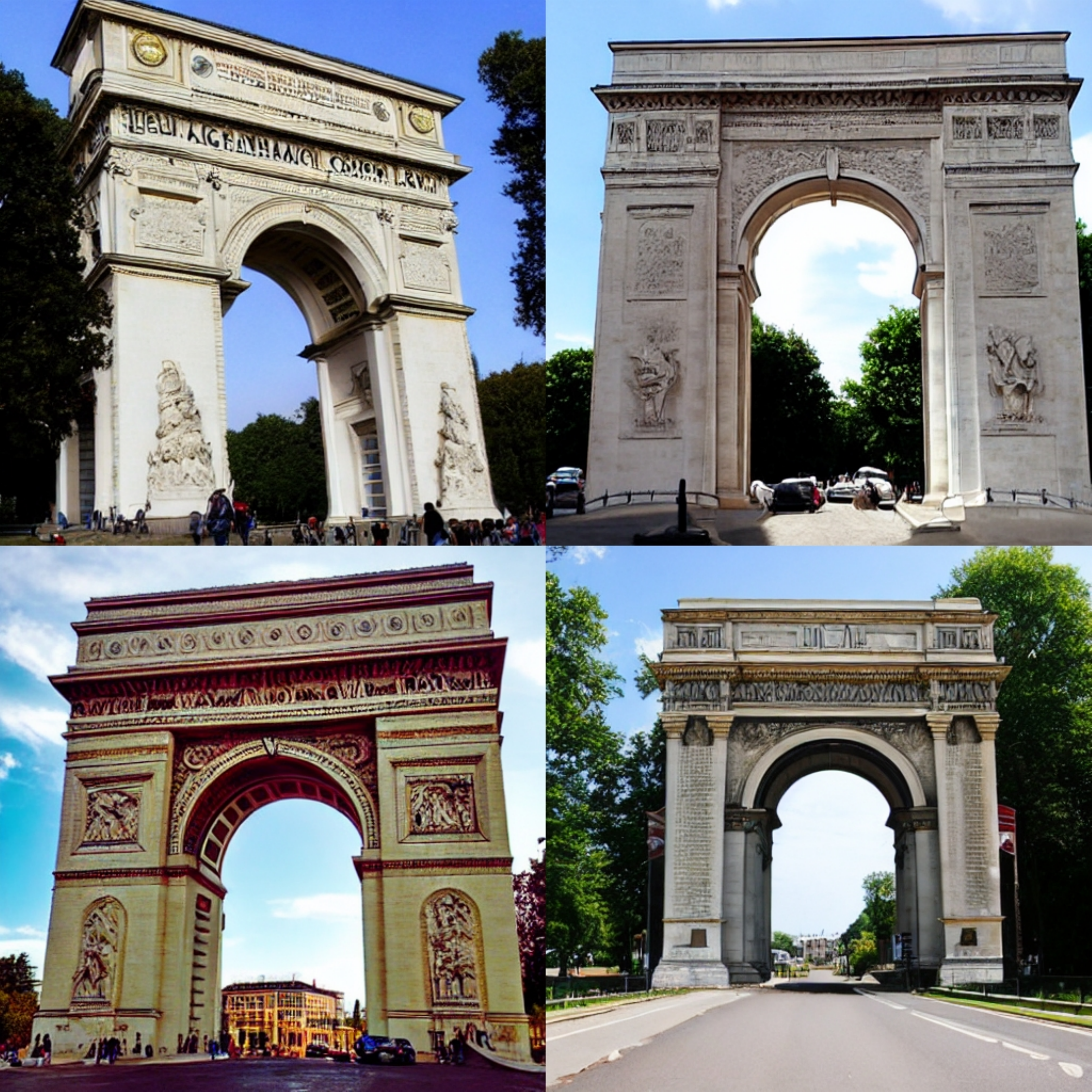}
\end{minipage} \\
\begin{minipage}[b]{0.3\linewidth}
\centering
\includegraphics[width=0.9\linewidth]{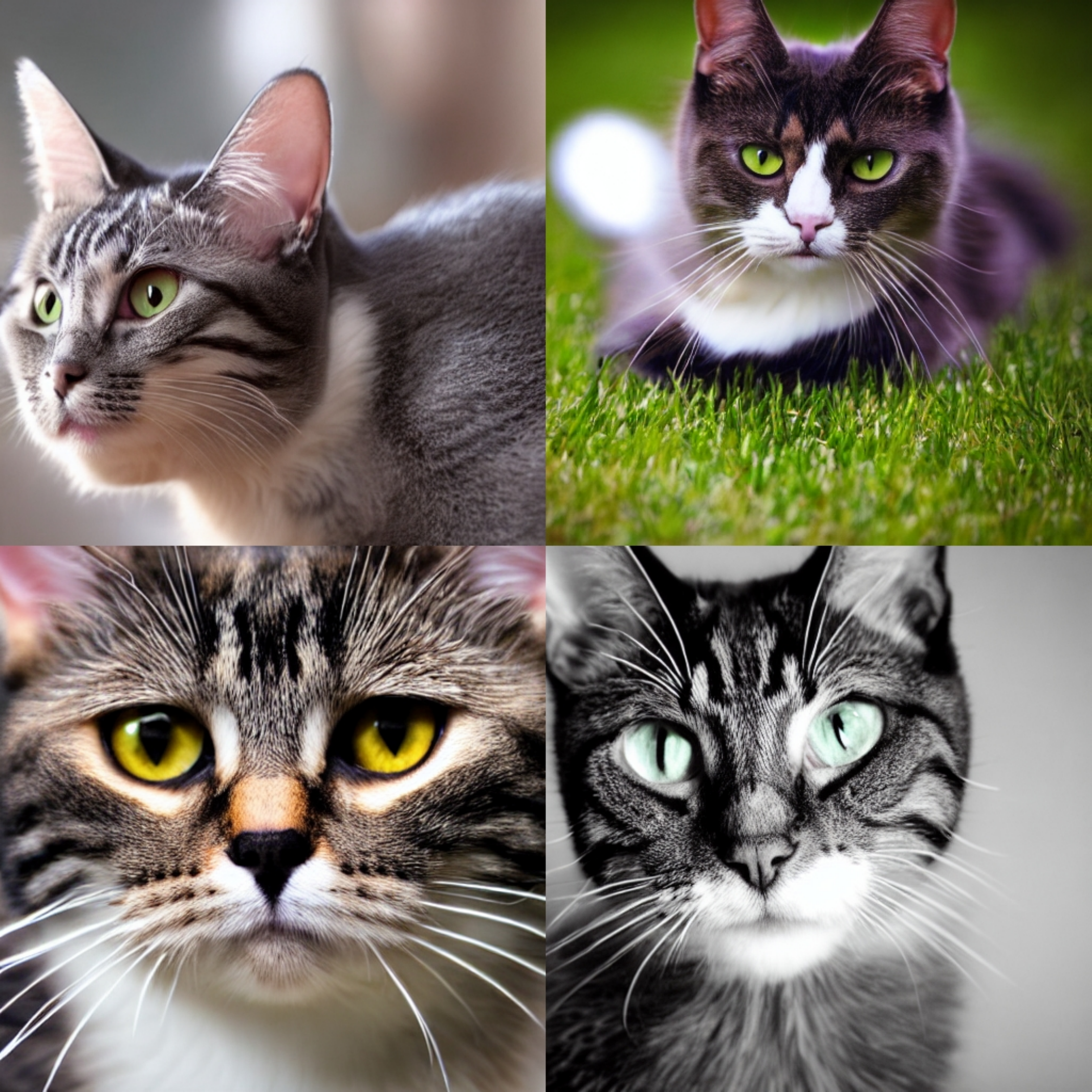}
\end{minipage} &
\begin{minipage}[b]{0.3\linewidth}
\centering
\includegraphics[width=0.9\linewidth]{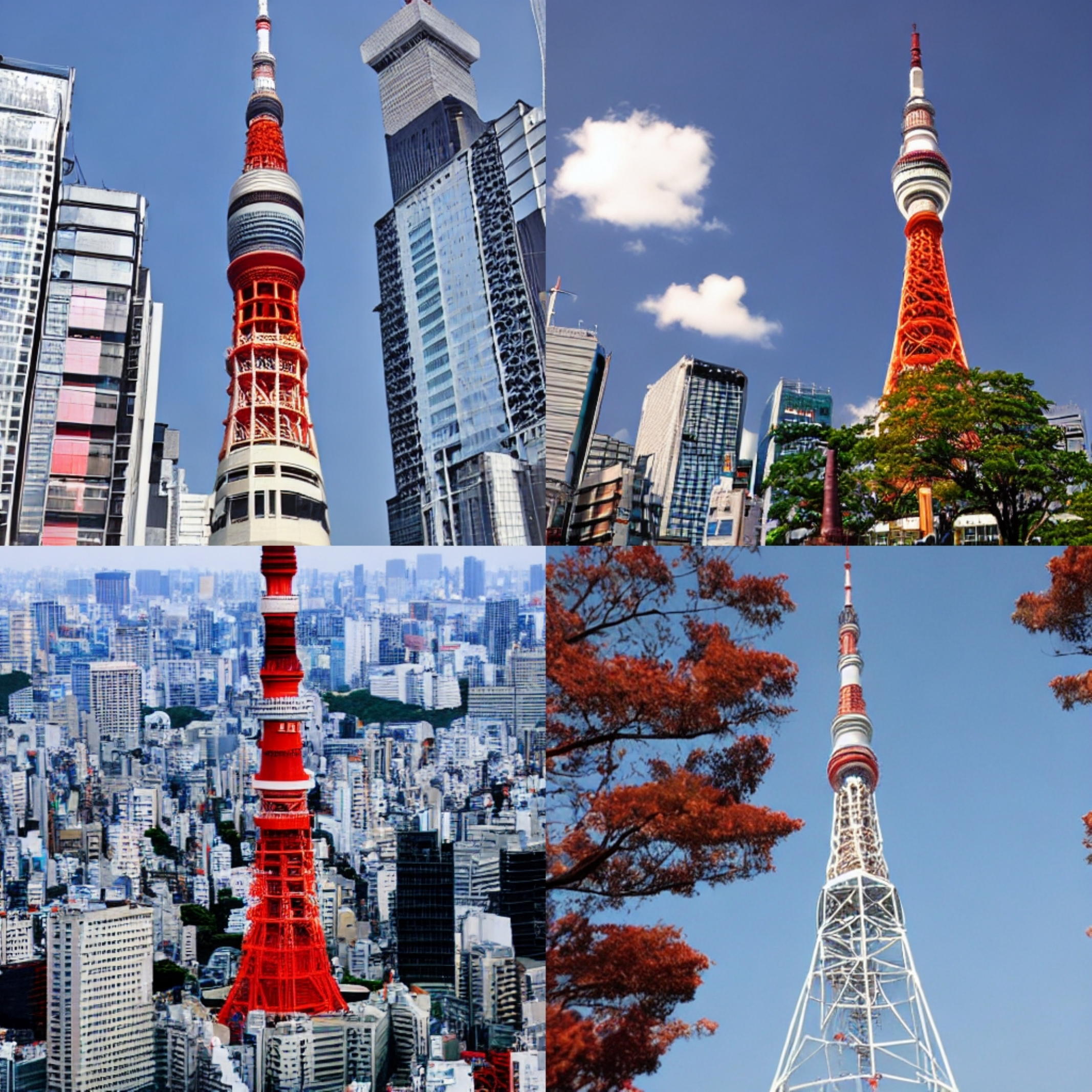}
\end{minipage} &
\begin{minipage}[b]{0.3\linewidth}
\centering
\includegraphics[width=0.9\linewidth]{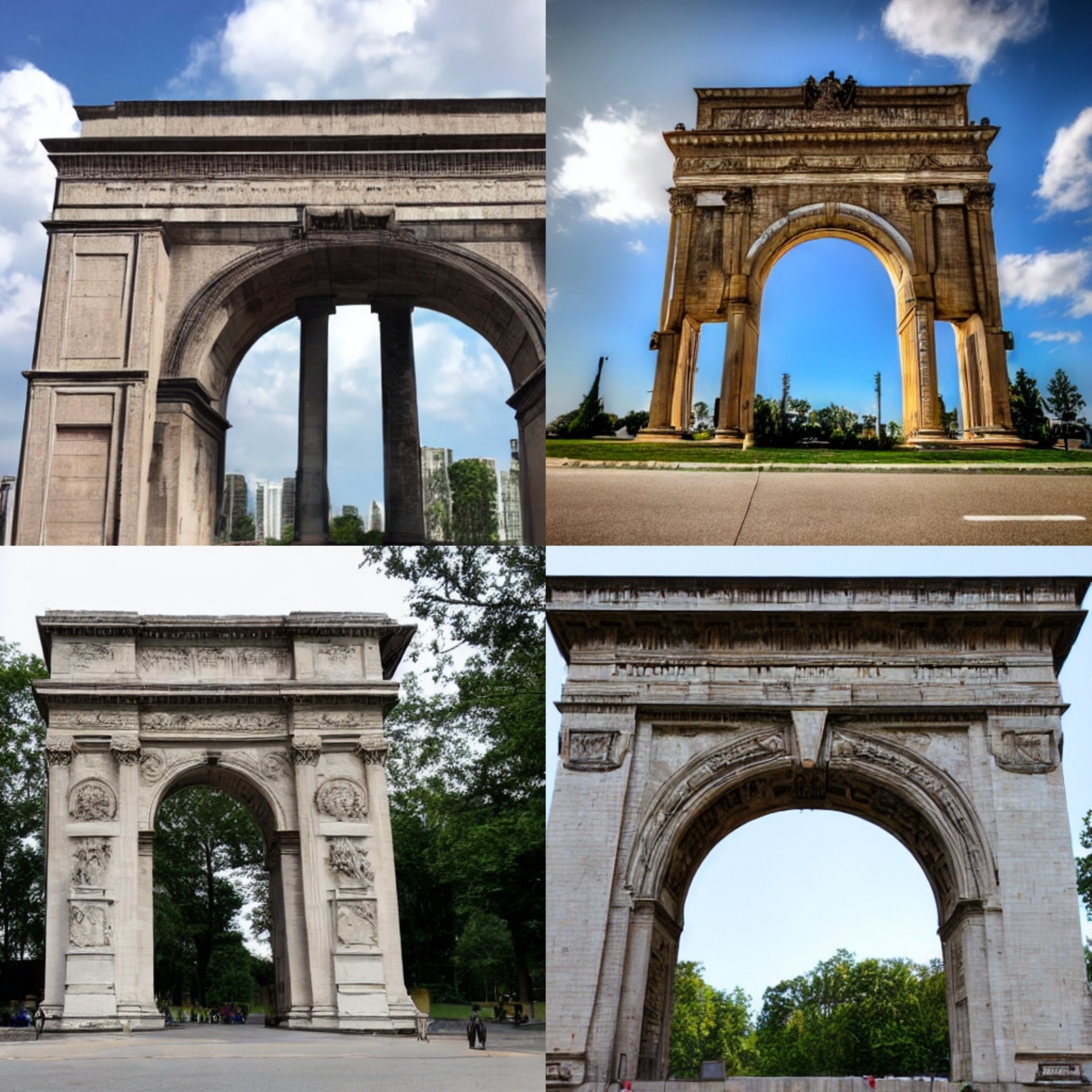}
\end{minipage} \\
\bottomrule
\end{tabular}
\caption{Comparison of the generated images between before (upper) and after (bottom)  erasing ``Eiffel Tower''. Caption we used when generating was ``a photo of (a) \{\textit{concept}\}''.}
\label{fig:erasing-eiffel-other-concepts}
\end{figure}

\paragraph{Effect on Other Concepts} We investigated the impact of removing one concept on the generation of other concepts. We used a text encoder with ``Eiffel Tower'' erased to generate other concepts. The generated concepts were ``cat'', which is an unrelated concept, another landmark in Paris ``Arc de Triomphe'' (the caption we used ``Triumphal arch''), and a similar-shaped landmark “Tokyo Tower”. The results are shown in \cref{fig:erasing-eiffel-other-concepts}. It can be seen that other concepts could also be generated. The generated images of the Tokyo Tower differ in actual shape\footnote{The actual shape of Tokyo Tower can be checked at \url{https://www.japan.travel/en/spot/1709/}}. However, because this is already observed in the images from the Original SD, we did not consider it is a significant drawback in the qualitative evaluations.

In previous studies~\cite{Gandikota_2023_ICCV, Kumari_2023_ICCV, Gandikota_2024_WACV, Zhang_2024_CVPR, zhao2024separable}, the generated images from Original SD were treated as ground truth. However, we consider this approach lacks justification. As observed in the example of ``Tokyo Tower'' in \cref{fig:erasing-eiffel-other-concepts}, the generated images from the Original SD are not perfect. Like Fan et al.~\yrcite{fan2024salun}, we believe that the generated images using a retrain model\footnote{Details of a retrain model are described in \cref{appendix:retrain}.} should be the ground truth. However, a retrain model involves significant computational costs, surpassing the scope of the type of experimentation we can conduct. As the training data of CLIP are not publicly available, reproducing experiments is also impossible. Therefore, we primarily conducted a qualitative evaluation.

\begin{figure}[!htbp]
\centering
\includegraphics[width=\linewidth]{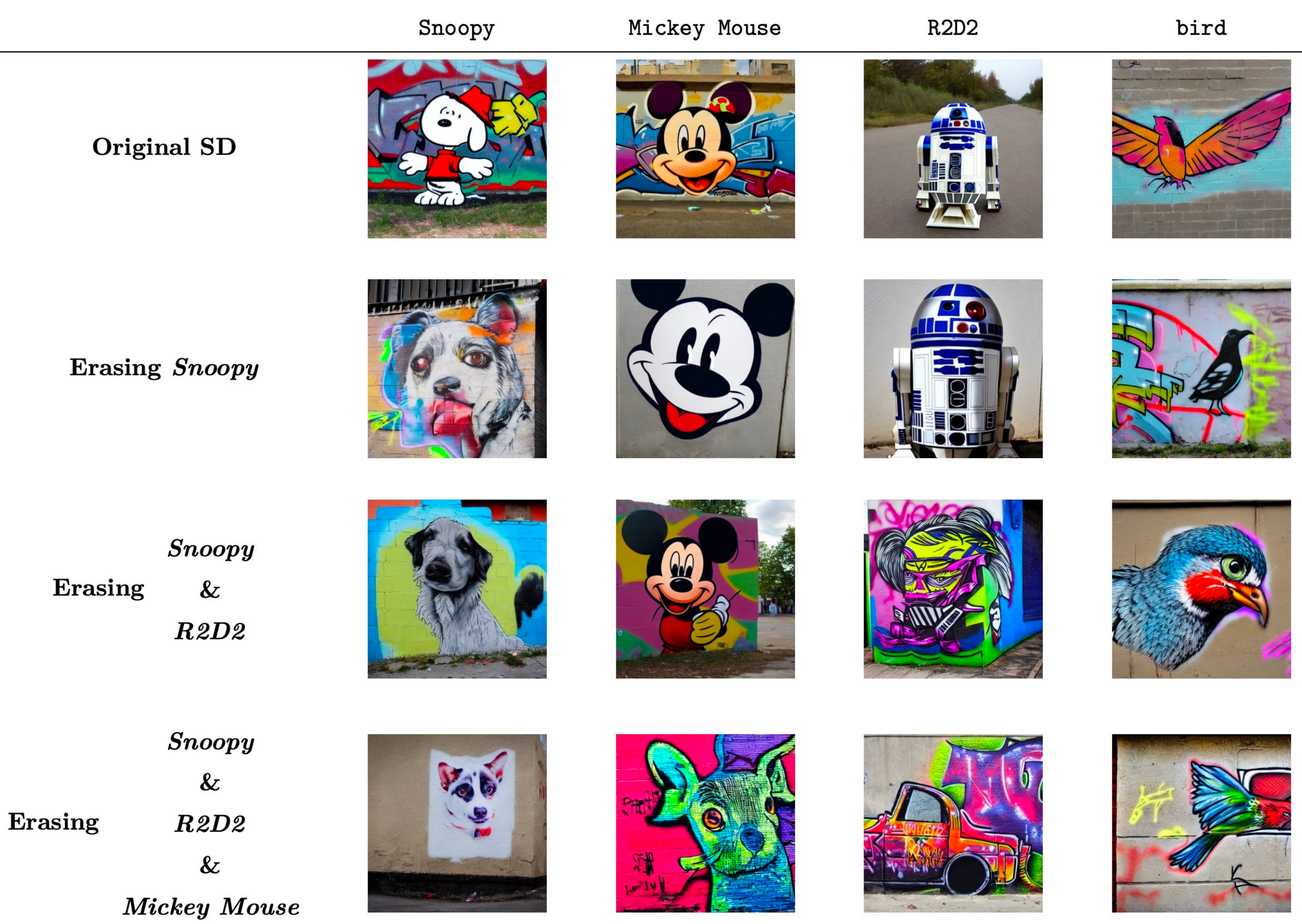}
\caption{Samples of ``graffiti of the \{\textit{Concept}\}'' when erasing multiple concepts}
\label{fig:multi}
\end{figure}

\paragraph{Erasing Multiple Concepts}
We considered the erasure of multiple concepts. We erase ``Snoopy'', ``R2D2'', and ``Mickey Mouse'' in that order. The results are shown in \cref{fig:multi}. Even after removing multiple concepts, the erased concepts did not reappear. While the change in generated images could be considered a future challenge, as previously mentioned, a retrain model served as the ground truth in our research, mitigating this issue. It can be observed that for the concepts that were not erased, the contents of the prompts were reflected in the generated images. The generated images of R2D2 from Original SD also did not match the prompt, which serves as another example of why Original SD is not considered the ground truth.

\subsection{Quantitative Results}
We used CLIP Score~\cite{hessel-etal-2021-clipscore} as the evaluation metric. While FID is suitable for evaluation on limited datasets, there are concerns about its inability to reflect diversity in other cases~\cite{Jayasumana_2024_CVPR}, so we choose not to use this standard metric. CLIP Score faces similar challenges in evaluating concepts not learned by CLIP~\cite{Otani_2023_CVPR}. We address this issue by not using such captions during evaluation. Since our method does not update any parameters of the U-Net, image fidelity should not degrade. However, as the parameters of the text encoder are updated, text-image alignment may decrease.

It should be noted that while a higher CLIP Score is desirable for concepts that have not been erased, the same may not hold true for erased concepts. Considering the example of the ``Eiffel Tower'' in \cref{fig:comparison}, for instance, it is anticipated that ESD-x-1 would have a lower CLIP Score compared with our proposed method. This is because when there is no relevance between the text and generated image, the CLIP Score is naturally lower. We conducted a quantitative comparison following prior studies~\cite{Gandikota_2023_ICCV, Kumari_2023_ICCV, Gandikota_2024_WACV, Zhang_2024_CVPR, basu2024localizing, Lyu_2024_CVPR}. However, the metrics used in those studies do not evaluate ``how well concepts have been erased?''. To the best of our knowledge, such evaluation metrics do not currently exist.

We created the prompts using CLIP ImageNet Template small and calculated the CLIP Score. Ten images were generated for each prompt. Due to resource constraints, generating ten images simultaneously is not feasible, so five images were generated at a time with seed values of 0 and 2024. Generation was executed using DDIM~\cite{song2021denoising} Scheduler with 7.5 guidance scale and 50 inference steps.

\begin{table}[htbp]
\caption{Comparison of the CLIP Scores}
\label{tab:clip-score}
\centering
\begin{small}
\begin{tabular}{lccc}
\toprule
\multicolumn{4}{c}{Erasing \textit{Eiffel Tower}} \\
\midrule
& Eiffel Tower & Tokyo Tower & Triumphal arch \\
\midrule
Original SD & 0.256 & 0.283 & 0.286 \\
ESD-x-1 & 0.111 & 0.177 & 0.274 \\
UCE & 0.213 & 0.241 & 0.284 \\
SPM & 0.190 & 0.277 & 0.283 \\
Ours & 0.212 & 0.262 & 0.281 \\
\midrule
\multicolumn{4}{c}{Erasing \textit{Monet Style}} \\
\midrule
& Monet Style & Gogh Style & Picasso Style \\
\midrule
Original SD & 0.276 & 0.264 & 0.267 \\
ESD-x-1 & 0.276 & 0.264 & 0.267 \\
UCE & 0.273 & 0.264 & 0.267 \\
SPM & 0.235 & 0.263 & 0.267 \\
Ours & 0.179 & 0.207 & 0.241 \\
\bottomrule
\end{tabular}
\end{small}
\end{table}

\cref{tab:clip-score} lists the results. Following previous studies, it can be observed that our proposed method is capable of erasing styles more effectively. Regarding objects, while successfully removing the target concept, our method maintained CLIP Scores for other concepts as much as possible. However, the results of the quantitative evaluation in \cref{tab:clip-score} do not align with the results presented in \cref{fig:comparison}. Similar to the earlier discussion, upon examining the generated Eiffel Tower images in \cref{fig:comparison}, it is evident that the Eiffel Tower was erased with all methods. According to previous studies, ESD-x-1, which had the lowest CLIP Score, would be considered the best-performing method. However, it is difficult to claim that ESD-x-1 exhibits the best performance through qualitative comparison. Therefore, we believe that using CLIP Score for evaluating concept erasure is not be appropriate. The same argument applies when using evaluation metrics such as FID or Learned perceptual Image Patch Similarity (LPIPS)~\cite{Zhang_2018_CVPR}. LPIPS is a metric that represents the distance from the Original SD. However, in the context of machine unlearning, a retrain model is the gold standard as the ground truth~\cite{thudi2022unrolling, jia2023model, fan2024salun}. If FID or LPIPS were to be used, a retrain model would need to be prepared, which is impractical as an evaluation metric.

\section{Ablation Studies}
We decomposed our proposed method into several elements and conducted additional experiments to confirm the effectiveness of the elements. 

\subsection{$k$-shot Erasing}
\begin{figure}[!htbp]
\centering
\includegraphics[width=0.9\linewidth]{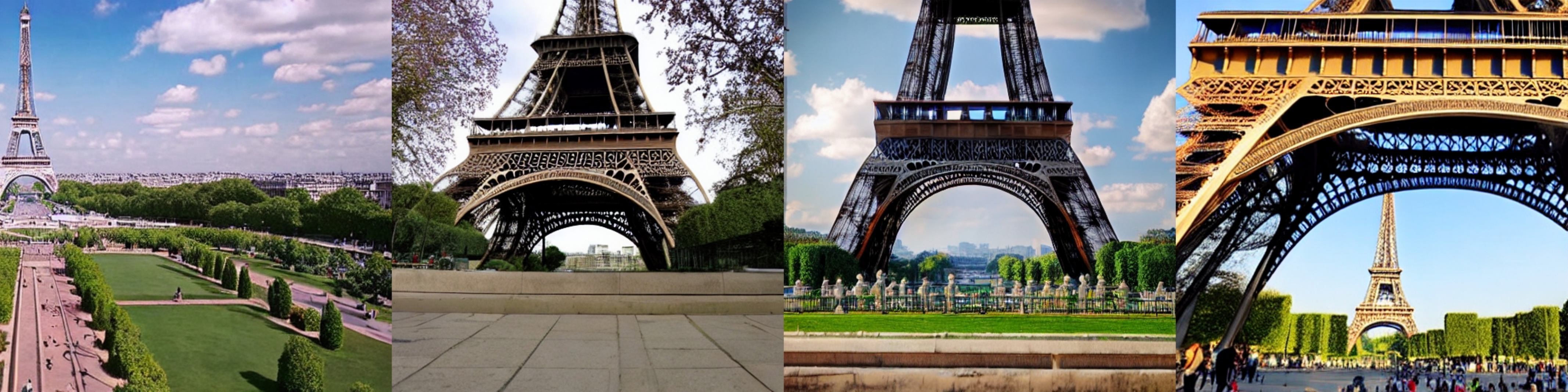}
\caption{Generated images in case with zero-shot erasing}
\label{fig:0-shot-eiffel}
\end{figure}

\begin{figure}[!htbp]
\centering
\includegraphics[width=0.9\linewidth]{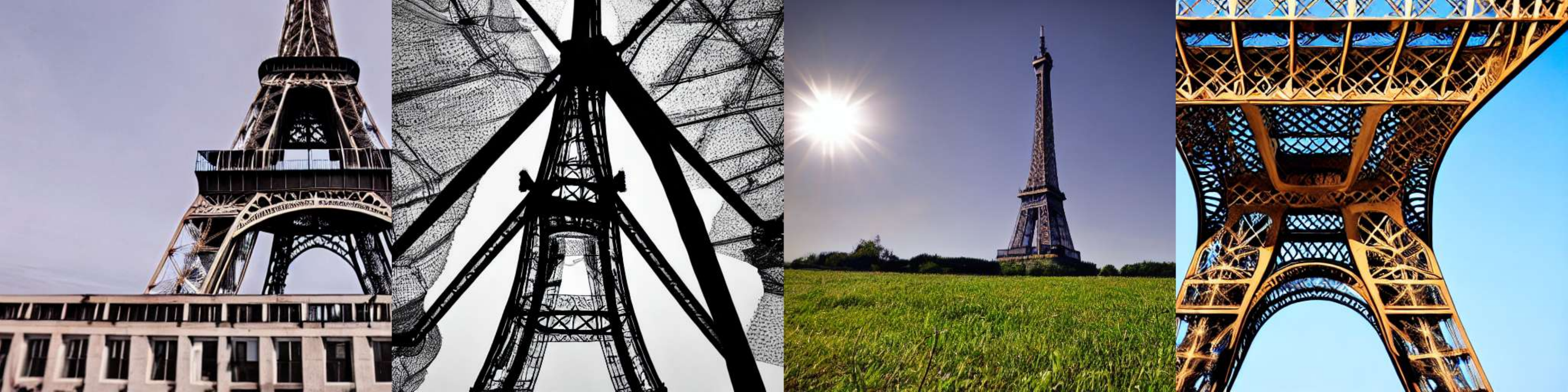}
\caption{Generated images in case with two-shot erasing}
\label{fig:2-shot-eiffel}
\end{figure}

We consider reducing the number of images provided during training. We conducted these experiments with zero-shot and two-shot. \cref{fig:0-shot-eiffel} shows the results with zero-shot, and \cref{fig:2-shot-eiffel} shows the results with two-shot. In both cases, the ``Eiffel Tower'' is still present. However, with zero-shot, it is clearly recognizable, while with two-shot, some images make it difficult to distinguish the Eiffel Tower. This indicates the need for diverse images.

As evident from \cref{fig:comparison-epochs}, with four-shot, the concept is mostly disappeared after the second epoch (four iterations). From this observation, we consider that the number of iterations is also sufficient for zero and two-shot setting.

\subsection{Number of Epochs}
\label{subsec:num-epochs}
\begin{figure}[!htbp]
\centering
\includegraphics[width=\linewidth]{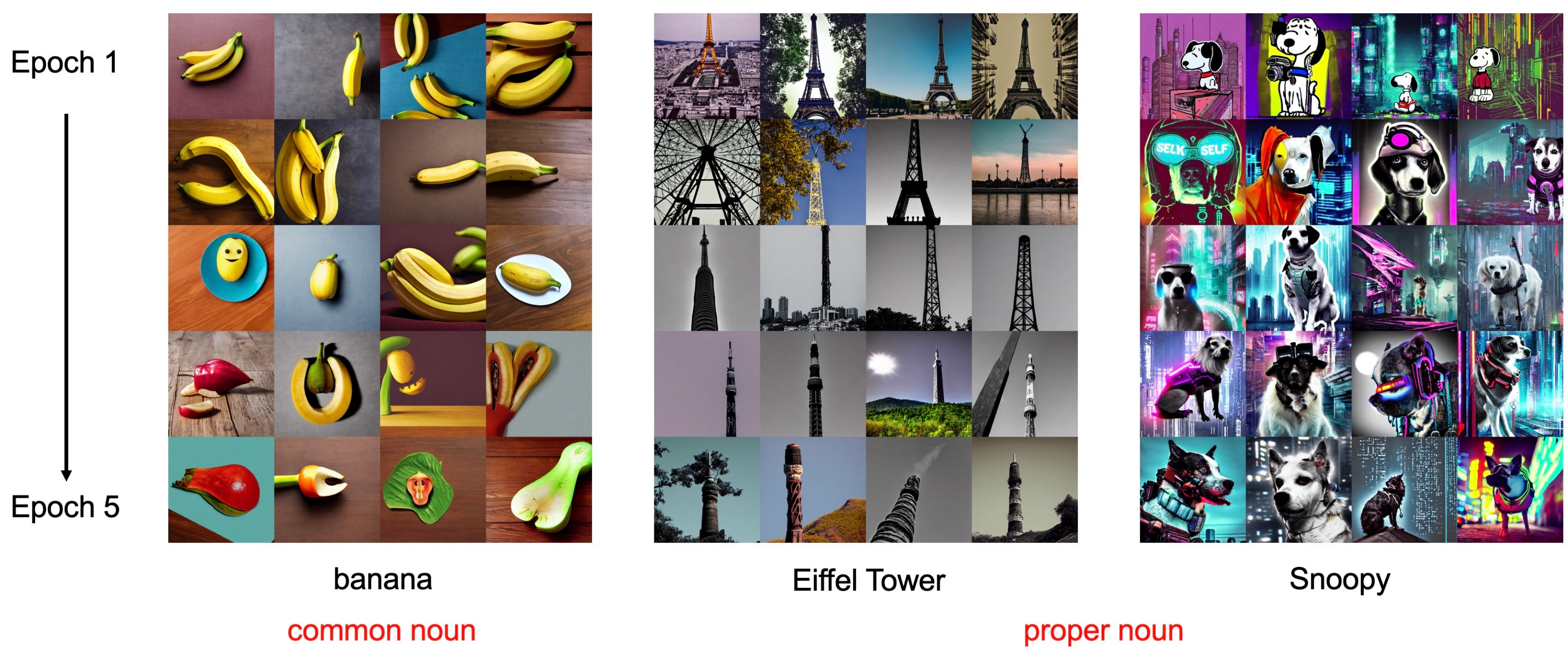}
\caption{Transition due to change in number of epochs. Top row is the result at end of first epoch and bottom row is result at end of fifth epoch. Captions we used in generating are ``a banana on the table'', ``a photo of Eiffel Tower'' and ``Snoopy in cyberpunk style'' respectively.}
\label{fig:comparison-epochs}
\end{figure}

We confirm the transition over epochs using ``Snoopy'', ``Eiffel Tower'', and ``banana''. Five epochs were trained for all concepts in relation to \cref{subsec:update-parameters}. \cref{fig:comparison-epochs} shows the results. The concepts ``Snoopy'' and ``Eiffel Tower'', which are proper nouns, disappeared early on. Therefore, there is no possibility of concept erasure failure due to the low number of iterations in the two-shot setting mentioned in the previous subsection. For ``banana'', however, the concept disappeared at the end of the fifth epoch. This indicates that the difficulty of the task differs between common and proper nouns. It can be inferred that more common nouns than proper nouns are included in the training data when training with CLIP and stable diffusion. This suggests that it is more difficult to erase knowledge that is more deeply rooted in the model.

\subsection{Real or Synthesized Images}
\begin{figure}[!htbp]
\centering
\includegraphics[width=0.9\linewidth]{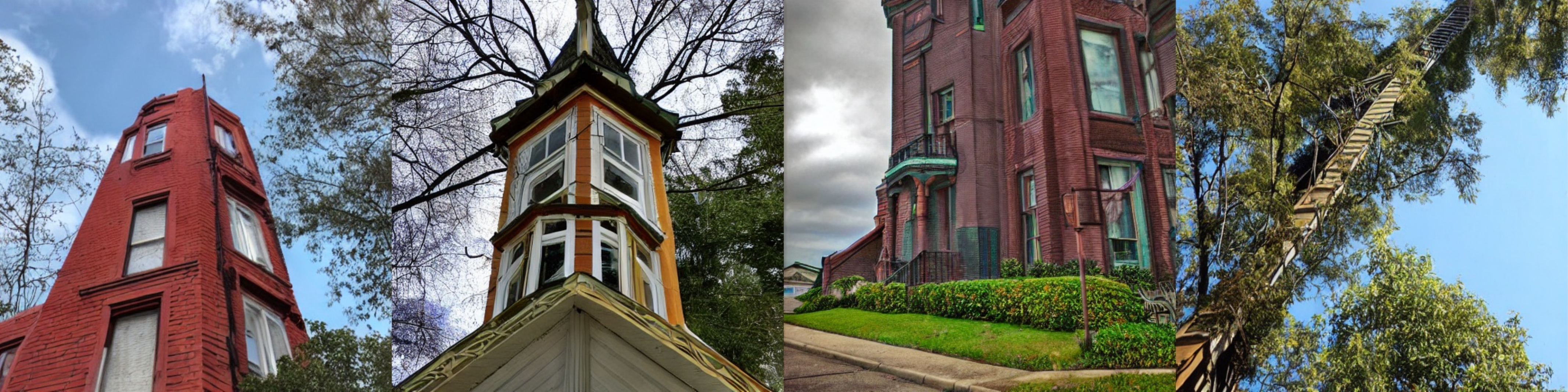}
\caption{Results when using synthesized images}
\label{fig:syn-eiffel}
\end{figure}

All experiments were conducted using real images. This experimental setting is based on our belief that a retrain model serves as the ground truth. In this case, it is most appropriate to use a subset of the dataset $\mathcal{D}_{f}$ containing the concepts to erase. However, experiments can be conducted using synthesized images. Therefore, we conducted experiments using four synthesized images (referred as \cref{fig:comparison}). The results are shown in \cref{fig:syn-eiffel}. Eiffel Tower was completely absent in some images, while in others, it is unclear what is being generated. Additionally, while in real images, the concept of ``Eiffel Tower'' was mapped to ``Tower'', in the case of generated images, ``Eiffel Tower'' was mapped to ``Building''. Since the appropriateness of each approach depends largely on subjective interpretation, we cannot make a definitive assertion. However, on the basis of the idea of slight change in text encoders, we believe that ``Eiffel Tower'' should be mapped to the ``Tower''.

\subsection{Update Parameters}
\label{subsec:update-parameters}
There are studies suggesting that knowledge acquired during training is stored in MLPs~\cite{NEURIPS2022_6f1d43d5, dai-etal-2022-knowledge, geva-etal-2021-transformer} and in the first self-attention modules~\cite{basu2024localizing} of transformer-based models, but it is unclear which is more effective in text-to-image tasks. We conducted experiments by limiting the parameters updated to either MLPs or self-attention modules to determine which part of CLIP contributes to knowledge updating. We presented the results from full parameter tuning. In the first self-attention layer, there are four trainable weight matrices: $W_q$, $W_k$, $W_v$, and $W_{out}$. We conducted the experiments under two settings: (i) updating only $W_{out}$ following the approach of DiffQuickFix~\cite{basu2024localizing} and (ii) updating all four matrices.

\begin{figure*}[!htbp]
\centering
\includegraphics[width=\linewidth]{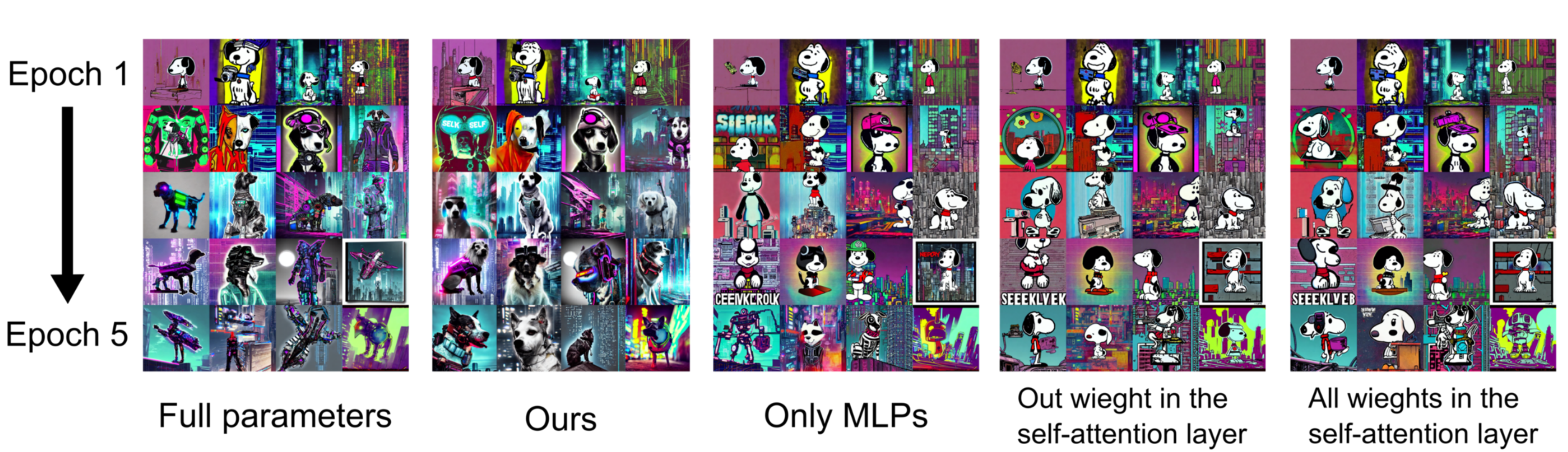}
\caption{Change in concept erasure depending on parameters to be updated}
\label{fig:parameters}
\end{figure*}

The results are presented in \cref{fig:parameters}. In the updates of the first self-attention layer, the concept of ``Snoopy'' was not erased in either setting, and almost identical results were obtained. It can be inferred that slightly updating only the weights of the first self-attention layer contributes minimally to concept erasure. When only updating MLPs, it can be observed that the concept disappeared by the end of the fifth epoch. This suggests that knowledge is accumulated in MLPs. It is evident that the concept disappeared at earlier stages with both our method and the full-parameter-tuning method. Therefore, it can be inferred that there are parameters related to knowledge other than MLPs. In comparison with full parameter tuning, the transition of generated images is generally the same. Up to the second epoch, the results were mostly identical when viewed as a whole despite some differences in detail. From this observation, the key other than MLPs is the final self-attention layer.

This comparison suggests that even with a transformer-based architecture such as CLIP trained on image-text pairs, the primary knowledge may be accumulated in MLPs. This indicates that in text-to-image diffusion models, using a transformer-based text encoder, regardless of the specific model (such as T5, BERT~\cite{devlin-etal-2019-bert}), updating the parameters of MLPs may enable concept erasure. However, since we do not have access to the weights of models trained using text encoders trained only with text, we cannot verify this.

\section{Limitations}
\label{sec:limitations}
\begin{figure}[!htbp]
\centering
\begin{tabular}{@{}cc@{}}
\begin{tabular}{@{}c@{}}
parachute \\
\end{tabular} &
\includegraphics[width=0.75\linewidth,valign=c]{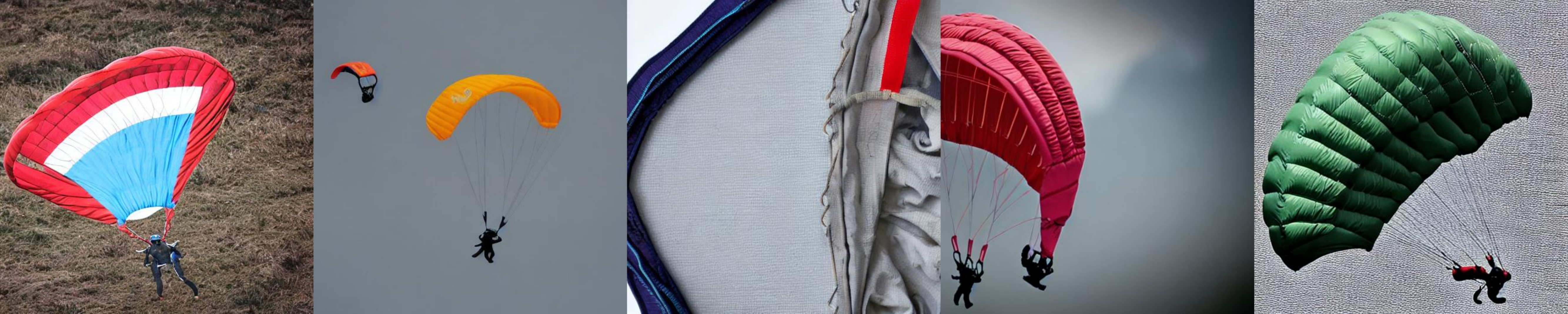} \\ \\
\begin{tabular}{@{}c@{}}
church \\
\end{tabular} &
\includegraphics[width=0.75\linewidth,valign=c]{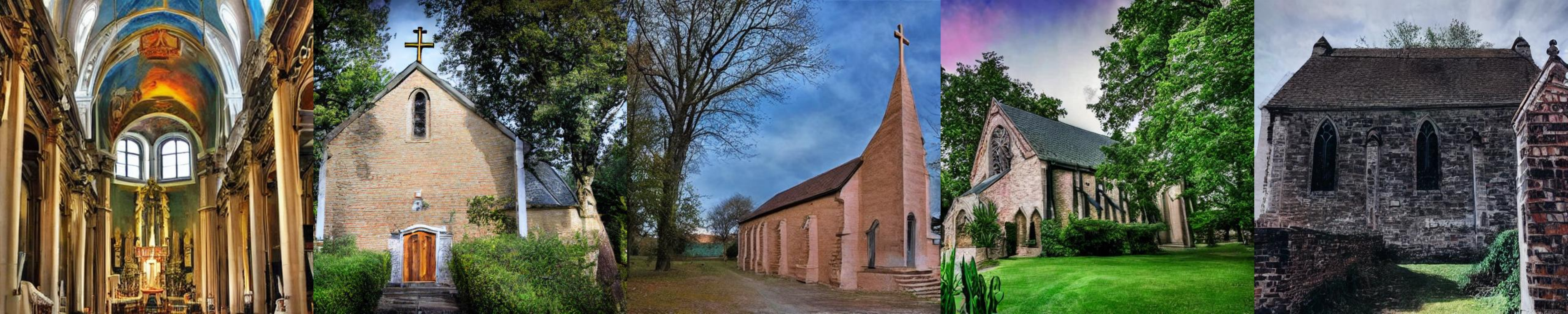}
\end{tabular}
\caption{Cases of failed erasing of parachute and church}
\label{fig:fails}
\end{figure}

The experiments presented in \cref{sec:exp} demonstrated the effectiveness of our proposed method. However, challenges remain for more general concept erasure. In \cref{fig:fails}, we show the cases of failed erasing. We consider ``church'' as an example, this failure. It may stem from the large domain represented by ``church'' in the text encoder's latent space. Since our method aims to map to similar semantic spaces by fine-tuning the text encoder, it becomes challenging to remove concepts associated with large semantic spaces. In such cases, combining our method with other concept erasure methods is feasible. DiffQuickFix~\cite{basu2024localizing} updates one of the weight matrices in the first self-attention layer, which is not updated with our proposed method. Hence, it can be effectively combined with our method to improve concept erasure in challenging cases.
    
\section{Related Works}
\subsection{Text-to-Image Diffusion Models and Text Encoder}
Denoising diffusion models~\cite{pmlr-v37-sohl-dickstein15, NEURIPS2020_4c5bcfec} have achieved success even on large-scale datasets with high variance, such as ImageNet~\cite{5206848}, by using a simple objective function. They have also demonstrated the ability to handle various resolutions~\cite{JMLR:v23:21-0635}. While it had been necessary to prepare a separate classifier for conditional generation tasks~\cite{dhariwal2021diffusion}, recent advancements have enabled image generation without the need for a classifier~\cite{ho2021classifierfree}.

CLIP~\cite{pmlr-v139-radford21a}, trained on a large corpus of image-text pairs from the Internet, has established a robust connection between images and natural language. The evolution of image generation with natural language has been accelerated by the fusion of CLIP and diffusion models~\cite{saharia2022photorealistic, ramesh2022hierarchical, openai2023dalle-3, Rombach_2022_CVPR}. Research has also been conducted on how to transfer natural-language information to generation models. Qualitatively, models trained with more natural-language data outperformed CLIP, even if there was no quantitative difference. The scaling effect of U-Net in generating images was also found to be marginal~\cite{saharia2022photorealistic}. Combining CLIP with other models allows for handling both visual and linguistic information more effectively~\cite{balaji2023ediffi}. In Stable Diffusion 3~\cite{pmlr-v235-esser24a}, the adoption of three text encoders significantly contributes to faithful generation aligned with prompts, owing to high-quality text embeddings.

We focused on the quality of text conditions in the text-to-image diffusion models. This idea is inspired by the notion that improving the quality of image captions enhances the overall quality of text-to-image generation. Specifically, we intentionally degrade the quality of text embeddings used for conditioning U-Net to attempt concept erasure.

\subsection{Erasing Concepts from Text-to-Image Diffusion Models}
Many image-generation models are trained on a vast amount of data collected from the Internet. Consequently, such datasets often contain undesired images, such as those containing NSFW content or copyrighted material, necessitating measures to address these issues when deploying the models to the market. Several studies have been conducted on methods to remove specific concepts or objects from generated images, many of which involve modifying the weights associated with U-Net, an image generative module~\cite{Gandikota_2023_ICCV, Kumari_2023_ICCV, Gandikota_2024_WACV, Zhang_2024_CVPR}. 
Some of these methods update the parameters of U-Net itself or adapters attached to U-Net~\cite{Gandikota_2023_ICCV, Kumari_2023_ICCV, zhao2024separable, Lu_2024_CVPR, heng2023selective, kim2023safeselfdistillationinternetscaletexttoimage, huang2024recelerreliableconcepterasing}, update the weights of cross-attention modules conditioning the U-Net by closed-form equation~\cite{Zhang_2024_CVPR, Gandikota_2024_WACV, Lu_2024_CVPR}, and induce the prevention of harmful content being generated when generating~\cite{Schramowski_2023_CVPR}.
There are also methods for investigating the flow of knowledge between U-Net and the text encoder, updating only parts of the text encoder~\cite{basu2024localizing}. Particularly when updating parameters associated with U-Net, there is a possibility of decreasing the image fidelity of the generated models, as well as potential reductions in the transmission capability of conditions. The computational complexity of U-Net can lead to time-consuming concept erasure. Most methods require the anchor concepts like the target concept. These often necessitate human or large language models intervention for their generation, without clear guidelines on their quality.

Since our method automatically sets the anchor concept in accordance with the latent space of the text encoder, the quality of the anchor concept does not affect the ablation result.

\subsection{Knowledge in Transformer-based Models}
Research have progressed in understanding where the knowledge of transformer-based large language models is accumulated. Large language models trained only on natural language are believed to accumulate knowledge in the MLP blocks in the transformer~\cite{NEURIPS2022_6f1d43d5, dai-etal-2022-knowledge, geva-etal-2021-transformer}, and updating the parameters of MLP enables knowledge editing~\cite{yao2022kformer, dong-etal-2022-calibrating}. However, regarding the text encoder of CLIP trained using both natural language and images, findings also suggesting that knowledge is primarily accumulated in the first self-attention layer~\cite{basu2024localizing}.

We have also obtained results suggesting similar knowledge accumulation in encoder-only models by treating the CLIP text encoder as a transformer-based language model and limiting the parameters to update. Our research suggests that, similar to prior studies, knowledge editing is achievable by updating MLP blocks. We also confirm that more effective concept erasure is possible by updating not only MLP blocks but also the final self-attention layer.

\section{Conclusion}
We interpreted concept erasure from text-to-image diffusion models as the disruption of text-image alignment and introduced a method for achieving this by making minor changes to the text encoder. Our method, which involves making slight modifications to the CLIP text encoder, requires fewer updates compared with current methods. Therefore, concept erasure can be executed very rapidly. When the text encoder undergoes minor changes, it is mapped to concepts that are closer in the latent space of the text encoder, leading to more natural changes than those generated by human-selected anchor concepts. In our experiments, we confirmed that several target concepts disappeared while suppressing the effect on other concepts. By varying the parameters to update, we confirmed that knowledge is primarily accumulated in the MLP blocks of the transformer, consistent with previous research. Like GPT and BERT, it has been suggested that knowledge accumulates in the MLP blocks of the transformer regardless of the training method. We also highlighted the inadequacy of the evaluation metrics used in previous research in the context of machine unlearning, emphasizing the need for proper evaluation.

In terms of future directions, there is potential for developing evaluation metrics for cases in which creating a retrain model is not feasible. While we used a very simple method of gradient ascent, there is scope for applying various other methods. Using saliency maps-based methods, such as SalUn~\cite{fan2024salun}, could be beneficial, as they adaptively control the updated parameters, potentially leading to more effective editing when combined with our proposed method. By updating the text encoder by using adapter tuning, similar to SPM, it is likely to achieve similar functionality to facilitated transport.

\bibliography{example_paper}

\begin{thebibliography}{72}
\providecommand{\natexlab}[1]{#1}
\providecommand{\url}[1]{\texttt{#1}}
\expandafter\ifx\csname urlstyle\endcsname\relax
  \providecommand{\doi}[1]{doi: #1}\else
  \providecommand{\doi}{doi: \begingroup \urlstyle{rm}\Url}\fi

\bibitem[Balaji et~al.(2023)Balaji, Nah, Huang, Vahdat, Song, Zhang, Kreis,
  Aittala, Aila, Laine, Catanzaro, Karras, and Liu]{balaji2023ediffi}
Balaji, Y., Nah, S., Huang, X., Vahdat, A., Song, J., Zhang, Q., Kreis, K.,
  Aittala, M., Aila, T., Laine, S., Catanzaro, B., Karras, T., and Liu, M.-Y.
\newblock ediff-i: Text-to-image diffusion models with an ensemble of expert
  denoisers, 2023.

\bibitem[Basu et~al.(2024)Basu, Zhao, Morariu, Feizi, and
  Manjunatha]{basu2024localizing}
Basu, S., Zhao, N., Morariu, V.~I., Feizi, S., and Manjunatha, V.
\newblock Localizing and editing knowledge in text-to-image generative models.
\newblock In \emph{The Twelfth International Conference on Learning
  Representations}, 2024.
\newblock URL \url{https://openreview.net/forum?id=Qmw9ne6SOQ}.

\bibitem[Bommasani et~al.(2022)Bommasani, Hudson, Adeli, Altman, Arora, von
  Arx, Bernstein, Bohg, Bosselut, Brunskill, Brynjolfsson, Buch, Card,
  Castellon, Chatterji, Chen, Creel, Davis, Demszky, Donahue, Doumbouya,
  Durmus, Ermon, Etchemendy, Ethayarajh, Fei-Fei, Finn, Gale, Gillespie, Goel,
  Goodman, Grossman, Guha, Hashimoto, Henderson, Hewitt, Ho, Hong, Hsu, Huang,
  Icard, Jain, Jurafsky, Kalluri, Karamcheti, Keeling, Khani, Khattab, Koh,
  Krass, Krishna, Kuditipudi, Kumar, Ladhak, Lee, Lee, Leskovec, Levent, Li,
  Li, Ma, Malik, Manning, Mirchandani, Mitchell, Munyikwa, Nair, Narayan,
  Narayanan, Newman, Nie, Niebles, Nilforoshan, Nyarko, Ogut, Orr,
  Papadimitriou, Park, Piech, Portelance, Potts, Raghunathan, Reich, Ren, Rong,
  Roohani, Ruiz, Ryan, Ré, Sadigh, Sagawa, Santhanam, Shih, Srinivasan,
  Tamkin, Taori, Thomas, Tramèr, Wang, Wang, Wu, Wu, Wu, Xie, Yasunaga, You,
  Zaharia, Zhang, Zhang, Zhang, Zhang, Zheng, Zhou, and
  Liang]{bommasani2022opportunities}
Bommasani, R., Hudson, D.~A., Adeli, E., Altman, R., Arora, S., von Arx, S.,
  Bernstein, M.~S., Bohg, J., Bosselut, A., Brunskill, E., Brynjolfsson, E.,
  Buch, S., Card, D., Castellon, R., Chatterji, N., Chen, A., Creel, K., Davis,
  J.~Q., Demszky, D., Donahue, C., Doumbouya, M., Durmus, E., Ermon, S.,
  Etchemendy, J., Ethayarajh, K., Fei-Fei, L., Finn, C., Gale, T., Gillespie,
  L., Goel, K., Goodman, N., Grossman, S., Guha, N., Hashimoto, T., Henderson,
  P., Hewitt, J., Ho, D.~E., Hong, J., Hsu, K., Huang, J., Icard, T., Jain, S.,
  Jurafsky, D., Kalluri, P., Karamcheti, S., Keeling, G., Khani, F., Khattab,
  O., Koh, P.~W., Krass, M., Krishna, R., Kuditipudi, R., Kumar, A., Ladhak,
  F., Lee, M., Lee, T., Leskovec, J., Levent, I., Li, X.~L., Li, X., Ma, T.,
  Malik, A., Manning, C.~D., Mirchandani, S., Mitchell, E., Munyikwa, Z., Nair,
  S., Narayan, A., Narayanan, D., Newman, B., Nie, A., Niebles, J.~C.,
  Nilforoshan, H., Nyarko, J., Ogut, G., Orr, L., Papadimitriou, I., Park,
  J.~S., Piech, C., Portelance, E., Potts, C., Raghunathan, A., Reich, R., Ren,
  H., Rong, F., Roohani, Y., Ruiz, C., Ryan, J., Ré, C., Sadigh, D., Sagawa,
  S., Santhanam, K., Shih, A., Srinivasan, K., Tamkin, A., Taori, R., Thomas,
  A.~W., Tramèr, F., Wang, R.~E., Wang, W., Wu, B., Wu, J., Wu, Y., Xie,
  S.~M., Yasunaga, M., You, J., Zaharia, M., Zhang, M., Zhang, T., Zhang, X.,
  Zhang, Y., Zheng, L., Zhou, K., and Liang, P.
\newblock On the opportunities and risks of foundation models, 2022.

\bibitem[Brock et~al.(2019)Brock, Donahue, and Simonyan]{brock2018large}
Brock, A., Donahue, J., and Simonyan, K.
\newblock Large scale {GAN} training for high fidelity natural image synthesis.
\newblock In \emph{International Conference on Learning Representations}, 2019.
\newblock URL \url{https://openreview.net/forum?id=B1xsqj09Fm}.

\bibitem[Chen et~al.(2023)Chen, Hu, LI, Ruiz, Jia, Chang, and
  Cohen]{chen2023subjectdriven}
Chen, W., Hu, H., LI, Y., Ruiz, N., Jia, X., Chang, M.-W., and Cohen, W.~W.
\newblock Subject-driven text-to-image generation via apprenticeship learning.
\newblock In \emph{Thirty-seventh Conference on Neural Information Processing
  Systems}, 2023.
\newblock URL \url{https://openreview.net/forum?id=wv3bHyQbX7}.

\bibitem[Dai et~al.(2022)Dai, Dong, Hao, Sui, Chang, and
  Wei]{dai-etal-2022-knowledge}
Dai, D., Dong, L., Hao, Y., Sui, Z., Chang, B., and Wei, F.
\newblock Knowledge neurons in pretrained transformers.
\newblock In Muresan, S., Nakov, P., and Villavicencio, A. (eds.),
  \emph{Proceedings of the 60th Annual Meeting of the Association for
  Computational Linguistics (Volume 1: Long Papers)}, pp.\  8493--8502, Dublin,
  Ireland, May 2022. Association for Computational Linguistics.
\newblock \doi{10.18653/v1/2022.acl-long.581}.
\newblock URL \url{https://aclanthology.org/2022.acl-long.581}.

\bibitem[Dai et~al.(2023)Dai, Hou, Ma, Tsai, Wang, Wang, Zhang, Vandenhende,
  Wang, Dubey, Yu, Kadian, Radenovic, Mahajan, Li, Zhao, Petrovic, Singh,
  Motwani, Wen, Song, Sumbaly, Ramanathan, He, Vajda, and Parikh]{dai2023emu}
Dai, X., Hou, J., Ma, C.-Y., Tsai, S., Wang, J., Wang, R., Zhang, P.,
  Vandenhende, S., Wang, X., Dubey, A., Yu, M., Kadian, A., Radenovic, F.,
  Mahajan, D., Li, K., Zhao, Y., Petrovic, V., Singh, M.~K., Motwani, S., Wen,
  Y., Song, Y., Sumbaly, R., Ramanathan, V., He, Z., Vajda, P., and Parikh, D.
\newblock Emu: Enhancing image generation models using photogenic needles in a
  haystack, 2023.

\bibitem[Deng et~al.(2009)Deng, Dong, Socher, Li, Li, and Fei-Fei]{5206848}
Deng, J., Dong, W., Socher, R., Li, L.-J., Li, K., and Fei-Fei, L.
\newblock Imagenet: A large-scale hierarchical image database.
\newblock In \emph{2009 IEEE Conference on Computer Vision and Pattern
  Recognition}, pp.\  248--255, 2009.
\newblock \doi{10.1109/CVPR.2009.5206848}.

\bibitem[Devlin et~al.(2019)Devlin, Chang, Lee, and
  Toutanova]{devlin-etal-2019-bert}
Devlin, J., Chang, M.-W., Lee, K., and Toutanova, K.
\newblock {BERT}: Pre-training of deep bidirectional transformers for language
  understanding.
\newblock In Burstein, J., Doran, C., and Solorio, T. (eds.), \emph{Proceedings
  of the 2019 Conference of the North {A}merican Chapter of the Association for
  Computational Linguistics: Human Language Technologies, Volume 1 (Long and
  Short Papers)}, pp.\  4171--4186, Minneapolis, Minnesota, June 2019.
  Association for Computational Linguistics.
\newblock \doi{10.18653/v1/N19-1423}.
\newblock URL \url{https://aclanthology.org/N19-1423}.

\bibitem[Dhariwal \& Nichol(2021)Dhariwal and Nichol]{dhariwal2021diffusion}
Dhariwal, P. and Nichol, A.~Q.
\newblock Diffusion models beat {GAN}s on image synthesis.
\newblock In Beygelzimer, A., Dauphin, Y., Liang, P., and Vaughan, J.~W.
  (eds.), \emph{Advances in Neural Information Processing Systems}, 2021.
\newblock URL \url{https://openreview.net/forum?id=AAWuCvzaVt}.

\bibitem[Dong et~al.(2022)Dong, Dai, Song, Xu, Sui, and
  Li]{dong-etal-2022-calibrating}
Dong, Q., Dai, D., Song, Y., Xu, J., Sui, Z., and Li, L.
\newblock Calibrating factual knowledge in pretrained language models.
\newblock In Goldberg, Y., Kozareva, Z., and Zhang, Y. (eds.), \emph{Findings
  of the Association for Computational Linguistics: EMNLP 2022}, pp.\
  5937--5947, Abu Dhabi, United Arab Emirates, December 2022. Association for
  Computational Linguistics.
\newblock \doi{10.18653/v1/2022.findings-emnlp.438}.
\newblock URL \url{https://aclanthology.org/2022.findings-emnlp.438}.

\bibitem[Esser et~al.(2024)Esser, Kulal, Blattmann, Entezari, M\"{u}ller,
  Saini, Levi, Lorenz, Sauer, Boesel, Podell, Dockhorn, English, and
  Rombach]{pmlr-v235-esser24a}
Esser, P., Kulal, S., Blattmann, A., Entezari, R., M\"{u}ller, J., Saini, H.,
  Levi, Y., Lorenz, D., Sauer, A., Boesel, F., Podell, D., Dockhorn, T.,
  English, Z., and Rombach, R.
\newblock Scaling rectified flow transformers for high-resolution image
  synthesis.
\newblock In Salakhutdinov, R., Kolter, Z., Heller, K., Weller, A., Oliver, N.,
  Scarlett, J., and Berkenkamp, F. (eds.), \emph{Proceedings of the 41st
  International Conference on Machine Learning}, volume 235 of
  \emph{Proceedings of Machine Learning Research}, pp.\  12606--12633. PMLR,
  21--27 Jul 2024.
\newblock URL \url{https://proceedings.mlr.press/v235/esser24a.html}.

\bibitem[Fan et~al.(2024)Fan, Liu, Zhang, Wong, Wei, and Liu]{fan2024salun}
Fan, C., Liu, J., Zhang, Y., Wong, E., Wei, D., and Liu, S.
\newblock Salun: Empowering machine unlearning via gradient-based weight
  saliency in both image classification and generation.
\newblock In \emph{The Twelfth International Conference on Learning
  Representations}, 2024.
\newblock URL \url{https://openreview.net/forum?id=gn0mIhQGNM}.

\bibitem[Fang et~al.(2024)Fang, Sun, Wang, Huang, Wang, and
  Cao]{FANG2024105171}
Fang, Y., Sun, Q., Wang, X., Huang, T., Wang, X., and Cao, Y.
\newblock Eva-02: A visual representation for neon genesis.
\newblock \emph{Image and Vision Computing}, pp.\  105171, 2024.
\newblock ISSN 0262-8856.
\newblock \doi{https://doi.org/10.1016/j.imavis.2024.105171}.
\newblock URL
  \url{https://www.sciencedirect.com/science/article/pii/S0262885624002762}.

\bibitem[Fei et~al.(2023)Fei, Fan, and Huang]{10.1145/3581783.3612599}
Fei, Z., Fan, M., and Huang, J.
\newblock Gradient-free textual inversion.
\newblock In \emph{Proceedings of the 31st ACM International Conference on
  Multimedia}, MM '23, pp.\  1364--1373, New York, NY, USA, 2023. Association
  for Computing Machinery.
\newblock ISBN 9798400701085.
\newblock \doi{10.1145/3581783.3612599}.
\newblock URL \url{https://doi.org/10.1145/3581783.3612599}.

\bibitem[Gal et~al.(2023)Gal, Alaluf, Atzmon, Patashnik, Bermano, Chechik, and
  Cohen-or]{gal2023an}
Gal, R., Alaluf, Y., Atzmon, Y., Patashnik, O., Bermano, A.~H., Chechik, G.,
  and Cohen-or, D.
\newblock An image is worth one word: Personalizing text-to-image generation
  using textual inversion.
\newblock In \emph{The Eleventh International Conference on Learning
  Representations}, 2023.
\newblock URL \url{https://openreview.net/forum?id=NAQvF08TcyG}.

\bibitem[Gandikota et~al.(2023)Gandikota, Materzynska, Fiotto-Kaufman, and
  Bau]{Gandikota_2023_ICCV}
Gandikota, R., Materzynska, J., Fiotto-Kaufman, J., and Bau, D.
\newblock Erasing concepts from diffusion models.
\newblock In \emph{Proceedings of the IEEE/CVF International Conference on
  Computer Vision (ICCV)}, pp.\  2426--2436, October 2023.

\bibitem[Gandikota et~al.(2024)Gandikota, Orgad, Belinkov, Materzy\'nska, and
  Bau]{Gandikota_2024_WACV}
Gandikota, R., Orgad, H., Belinkov, Y., Materzy\'nska, J., and Bau, D.
\newblock Unified concept editing in diffusion models.
\newblock In \emph{Proceedings of the IEEE/CVF Winter Conference on
  Applications of Computer Vision (WACV)}, pp.\  5111--5120, January 2024.

\bibitem[Geva et~al.(2021)Geva, Schuster, Berant, and
  Levy]{geva-etal-2021-transformer}
Geva, M., Schuster, R., Berant, J., and Levy, O.
\newblock Transformer feed-forward layers are key-value memories.
\newblock In Moens, M.-F., Huang, X., Specia, L., and Yih, S. W.-t. (eds.),
  \emph{Proceedings of the 2021 Conference on Empirical Methods in Natural
  Language Processing}, pp.\  5484--5495, Online and Punta Cana, Dominican
  Republic, November 2021. Association for Computational Linguistics.
\newblock \doi{10.18653/v1/2021.emnlp-main.446}.
\newblock URL \url{https://aclanthology.org/2021.emnlp-main.446}.

\bibitem[Goodfellow et~al.(2014)Goodfellow, Pouget-Abadie, Mirza, Xu,
  Warde-Farley, Ozair, Courville, and Bengio]{NIPS2014_5ca3e9b1}
Goodfellow, I., Pouget-Abadie, J., Mirza, M., Xu, B., Warde-Farley, D., Ozair,
  S., Courville, A., and Bengio, Y.
\newblock Generative adversarial nets.
\newblock In Ghahramani, Z., Welling, M., Cortes, C., Lawrence, N., and
  Weinberger, K. (eds.), \emph{Advances in Neural Information Processing
  Systems}, volume~27. Curran Associates, Inc., 2014.
\newblock URL
  \url{https://proceedings.neurips.cc/paper_files/paper/2014/file/5ca3e9b122f61f8f06494c97b1afccf3-Paper.pdf}.

\bibitem[Heng \& Soh(2023)Heng and Soh]{heng2023selective}
Heng, A. and Soh, H.
\newblock Selective amnesia: A continual learning approach to forgetting in
  deep generative models.
\newblock In \emph{Thirty-seventh Conference on Neural Information Processing
  Systems}, 2023.
\newblock URL \url{https://openreview.net/forum?id=BC1IJdsuYB}.

\bibitem[Hessel et~al.(2021)Hessel, Holtzman, Forbes, Le~Bras, and
  Choi]{hessel-etal-2021-clipscore}
Hessel, J., Holtzman, A., Forbes, M., Le~Bras, R., and Choi, Y.
\newblock {CLIPS}core: A reference-free evaluation metric for image captioning.
\newblock In Moens, M.-F., Huang, X., Specia, L., and Yih, S. W.-t. (eds.),
  \emph{Proceedings of the 2021 Conference on Empirical Methods in Natural
  Language Processing}, pp.\  7514--7528, Online and Punta Cana, Dominican
  Republic, November 2021. Association for Computational Linguistics.
\newblock \doi{10.18653/v1/2021.emnlp-main.595}.
\newblock URL \url{https://aclanthology.org/2021.emnlp-main.595}.

\bibitem[Heusel et~al.(2017)Heusel, Ramsauer, Unterthiner, Nessler, and
  Hochreiter]{NIPS2017_8a1d6947}
Heusel, M., Ramsauer, H., Unterthiner, T., Nessler, B., and Hochreiter, S.
\newblock Gans trained by a two time-scale update rule converge to a local nash
  equilibrium.
\newblock In Guyon, I., Luxburg, U.~V., Bengio, S., Wallach, H., Fergus, R.,
  Vishwanathan, S., and Garnett, R. (eds.), \emph{Advances in Neural
  Information Processing Systems}, volume~30. Curran Associates, Inc., 2017.
\newblock URL
  \url{https://proceedings.neurips.cc/paper_files/paper/2017/file/8a1d694707eb0fefe65871369074926d-Paper.pdf}.

\bibitem[Ho \& Salimans(2021)Ho and Salimans]{ho2021classifierfree}
Ho, J. and Salimans, T.
\newblock Classifier-free diffusion guidance.
\newblock In \emph{NeurIPS 2021 Workshop on Deep Generative Models and
  Downstream Applications}, 2021.
\newblock URL \url{https://openreview.net/forum?id=qw8AKxfYbI}.

\bibitem[Ho et~al.(2020)Ho, Jain, and Abbeel]{NEURIPS2020_4c5bcfec}
Ho, J., Jain, A., and Abbeel, P.
\newblock Denoising diffusion probabilistic models.
\newblock In Larochelle, H., Ranzato, M., Hadsell, R., Balcan, M., and Lin, H.
  (eds.), \emph{Advances in Neural Information Processing Systems}, volume~33,
  pp.\  6840--6851. Curran Associates, Inc., 2020.
\newblock URL
  \url{https://proceedings.neurips.cc/paper_files/paper/2020/file/4c5bcfec8584af0d967f1ab10179ca4b-Paper.pdf}.

\bibitem[Ho et~al.(2022)Ho, Saharia, Chan, Fleet, Norouzi, and
  Salimans]{JMLR:v23:21-0635}
Ho, J., Saharia, C., Chan, W., Fleet, D.~J., Norouzi, M., and Salimans, T.
\newblock Cascaded diffusion models for high fidelity image generation.
\newblock \emph{Journal of Machine Learning Research}, 23\penalty0
  (47):\penalty0 1--33, 2022.
\newblock URL \url{http://jmlr.org/papers/v23/21-0635.html}.

\bibitem[Howard \& Gugger(2020)Howard and Gugger]{info11020108}
Howard, J. and Gugger, S.
\newblock Fastai: A layered api for deep learning.
\newblock \emph{Information}, 11\penalty0 (2), 2020.
\newblock ISSN 2078-2489.
\newblock \doi{10.3390/info11020108}.
\newblock URL \url{https://www.mdpi.com/2078-2489/11/2/108}.

\bibitem[Hu et~al.(2022)Hu, yelong shen, Wallis, Allen-Zhu, Li, Wang, Wang, and
  Chen]{hu2022lora}
Hu, E.~J., yelong shen, Wallis, P., Allen-Zhu, Z., Li, Y., Wang, S., Wang, L.,
  and Chen, W.
\newblock Lo{RA}: Low-rank adaptation of large language models.
\newblock In \emph{International Conference on Learning Representations}, 2022.
\newblock URL \url{https://openreview.net/forum?id=nZeVKeeFYf9}.

\bibitem[Huang et~al.(2024)Huang, Chang, Tsai, Lai, Yang, and
  Wang]{huang2024recelerreliableconcepterasing}
Huang, C.-P., Chang, K.-P., Tsai, C.-T., Lai, Y.-H., Yang, F.-E., and Wang,
  Y.-C.~F.
\newblock Receler: Reliable concept erasing of text-to-image diffusion models
  via lightweight erasers, 2024.
\newblock URL \url{https://arxiv.org/abs/2311.17717}.

\bibitem[Ilharco et~al.(2023)Ilharco, Ribeiro, Wortsman, Schmidt, Hajishirzi,
  and Farhadi]{ilharco2023editing}
Ilharco, G., Ribeiro, M.~T., Wortsman, M., Schmidt, L., Hajishirzi, H., and
  Farhadi, A.
\newblock Editing models with task arithmetic.
\newblock In \emph{The Eleventh International Conference on Learning
  Representations}, 2023.
\newblock URL \url{https://openreview.net/forum?id=6t0Kwf8-jrj}.

\bibitem[Jang et~al.(2023)Jang, Yoon, Yang, Cha, Lee, Logeswaran, and
  Seo]{jang-etal-2023-knowledge}
Jang, J., Yoon, D., Yang, S., Cha, S., Lee, M., Logeswaran, L., and Seo, M.
\newblock Knowledge unlearning for mitigating privacy risks in language models.
\newblock In Rogers, A., Boyd-Graber, J., and Okazaki, N. (eds.),
  \emph{Proceedings of the 61st Annual Meeting of the Association for
  Computational Linguistics (Volume 1: Long Papers)}, pp.\  14389--14408,
  Toronto, Canada, July 2023. Association for Computational Linguistics.
\newblock \doi{10.18653/v1/2023.acl-long.805}.
\newblock URL \url{https://aclanthology.org/2023.acl-long.805}.

\bibitem[Jayasumana et~al.(2024)Jayasumana, Ramalingam, Veit, Glasner,
  Chakrabarti, and Kumar]{Jayasumana_2024_CVPR}
Jayasumana, S., Ramalingam, S., Veit, A., Glasner, D., Chakrabarti, A., and
  Kumar, S.
\newblock Rethinking fid: Towards a better evaluation metric for image
  generation.
\newblock In \emph{Proceedings of the IEEE/CVF Conference on Computer Vision
  and Pattern Recognition (CVPR)}, pp.\  9307--9315, June 2024.

\bibitem[Jia et~al.(2023)Jia, Liu, Ram, Yao, Liu, Liu, Sharma, and
  Liu]{jia2023model}
Jia, J., Liu, J., Ram, P., Yao, Y., Liu, G., Liu, Y., Sharma, P., and Liu, S.
\newblock Model sparsity can simplify machine unlearning.
\newblock In \emph{Thirty-seventh Conference on Neural Information Processing
  Systems}, 2023.
\newblock URL \url{https://openreview.net/forum?id=0jZH883i34}.

\bibitem[Karras et~al.(2020)Karras, Laine, Aittala, Hellsten, Lehtinen, and
  Aila]{Karras_2020_CVPR}
Karras, T., Laine, S., Aittala, M., Hellsten, J., Lehtinen, J., and Aila, T.
\newblock Analyzing and improving the image quality of stylegan.
\newblock In \emph{Proceedings of the IEEE/CVF Conference on Computer Vision
  and Pattern Recognition (CVPR)}, June 2020.

\bibitem[Kim et~al.(2023)Kim, Jung, Kim, Choi, Shin, and
  Lee]{kim2023safeselfdistillationinternetscaletexttoimage}
Kim, S., Jung, S., Kim, B., Choi, M., Shin, J., and Lee, J.
\newblock Towards safe self-distillation of internet-scale text-to-image
  diffusion models, 2023.
\newblock URL \url{https://arxiv.org/abs/2307.05977}.

\bibitem[Kingma \& Ba(2017)Kingma and Ba]{kingma2017adam}
Kingma, D.~P. and Ba, J.
\newblock Adam: A method for stochastic optimization, 2017.

\bibitem[Kumari et~al.(2023)Kumari, Zhang, Wang, Shechtman, Zhang, and
  Zhu]{Kumari_2023_ICCV}
Kumari, N., Zhang, B., Wang, S.-Y., Shechtman, E., Zhang, R., and Zhu, J.-Y.
\newblock Ablating concepts in text-to-image diffusion models.
\newblock In \emph{Proceedings of the IEEE/CVF International Conference on
  Computer Vision (ICCV)}, pp.\  22691--22702, October 2023.

\bibitem[Li et~al.(2024)Li, Liu, Kag, Hu, Idelbayev, Sagar, Wang, Tulyakov, and
  Ren]{Li_2024_CVPR}
Li, Y., Liu, X., Kag, A., Hu, J., Idelbayev, Y., Sagar, D., Wang, Y., Tulyakov,
  S., and Ren, J.
\newblock Textcraftor: Your text encoder can be image quality controller.
\newblock In \emph{Proceedings of the IEEE/CVF Conference on Computer Vision
  and Pattern Recognition (CVPR)}, pp.\  7985--7995, June 2024.

\bibitem[Liu et~al.(2022)Liu, Ren, Lin, and Zhao]{liu2022pseudo}
Liu, L., Ren, Y., Lin, Z., and Zhao, Z.
\newblock Pseudo numerical methods for diffusion models on manifolds.
\newblock In \emph{International Conference on Learning Representations}, 2022.
\newblock URL \url{https://openreview.net/forum?id=PlKWVd2yBkY}.

\bibitem[Lu et~al.(2024)Lu, Wang, Li, Liu, and Kong]{Lu_2024_CVPR}
Lu, S., Wang, Z., Li, L., Liu, Y., and Kong, A. W.-K.
\newblock Mace: Mass concept erasure in diffusion models.
\newblock In \emph{Proceedings of the IEEE/CVF Conference on Computer Vision
  and Pattern Recognition (CVPR)}, pp.\  6430--6440, June 2024.

\bibitem[Lyu et~al.(2024)Lyu, Yang, Hong, Chen, Jin, He, Xue, Han, and
  Ding]{Lyu_2024_CVPR}
Lyu, M., Yang, Y., Hong, H., Chen, H., Jin, X., He, Y., Xue, H., Han, J., and
  Ding, G.
\newblock One-dimensional adapter to rule them all: Concepts diffusion models
  and erasing applications.
\newblock In \emph{Proceedings of the IEEE/CVF Conference on Computer Vision
  and Pattern Recognition (CVPR)}, pp.\  7559--7568, June 2024.

\bibitem[Meng et~al.(2022)Meng, Bau, Andonian, and
  Belinkov]{NEURIPS2022_6f1d43d5}
Meng, K., Bau, D., Andonian, A., and Belinkov, Y.
\newblock Locating and editing factual associations in gpt.
\newblock In Koyejo, S., Mohamed, S., Agarwal, A., Belgrave, D., Cho, K., and
  Oh, A. (eds.), \emph{Advances in Neural Information Processing Systems},
  volume~35, pp.\  17359--17372. Curran Associates, Inc., 2022.

\bibitem[OpenAI(2023{\natexlab{a}})]{openai2023dalle-3}
OpenAI.
\newblock Improving image generation with better captions, 2023{\natexlab{a}}.

\bibitem[OpenAI(2023{\natexlab{b}})]{openai2023gpt4}
OpenAI.
\newblock Gpt-4 technical report, 2023{\natexlab{b}}.

\bibitem[Otani et~al.(2023)Otani, Togashi, Sawai, Ishigami, Nakashima, Rahtu,
  Heikkil\"a, and Satoh]{Otani_2023_CVPR}
Otani, M., Togashi, R., Sawai, Y., Ishigami, R., Nakashima, Y., Rahtu, E.,
  Heikkil\"a, J., and Satoh, S.
\newblock Toward verifiable and reproducible human evaluation for text-to-image
  generation.
\newblock In \emph{Proceedings of the IEEE/CVF Conference on Computer Vision
  and Pattern Recognition (CVPR)}, pp.\  14277--14286, June 2023.

\bibitem[Pang et~al.(2024)Pang, Yin, Xie, Wang, Li, and Mao]{Pang_2024_CVPR}
Pang, L., Yin, J., Xie, H., Wang, Q., Li, Q., and Mao, X.
\newblock Cross initialization for face personalization of text-to-image
  models.
\newblock In \emph{Proceedings of the IEEE/CVF Conference on Computer Vision
  and Pattern Recognition (CVPR)}, pp.\  8393--8403, June 2024.

\bibitem[Parmar et~al.(2022)Parmar, Zhang, and Zhu]{parmar2021cleanfid}
Parmar, G., Zhang, R., and Zhu, J.-Y.
\newblock On aliased resizing and surprising subtleties in gan evaluation.
\newblock In \emph{CVPR}, 2022.

\bibitem[Podell et~al.(2024)Podell, English, Lacey, Blattmann, Dockhorn,
  M{\"u}ller, Penna, and Rombach]{podell2024sdxl}
Podell, D., English, Z., Lacey, K., Blattmann, A., Dockhorn, T., M{\"u}ller,
  J., Penna, J., and Rombach, R.
\newblock {SDXL}: Improving latent diffusion models for high-resolution image
  synthesis.
\newblock In \emph{The Twelfth International Conference on Learning
  Representations}, 2024.
\newblock URL \url{https://openreview.net/forum?id=di52zR8xgf}.

\bibitem[Radford et~al.(2021)Radford, Kim, Hallacy, Ramesh, Goh, Agarwal,
  Sastry, Askell, Mishkin, Clark, Krueger, and Sutskever]{pmlr-v139-radford21a}
Radford, A., Kim, J.~W., Hallacy, C., Ramesh, A., Goh, G., Agarwal, S., Sastry,
  G., Askell, A., Mishkin, P., Clark, J., Krueger, G., and Sutskever, I.
\newblock Learning transferable visual models from natural language
  supervision.
\newblock In Meila, M. and Zhang, T. (eds.), \emph{Proceedings of the 38th
  International Conference on Machine Learning}, volume 139 of
  \emph{Proceedings of Machine Learning Research}, pp.\  8748--8763. PMLR,
  18--24 Jul 2021.
\newblock URL \url{https://proceedings.mlr.press/v139/radford21a.html}.

\bibitem[Raffel et~al.(2020)Raffel, Shazeer, Roberts, Lee, Narang, Matena,
  Zhou, Li, and Liu]{10.5555/3455716.3455856}
Raffel, C., Shazeer, N., Roberts, A., Lee, K., Narang, S., Matena, M., Zhou,
  Y., Li, W., and Liu, P.~J.
\newblock Exploring the limits of transfer learning with a unified text-to-text
  transformer.
\newblock \emph{J. Mach. Learn. Res.}, 21\penalty0 (1), jan 2020.
\newblock ISSN 1532-4435.

\bibitem[Ramesh et~al.(2022)Ramesh, Dhariwal, Nichol, Chu, and
  Chen]{ramesh2022hierarchical}
Ramesh, A., Dhariwal, P., Nichol, A., Chu, C., and Chen, M.
\newblock Hierarchical text-conditional image generation with clip latents,
  2022.

\bibitem[Rombach et~al.(2022)Rombach, Blattmann, Lorenz, Esser, and
  Ommer]{Rombach_2022_CVPR}
Rombach, R., Blattmann, A., Lorenz, D., Esser, P., and Ommer, B.
\newblock High-resolution image synthesis with latent diffusion models.
\newblock In \emph{Proceedings of the IEEE/CVF Conference on Computer Vision
  and Pattern Recognition (CVPR)}, pp.\  10684--10695, June 2022.

\bibitem[Ronneberger et~al.(2015)Ronneberger, Fischer, and
  Brox]{10.1007/978-3-319-24574-4_28}
Ronneberger, O., Fischer, P., and Brox, T.
\newblock U-net: Convolutional networks for biomedical image segmentation.
\newblock In Navab, N., Hornegger, J., Wells, W.~M., and Frangi, A.~F. (eds.),
  \emph{Medical Image Computing and Computer-Assisted Intervention -- MICCAI
  2015}, pp.\  234--241, Cham, 2015. Springer International Publishing.
\newblock ISBN 978-3-319-24574-4.

\bibitem[Ruiz et~al.(2023)Ruiz, Li, Jampani, Pritch, Rubinstein, and
  Aberman]{Ruiz_2023_CVPR}
Ruiz, N., Li, Y., Jampani, V., Pritch, Y., Rubinstein, M., and Aberman, K.
\newblock Dreambooth: Fine tuning text-to-image diffusion models for
  subject-driven generation.
\newblock In \emph{Proceedings of the IEEE/CVF Conference on Computer Vision
  and Pattern Recognition (CVPR)}, pp.\  22500--22510, June 2023.

\bibitem[Saharia et~al.(2022)Saharia, Chan, Saxena, Li, Whang, Denton,
  Ghasemipour, Gontijo-Lopes, Ayan, Salimans, Ho, Fleet, and
  Norouzi]{saharia2022photorealistic}
Saharia, C., Chan, W., Saxena, S., Li, L., Whang, J., Denton, E., Ghasemipour,
  S. K.~S., Gontijo-Lopes, R., Ayan, B.~K., Salimans, T., Ho, J., Fleet, D.~J.,
  and Norouzi, M.
\newblock Photorealistic text-to-image diffusion models with deep language
  understanding.
\newblock In Oh, A.~H., Agarwal, A., Belgrave, D., and Cho, K. (eds.),
  \emph{Advances in Neural Information Processing Systems}, 2022.
\newblock URL \url{https://openreview.net/forum?id=08Yk-n5l2Al}.

\bibitem[Schramowski et~al.(2023)Schramowski, Brack, Deiseroth, and
  Kersting]{Schramowski_2023_CVPR}
Schramowski, P., Brack, M., Deiseroth, B., and Kersting, K.
\newblock Safe latent diffusion: Mitigating inappropriate degeneration in
  diffusion models.
\newblock In \emph{Proceedings of the IEEE/CVF Conference on Computer Vision
  and Pattern Recognition (CVPR)}, pp.\  22522--22531, June 2023.

\bibitem[Schuhmann et~al.(2022)Schuhmann, Beaumont, Vencu, Gordon, Wightman,
  Cherti, Coombes, Katta, Mullis, Wortsman, Schramowski, Kundurthy, Crowson,
  Schmidt, Kaczmarczyk, and Jitsev]{schuhmann2022laionb}
Schuhmann, C., Beaumont, R., Vencu, R., Gordon, C.~W., Wightman, R., Cherti,
  M., Coombes, T., Katta, A., Mullis, C., Wortsman, M., Schramowski, P.,
  Kundurthy, S.~R., Crowson, K., Schmidt, L., Kaczmarczyk, R., and Jitsev, J.
\newblock {LAION}-5b: An open large-scale dataset for training next generation
  image-text models.
\newblock In \emph{Thirty-sixth Conference on Neural Information Processing
  Systems Datasets and Benchmarks Track}, 2022.
\newblock URL \url{https://openreview.net/forum?id=M3Y74vmsMcY}.

\bibitem[Sohl-Dickstein et~al.(2015)Sohl-Dickstein, Weiss, Maheswaranathan, and
  Ganguli]{pmlr-v37-sohl-dickstein15}
Sohl-Dickstein, J., Weiss, E., Maheswaranathan, N., and Ganguli, S.
\newblock Deep unsupervised learning using nonequilibrium thermodynamics.
\newblock In Bach, F. and Blei, D. (eds.), \emph{Proceedings of the 32nd
  International Conference on Machine Learning}, volume~37 of \emph{Proceedings
  of Machine Learning Research}, pp.\  2256--2265, Lille, France, 07--09 Jul
  2015. PMLR.
\newblock URL \url{https://proceedings.mlr.press/v37/sohl-dickstein15.html}.

\bibitem[Song et~al.(2021)Song, Meng, and Ermon]{song2021denoising}
Song, J., Meng, C., and Ermon, S.
\newblock Denoising diffusion implicit models.
\newblock In \emph{International Conference on Learning Representations}, 2021.
\newblock URL \url{https://openreview.net/forum?id=St1giarCHLP}.

\bibitem[Thudi et~al.(2022)Thudi, Deza, Chandrasekaran, and
  Papernot]{thudi2022unrolling}
Thudi, A., Deza, G., Chandrasekaran, V., and Papernot, N.
\newblock Unrolling sgd: Understanding factors influencing machine unlearning.
\newblock In \emph{2022 IEEE 7th European Symposium on Security and Privacy
  (EuroS\&P)}, pp.\  303--319. IEEE, 2022.

\bibitem[van~der Maaten \& Hinton(2008)van~der Maaten and
  Hinton]{JMLR:v9:vandermaaten08a}
van~der Maaten, L. and Hinton, G.
\newblock Visualizing data using t-sne.
\newblock \emph{Journal of Machine Learning Research}, 9\penalty0
  (86):\penalty0 2579--2605, 2008.
\newblock URL \url{http://jmlr.org/papers/v9/vandermaaten08a.html}.

\bibitem[Wang et~al.(2020)Wang, Yao, Kwok, and Ni]{10.1145/3386252}
Wang, Y., Yao, Q., Kwok, J.~T., and Ni, L.~M.
\newblock Generalizing from a few examples: A survey on few-shot learning.
\newblock \emph{ACM Comput. Surv.}, 53\penalty0 (3), jun 2020.
\newblock ISSN 0360-0300.
\newblock \doi{10.1145/3386252}.
\newblock URL \url{https://doi.org/10.1145/3386252}.

\bibitem[Wei et~al.(2023)Wei, Zhang, Ji, Bai, Zhang, and Zuo]{Wei_2023_ICCV}
Wei, Y., Zhang, Y., Ji, Z., Bai, J., Zhang, L., and Zuo, W.
\newblock Elite: Encoding visual concepts into textual embeddings for
  customized text-to-image generation.
\newblock In \emph{Proceedings of the IEEE/CVF International Conference on
  Computer Vision (ICCV)}, pp.\  15943--15953, October 2023.

\bibitem[Weiss et~al.(2022)Weiss, Rahaman, Locatello, Pal, Bengio,
  Sch{\"o}lkopf, Li, and Ballas]{weiss2022neural}
Weiss, M., Rahaman, N., Locatello, F., Pal, C., Bengio, Y., Sch{\"o}lkopf, B.,
  Li, L.~E., and Ballas, N.
\newblock Neural attentive circuits.
\newblock In Oh, A.~H., Agarwal, A., Belgrave, D., and Cho, K. (eds.),
  \emph{Advances in Neural Information Processing Systems}, 2022.
\newblock URL \url{https://openreview.net/forum?id=q41xK9Bunq1}.

\bibitem[Wightman(2019)]{rw2019timm}
Wightman, R.
\newblock Pytorch image models.
\newblock \url{https://github.com/rwightman/pytorch-image-models}, 2019.

\bibitem[Xue et~al.(2023)Xue, Song, Guo, Liu, Zong, Liu, and
  Luo]{xue2023raphael}
Xue, Z., Song, G., Guo, Q., Liu, B., Zong, Z., Liu, Y., and Luo, P.
\newblock {RAPHAEL}: Text-to-image generation via large mixture of diffusion
  paths.
\newblock In \emph{Thirty-seventh Conference on Neural Information Processing
  Systems}, 2023.
\newblock URL \url{https://openreview.net/forum?id=jUdZCcoOu3}.

\bibitem[Yao et~al.(2022)Yao, Huang, Dong, Wei, Chen, and
  Zhang]{yao2022kformer}
Yao, Y., Huang, S., Dong, L., Wei, F., Chen, H., and Zhang, N.
\newblock Kformer: Knowledge injection in transformer feed-forward layers.
\newblock In \emph{CCF International Conference on Natural Language Processing
  and Chinese Computing}, pp.\  131--143. Springer, 2022.

\bibitem[Zhang et~al.(2024{\natexlab{a}})Zhang, Wang, Xu, Wang, and
  Shi]{Zhang_2024_CVPR}
Zhang, G., Wang, K., Xu, X., Wang, Z., and Shi, H.
\newblock Forget-me-not: Learning to forget in text-to-image diffusion models.
\newblock In \emph{Proceedings of the IEEE/CVF Conference on Computer Vision
  and Pattern Recognition (CVPR) Workshops}, pp.\  1755--1764, June
  2024{\natexlab{a}}.

\bibitem[Zhang \& Chen(2023)Zhang and Chen]{zhang2023fast}
Zhang, Q. and Chen, Y.
\newblock Fast sampling of diffusion models with exponential integrator.
\newblock In \emph{The Eleventh International Conference on Learning
  Representations}, 2023.
\newblock URL \url{https://openreview.net/forum?id=Loek7hfb46P}.

\bibitem[Zhang et~al.(2018)Zhang, Isola, Efros, Shechtman, and
  Wang]{Zhang_2018_CVPR}
Zhang, R., Isola, P., Efros, A.~A., Shechtman, E., and Wang, O.
\newblock The unreasonable effectiveness of deep features as a perceptual
  metric.
\newblock In \emph{Proceedings of the IEEE Conference on Computer Vision and
  Pattern Recognition (CVPR)}, June 2018.

\bibitem[Zhang et~al.(2024{\natexlab{b}})Zhang, Wei, Zhang, Wu, Zhang, Lei, and
  Li]{zhang2024survey}
Zhang, X., Wei, X.-Y., Zhang, W., Wu, J., Zhang, Z., Lei, Z., and Li, Q.
\newblock A survey on personalized content synthesis with diffusion models,
  2024{\natexlab{b}}.

\bibitem[Zhao et~al.(2024)Zhao, Zhang, Zheng, Kong, and Yin]{zhao2024separable}
Zhao, M., Zhang, L., Zheng, T., Kong, Y., and Yin, B.
\newblock Separable multi-concept erasure from diffusion models, 2024.

\end{thebibliography}
\bibliographystyle{icml2024}

\newpage
\appendix
\onecolumn

\section{Preliminaries}
\subsection{Diffusion Models}
Denoising diffusion models consist of two processes: forward and reverse process. In the forward process, the noise is gradually added to the input data $\boldsymbol{x}_0$, eventually resulting in pure Gaussian noise. In the reverse process, starting from Gaussian noise, the model predicts the noise added at each time step $t\in[0, T]$.

\subsection{Anchor Concepts}
\label{appendix:anchor-concepts}
We refer to concept $B$ as the anchor concept when transitioning from concept $A$ to concept $B$. For example, with UCE~\cite{Gandikota_2024_WACV}, the following optimization problem is formulated.

\begin{equation*}
\min_{W}\sum_{i=0}^m\|Wc_i-\underbrace{v_i^*}_{W^{\mathrm{old}}c_i^*}\|_2^2+\lambda\|W-W^{\mathrm{old}}\|_F^2
\end{equation*}

Here, $W$ is the projection matrix, $c_i$ is the text embedding of the prompt containing the concepts to be erased (e.g. ``Van Gogh style''), and $c_i^*$ is also text embeddings taken from the destination prompt (e.g. ``art''). In this case, we consider $c_i^*$ to be the anchor concept.

As another example, we consider Ablating Concepts~\cite{Kumari_2023_ICCV}. The method is formulated as a model-based approach as follows.

\begin{equation*}
\arg\min_{\varepsilon_{\theta}}\mathbb{E}_{\boldsymbol{x}_t, \boldsymbol{c}, \boldsymbol{c}^*, t}[w_t\|\varepsilon_{\theta_{\mathrm{fixed}}}(\boldsymbol{x}_t, \boldsymbol{c}, t)-\varepsilon_{\theta}(\boldsymbol{x}_t, \boldsymbol{c}^*, t)\|_2^2]
\end{equation*}

where $\boldsymbol{c}$ is a random prompt for the anchor concept (e.g. ``cat'') and $\boldsymbol{c}^*$ is modified from $\boldsymbol{c}$ to include the target concept (e.g. ``Grumpy Cat''). In this case, $\boldsymbol{c}^*$ is the anchor concept.

\subsection{Retrain Model}
\label{appendix:retrain}
We refer to a ``retrain model'' as one that is trained on a training dataset from which the forgetting dataset has been removed. Specifically, let $\mathcal{D}$ denote the training datasets and $\mathcal{D}_f\subseteq\mathcal{D}$ is the forgetting datasets. As an example, we assume a text-to-image model has been trained on LAION-5B and want to erase the concept of ``Grumpy Cat''. In this case, LAION-5B is $\mathcal{D}$ and the subset of LAION-5B, the elements of which contain ``Grumpy Cat'', is $\mathcal{D}_f$. The $\mathcal{D}_r=\mathcal{D}\setminus\mathcal{D}_f$, which is $\mathcal{D}_f$ removed from $\mathcal{D}$, denote the remaining datasets, and the model trained using $\mathcal{D}_r$ is called a retrain model.

\section{Updated Layers}
\label{appendix:updated-parameters}
\cref{table:updated-parameters} shows the names of the updated layers.

\begin{table}[htbp]
\caption{List of the updated layer name with our method.}
\label{table:updated-parameters}
\centering
\begin{small}
\begin{tabular}{c}
\toprule
Layer Name \\
\midrule
text\_model.encoder.layers.0.mlp.fc1 \\
text\_model.encoder.layers.0.mlp.fc2 \\
text\_model.encoder.layers.1.mlp.fc1 \\
text\_model.encoder.layers.1.mlp.fc2 \\
text\_model.encoder.layers.2.mlp.fc1 \\
text\_model.encoder.layers.2.mlp.fc2 \\
text\_model.encoder.layers.3.mlp.fc1 \\
text\_model.encoder.layers.3.mlp.fc2 \\
text\_model.encoder.layers.4.mlp.fc1 \\
text\_model.encoder.layers.4.mlp.fc2 \\
text\_model.encoder.layers.5.mlp.fc1 \\
text\_model.encoder.layers.5.mlp.fc2 \\
text\_model.encoder.layers.6.mlp.fc1 \\
text\_model.encoder.layers.6.mlp.fc2 \\
text\_model.encoder.layers.7.mlp.fc1 \\
text\_model.encoder.layers.7.mlp.fc2 \\
text\_model.encoder.layers.8.mlp.fc1 \\
text\_model.encoder.layers.8.mlp.fc2 \\
text\_model.encoder.layers.9.mlp.fc1 \\
text\_model.encoder.layers.9.mlp.fc2 \\
text\_model.encoder.layers.10.mlp.fc1 \\
text\_model.encoder.layers.10.mlp.fc2 \\
text\_model.encoder.layers.11.mlp.fc1 \\
text\_model.encoder.layers.11.mlp.fc2 \\
text\_model.encoder.layers.11.self\_attn.k\_proj \\
text\_model.encoder.layers.11.self\_attn.v\_proj \\
text\_model.encoder.layers.11.self\_attn.k\_proj \\
text\_model.encoder.layers.11.self\_attn.out\_proj \\
\bottomrule
\end{tabular}
\end{small}
\end{table}

\section{Experimental Details}
\subsection{Environments}
We conducted all experiments on a single NVIDIA RTX A5000. Our software environments were PyTorch 1.13.1, diffusers 0.21.4, and transformers 4.34.0. 

\subsection{Baselines}
\label{appendix-subsec:baselines}
We describe the baselines used in our experiments.

\begin{itemize}
\item ESD~\cite{Gandikota_2023_ICCV}: This method updates the parameters related to U-Net. It prepares two outputs: one from Original SD and the other from the fine-tuned model. The fine-tuned model is updated using the L2 Loss of each output. We use ESD-x-1, which updates the parameters related to cross-attention.
\item UCE~\cite{Gandikota_2024_WACV}: This method updates the weights of the cross-attention in U-Net by closed-form. Unlike ESD, it does not require the output of the Original SD. In the experiments, we used anchor concepts generated by ChatGPT.
\item SPM~\cite{Lyu_2024_CVPR}: This method uses adapter tuning using LoRA~\cite{hu2022lora} for the U-Net attention module. Therefore, transferring erased model to another model is easy.
\end{itemize}
  
\subsection{Implementation Details of Baselines}
We describe the details of the baseline implementations.

\begin{itemize}
\item \textbf{ESD-x}: We used the original implementation using diffusers\footnote{\url{https://huggingface.co/spaces/baulab/Erasing-Concepts-In-Diffusion/tree/main}}. Following the original paper, the learning rate was set to $10^{-5}$, number of iterations was 1,000, and $\eta=1$.
\item \textbf{UCE}: We used the original implementation\footnote{\url{https://github.com/rohitgandikota/unified-concept-editing/tree/main}}. The hyperparameters, except anchor concept, were those provided in the official implementation. The anchor concept was generated by ChatGPT~\footnote{\url{https://chat.openai.com/}} for each concept. The prompt given to ChatGPT was \texttt{when the erased concepts is "\{TARGET CONCEPT\}", what concepts to guide the erased concepts towards? Answer the one concept name.} 
\item \textbf{SPM}: We used the original implementation\footnote{\url{https://github.com/Con6924/SPM/tree/main}}. For configuration of training, we used the provided configuration\footnote{\url{https://github.com/Con6924/SPM/blob/main/configs/snoopy/config.yaml}}. When generating the images, we did not use the negative prompt for fair comparison with other methods. The setting of the generation phase is described in \cref{subsec:exp-setting}.
\end{itemize}

\subsection{Implementation Details of Our Proposed Method}
We used CLIP ImageNet Template to erase the concepts. This template is separated into object and style. We list the details of the templates in Tables~\ref{tab:template-object} and \ref{tab:template-style}. These prompts are also used in textual inversion for diffusers\footnote{\url{https://github.com/huggingface/diffusers/blob/main/examples/textual_inversion/textual_inversion.py}}. When updating the text encoder, prompts are randomly chosen for each iteration.

\begin{table}[htbp]
\begin{minipage}[t]{.45\textwidth}
    \caption{List of prompts used when erasing object}
    \label{tab:template-object}
    \centering
    \begin{small}
    \begin{tabular}{c}
    \toprule
    Object Prompt  \\
    \midrule
    a photo of a \{\textit{Object Name}\} \\
    a rendering of a \{\textit{Object Name}\} \\
    a cropped photo of the \{\textit{Object Name}\} \\
    the photo of a \{\textit{Object Name}\} \\
    a photo of a clean \{\textit{Object Name}\} \\
    a photo of a dirty \{\textit{Object Name}\} \\
    a dark photo of the \{\textit{Object Name}\} \\
    a photo of my \{\textit{Object Name}\} \\
    a photo of the cool \{\textit{Object Name}\} \\
    a close-up photo of a \{\textit{Object Name}\} \\
    a bright photo of the \{\textit{Object Name}\} \\
    a cropped photo of a \{\textit{Object Name}\} \\
    a photo of the \{\textit{Object Name}\} \\
    a good photo of the \{\textit{Object Name}\} \\
    a photo of one \{\textit{Object Name}\} \\
    a close-up photo of the \{\textit{Object Name}\} \\
    a rendition of the \{\textit{Object Name}\} \\
    a photo of the clean \{\textit{Object Name}\} \\
    a rendition of a \{\textit{Object Name}\} \\
    a photo of a nice \{\textit{Object Name}\} \\
    a good photo of a \{\textit{Object Name}\} \\
    a photo of the nice \{\textit{Object Name}\} \\
    a photo of the small \{\textit{Object Name}\} \\
    a photo of the weird \{\textit{Object Name}\} \\
    a photo of the large \{\textit{Object Name}\} \\
    a photo of a cool \{\textit{Object Name}\} \\
    a photo of a small \{\textit{Object Name}\} \\
    \bottomrule
    \end{tabular}
    \end{small}
\end{minipage}
\hfil
\begin{minipage}[t]{.45\textwidth}
    \caption{List of prompts used when erasing style}
    \label{tab:template-style}
    \centering
    \begin{small}
    \begin{tabular}{c}
    \toprule
    Style Prompt  \\
    \midrule
    a painting in the style of \{\textit{Style Name}\} \\
    a rendering in the style of \{\textit{Style Name}\} \\
    a cropped painting in the style of \{\textit{Style Name}\} \\
    the painting in the style of \{\textit{Style Name}\} \\
    a clean painting in the style of \{\textit{Style Name}\} \\
    a dirty painting in the style of \{\textit{Style Name}\} \\
    a dark painting in the style of \{\textit{Style Name}\} \\
    a picture in the style of \{\textit{Style Name}\} \\
    a cool painting in the style of \{\textit{Style Name}\} \\
    a close-up painting in the style of \{\textit{Style Name}\} \\
    a bright painting in the style of \{\textit{Style Name}\} \\
    a cropped painting in the style of \{\textit{Style Name}\} \\
    a good painting in the style of \{\textit{Style Name}\} \\
    a close-up painting in the style of \{\textit{Style Name}\} \\
    a rendition in the style of \{\textit{Style Name}\} \\
    a nice painting in the style of \{\textit{Style Name}\} \\
    a small painting in the style of \{\textit{Style Name}\} \\
    a weird painting in the style of \{\textit{Style Name}\} \\
    a large painting in the style of \{\textit{Style Name}\} \\
    \bottomrule
    \end{tabular}
    \end{small}
\end{minipage}
\end{table}

\section{Additional Results}
\label{supl:additional-results}
In this section, we present the results of further qualitative evaluation. Comparisons with current methods and further results of the proposed method only are presented.

\subsection{Additional Comparison}
\cref{tab:add-exp} shows the target concepts, the anchor concepts used with UCE, and the prompts used during generation.

\begin{table}[htbp]
\caption{List of target concepts, anchor concepts, and prompts in generating}
\label{tab:add-exp}
\centering
\begin{tabular}{lcc}
\toprule
Target Concept & Anchor Concept (UCE) & Prompt in Generating \\
\midrule
Snoopy & Peanuts & Snoopy in cyberpunk style \\
Grumpy Cat & Internet Meme & A Grumpy cat laying in the sun. \\
Mickey Mouse & Disney & drawing of Mickey Mouse walking along the river. \\
R2D2 & Star Wars & portrait of R2D2 \\
Gogh Style & Post-Impressionism & Painting of trees in bloom in the style of Van Gogh. \\
\bottomrule
\end{tabular}
\end{table}

\begin{figure}[!htbp]
\begin{minipage}[b]{0.19\linewidth}
\centering
\includegraphics[width=0.8\linewidth]{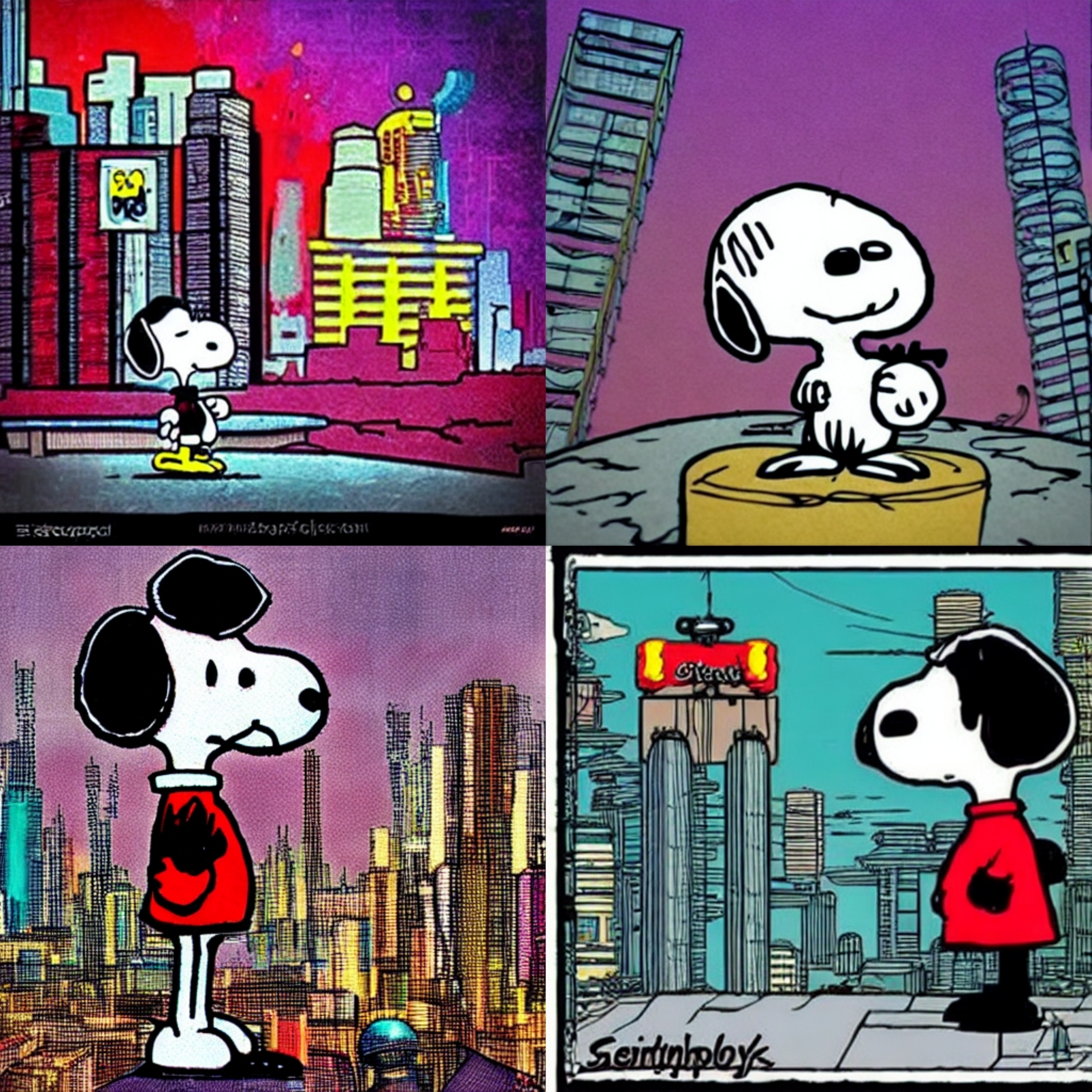}
\subcaption{Original SD}
\end{minipage}
\begin{minipage}[b]{0.19\linewidth}
\centering
\includegraphics[width=0.8\linewidth]{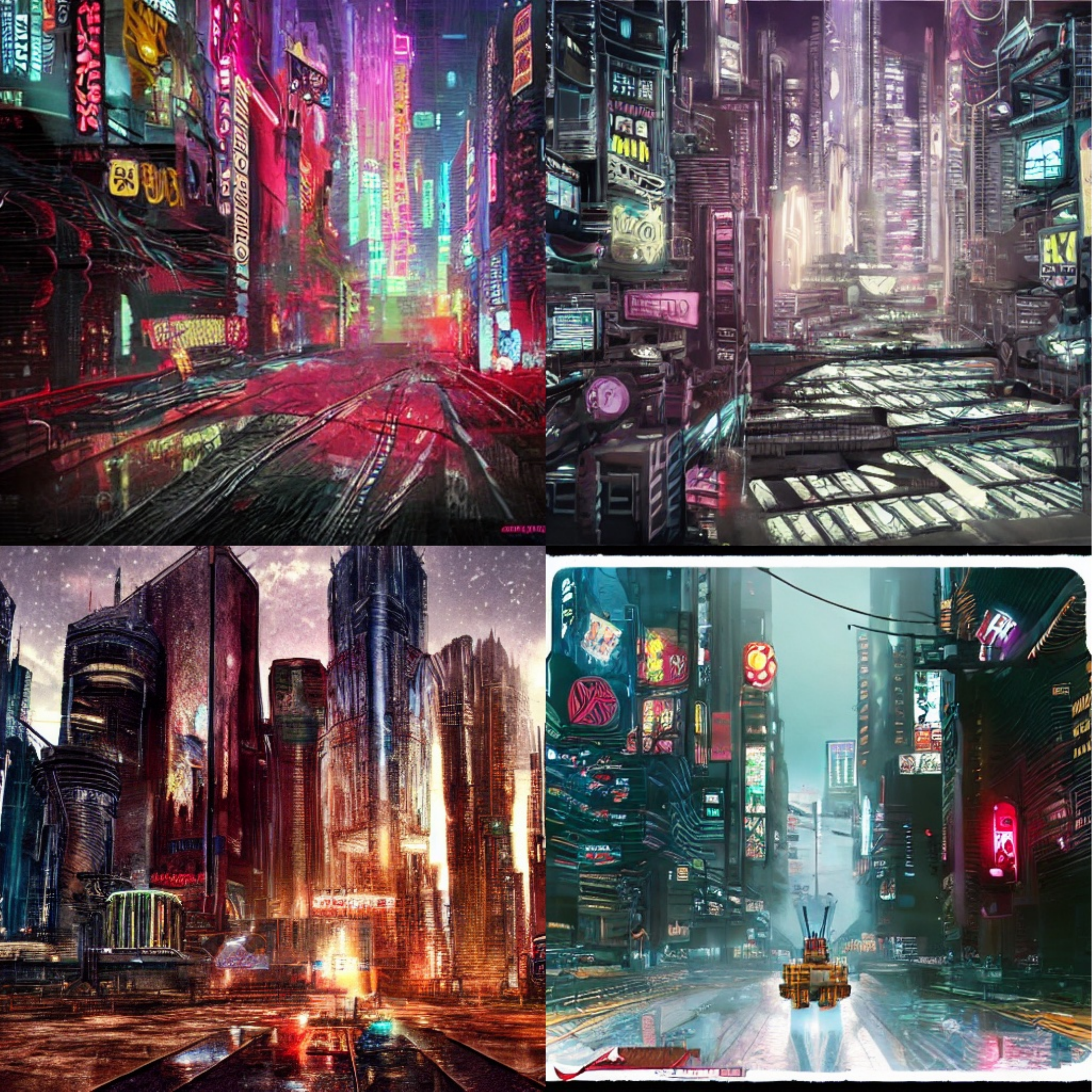}
\subcaption{ESD-x-1}
\end{minipage}
\begin{minipage}[b]{0.19\linewidth}
\centering
\includegraphics[width=0.8\linewidth]{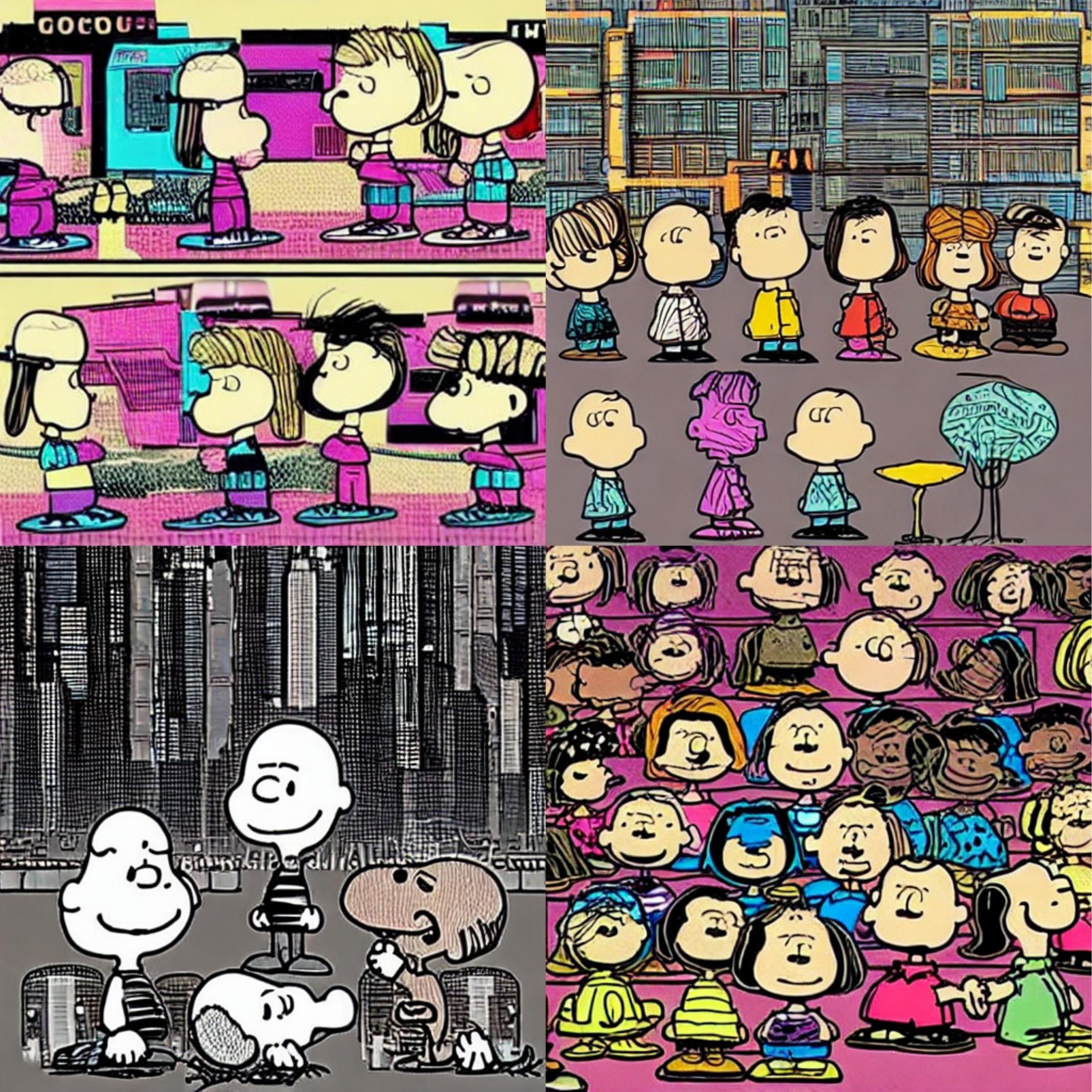}
\subcaption{UCE}
\end{minipage}
\begin{minipage}[b]{0.19\linewidth}
\centering
\includegraphics[width=0.8\linewidth]{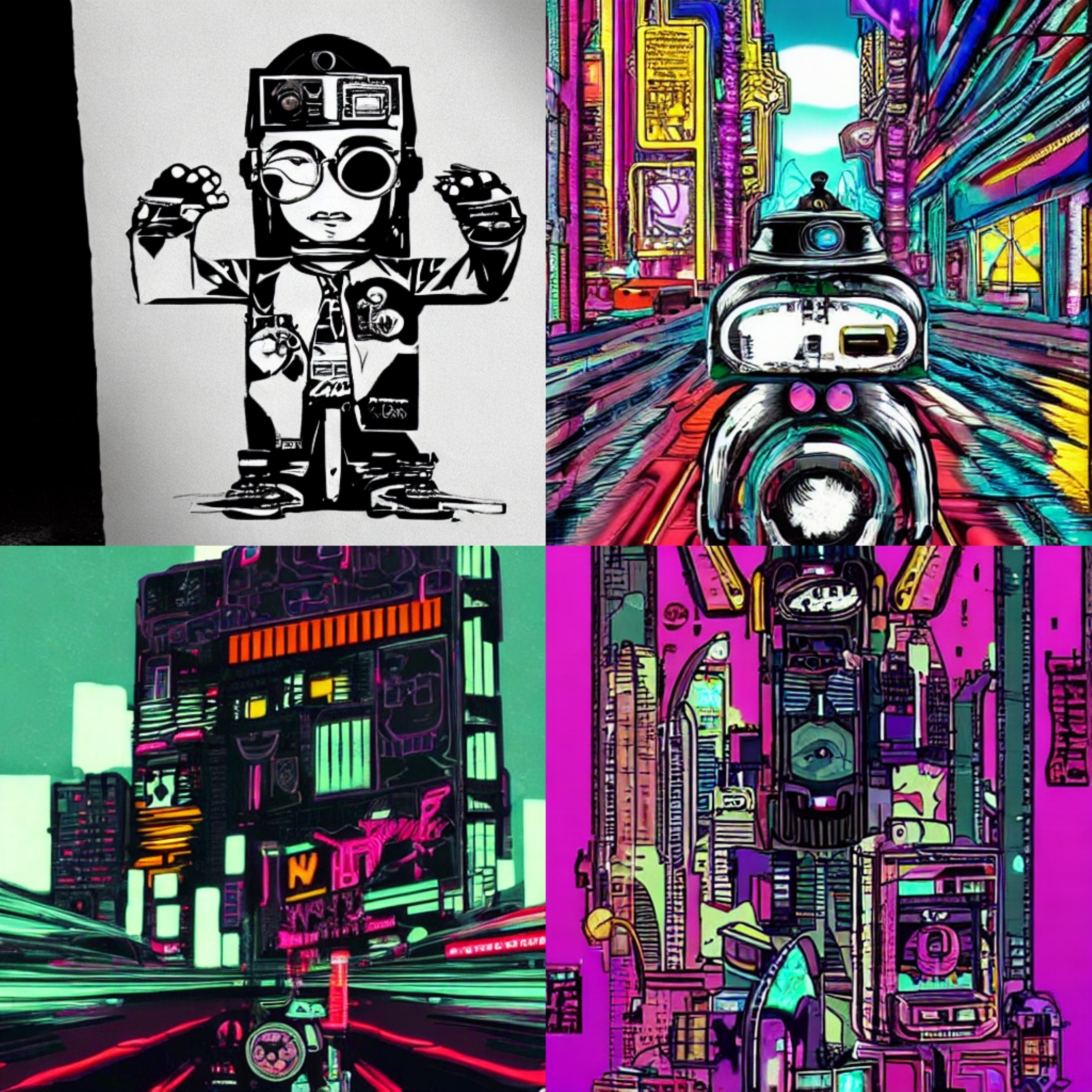}
\subcaption{SPM}
\end{minipage}
\begin{minipage}[b]{0.19\linewidth}
\centering
\includegraphics[width=0.8\linewidth]{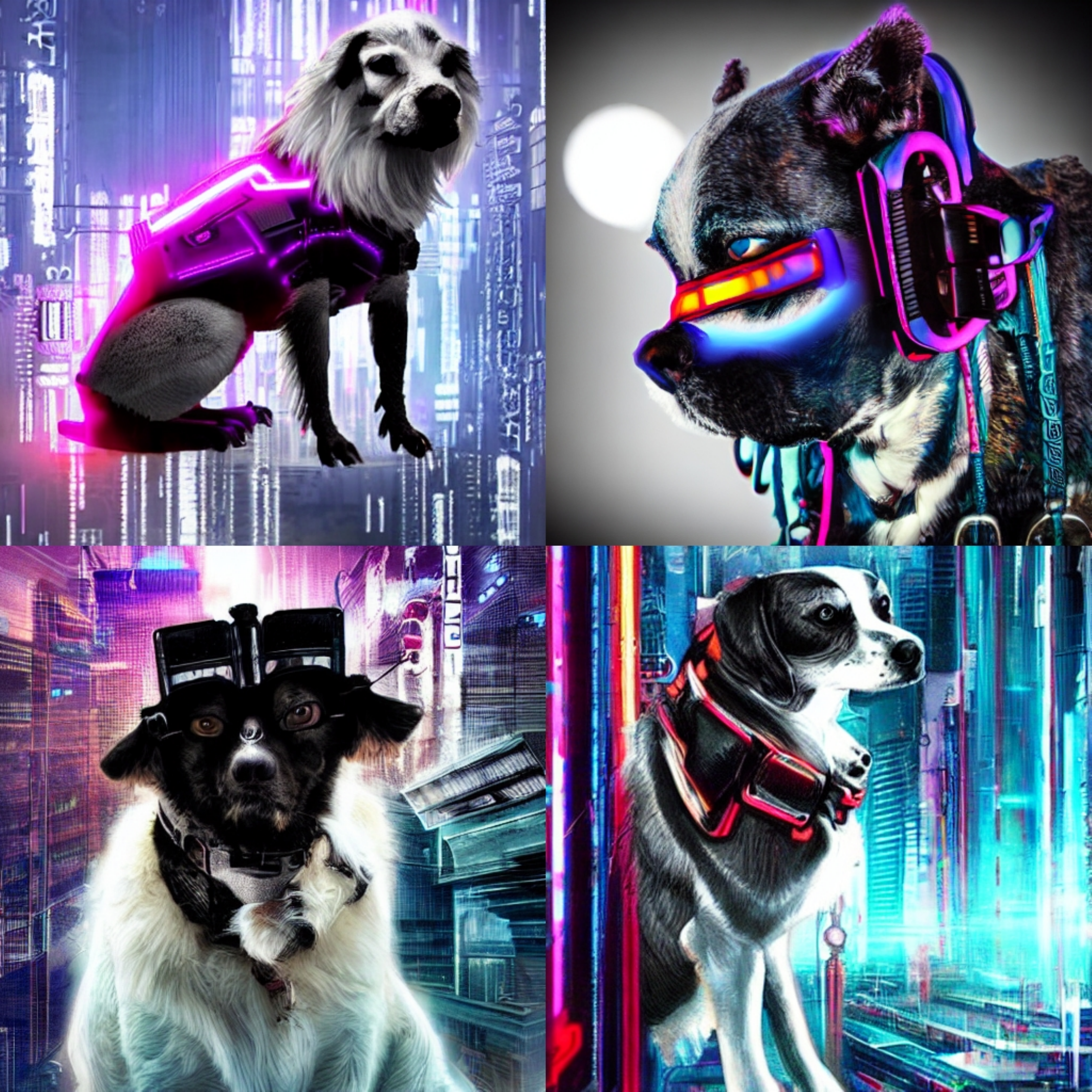}
\subcaption{Ours}
\end{minipage}   
\caption{Comparison of generated images when erasing ``Snoopy''}
\label{fig:erasing-snoopy}
\end{figure}

\cref{fig:erasing-snoopy} shows the results of erasing ``Snoopy''. Since the motif of Snoopy is a dog, it can be considered that the proposed method has transitioned to a concept close to it. Additionally, ESD-x-1 resulted in the absence of Snoopy, which is speculated to differ from the retrain results we consider as ground truth. Although the results of UCE were mapped by the concept ``Peanuts'', it is difficult to argue that the ``cyberpunk style'' was adequately reflected.

\begin{figure}[!htbp]
\begin{minipage}[b]{0.19\linewidth}
\centering
\includegraphics[width=0.8\linewidth]{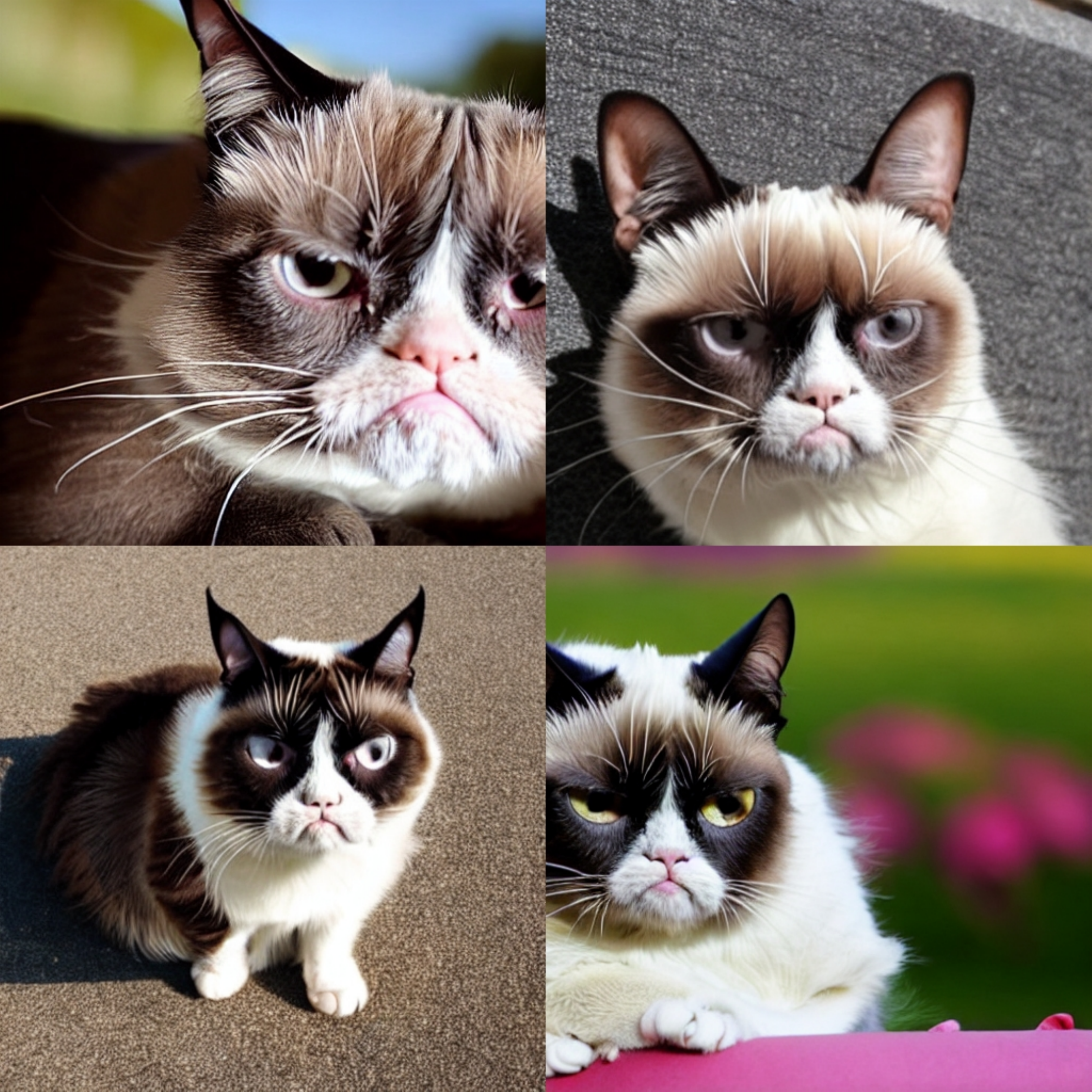}
\subcaption{Original SD}
\end{minipage}
\begin{minipage}[b]{0.19\linewidth}
\centering
\includegraphics[width=0.8\linewidth]{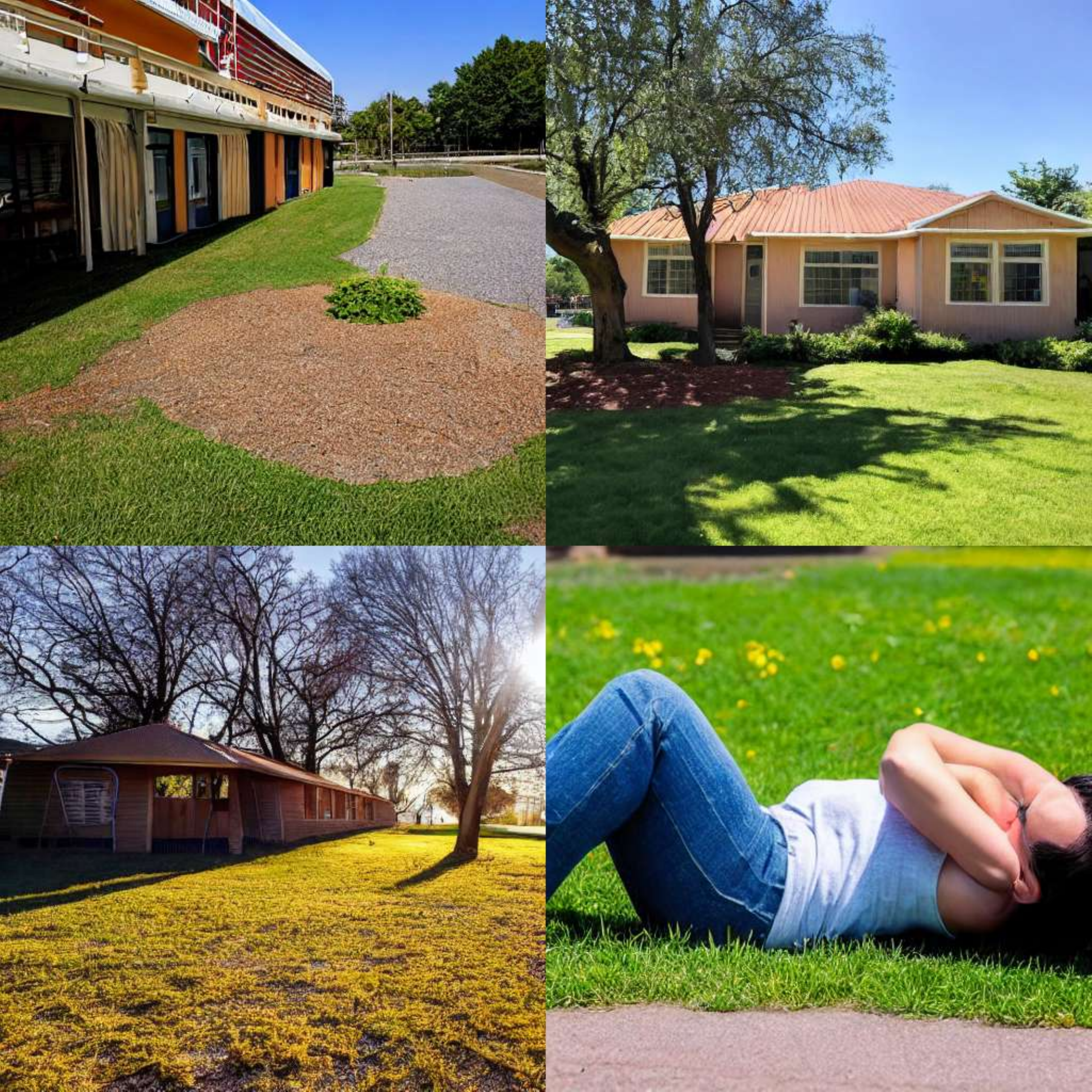}
\subcaption{ESD-x-1}
\end{minipage}
\begin{minipage}[b]{0.19\linewidth}
\centering
\includegraphics[width=0.8\linewidth]{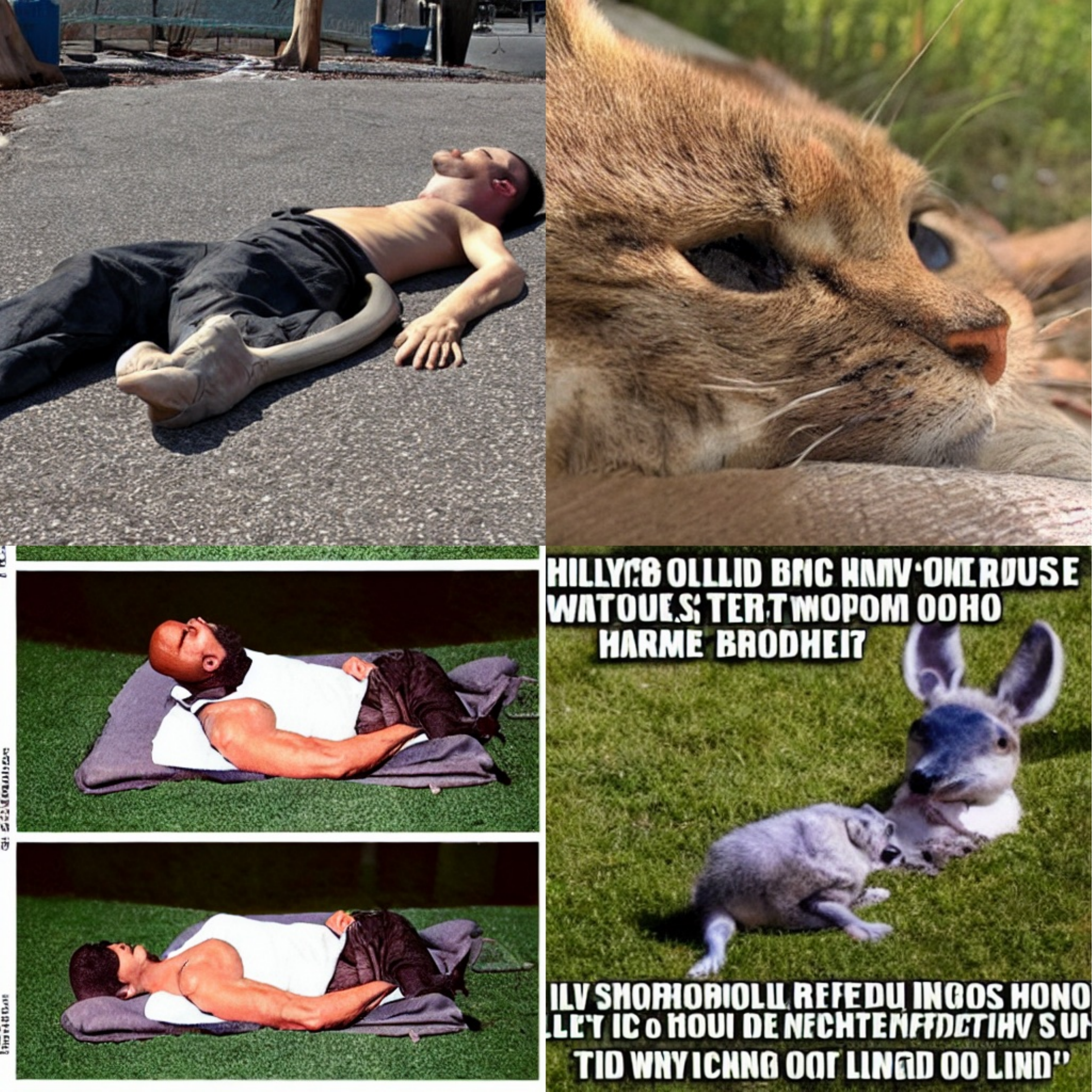}
\subcaption{UCE}
\end{minipage}
\begin{minipage}[b]{0.19\linewidth}
\centering
\includegraphics[width=0.8\linewidth]{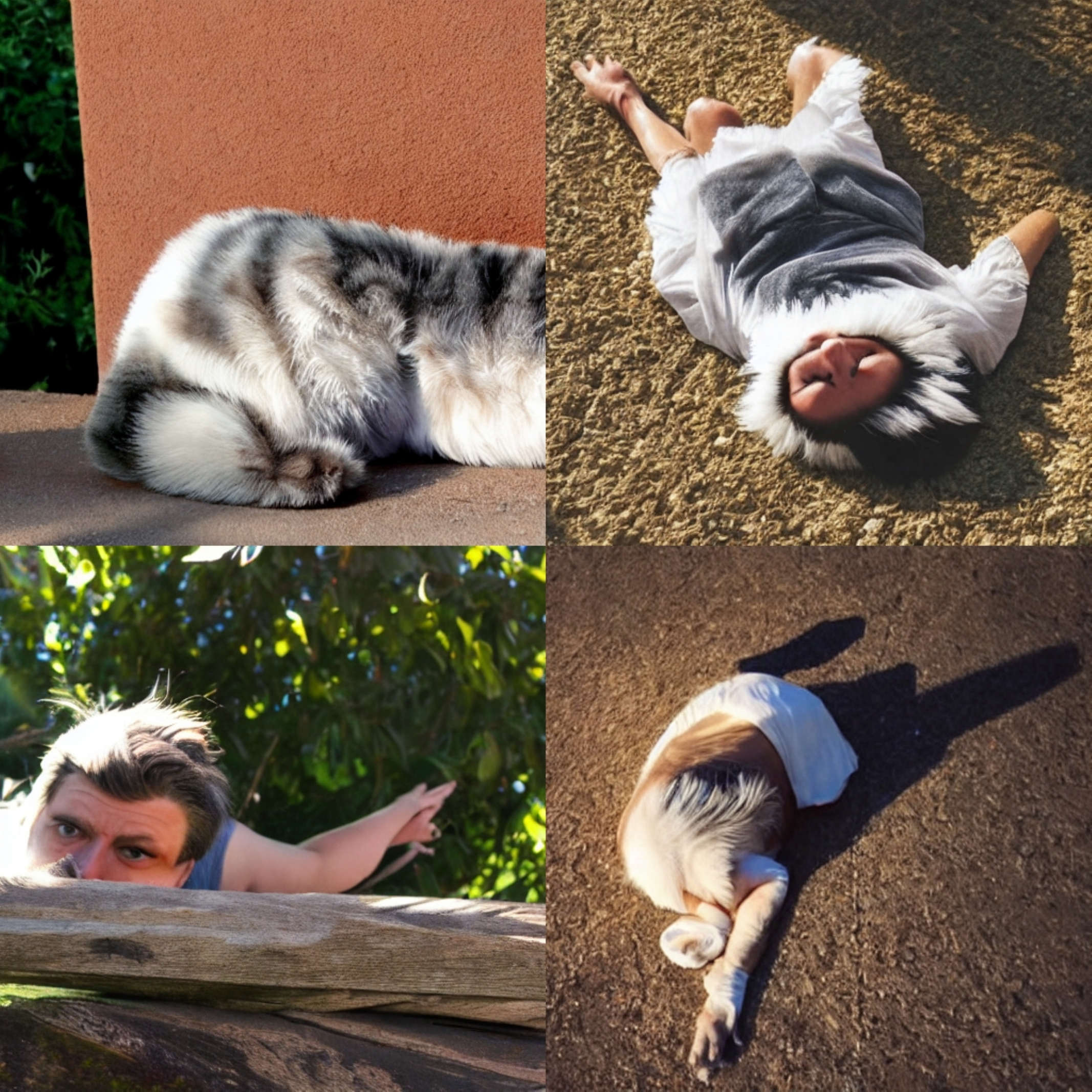}
\subcaption{SPM}
\end{minipage}
\begin{minipage}[b]{0.19\linewidth}
\centering
\includegraphics[width=0.8\linewidth]{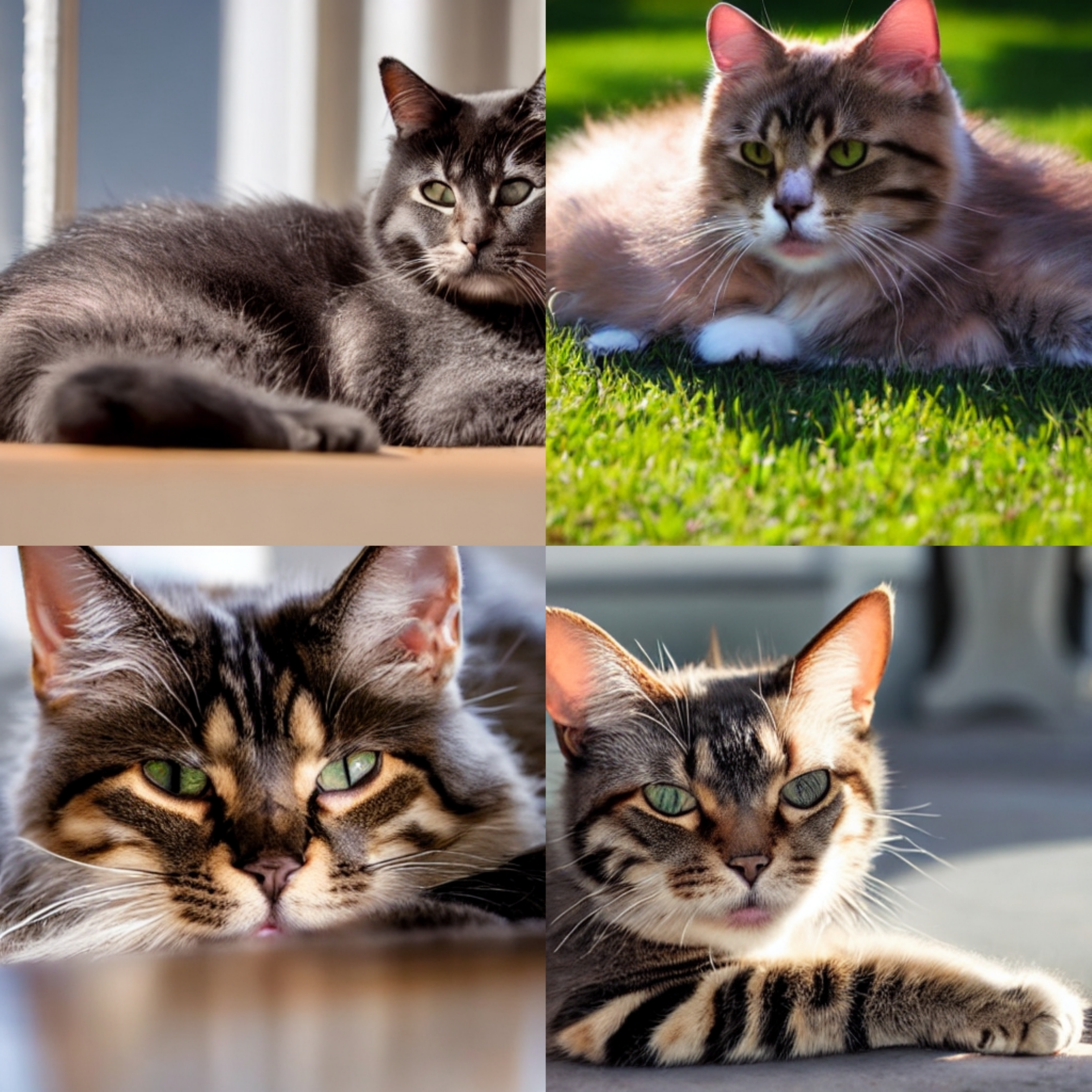}
\subcaption{Ours}
\end{minipage}   
\caption{Comparison of generated images when erasing ``Grumpy Cat''}
\label{fig:erasing-grumpy}
\end{figure}

\cref{fig:erasing-grumpy} shows the results of erasing ``Grumpy Cat''. ESD-x-1 generated an image of an unrelated building. Both SPM and UCE showed a significant decrease in generation quality. As shown in \cref{fig:comparison}, UCE maintained the quality of concept erasure and target concept generation when given an appropriate anchor concept. However, in this case, since ``Internet Meme'' is the anchor concept, it resulted in such a generation. ``Internet Meme'' was an anchor concept generated by ChatGPT, but whether such anchor concepts are appropriate depends on the knowledge of the human or model used to generate them.

\begin{figure}[!htbp]
\begin{minipage}[b]{0.19\linewidth}
\centering
\includegraphics[width=0.8\linewidth]{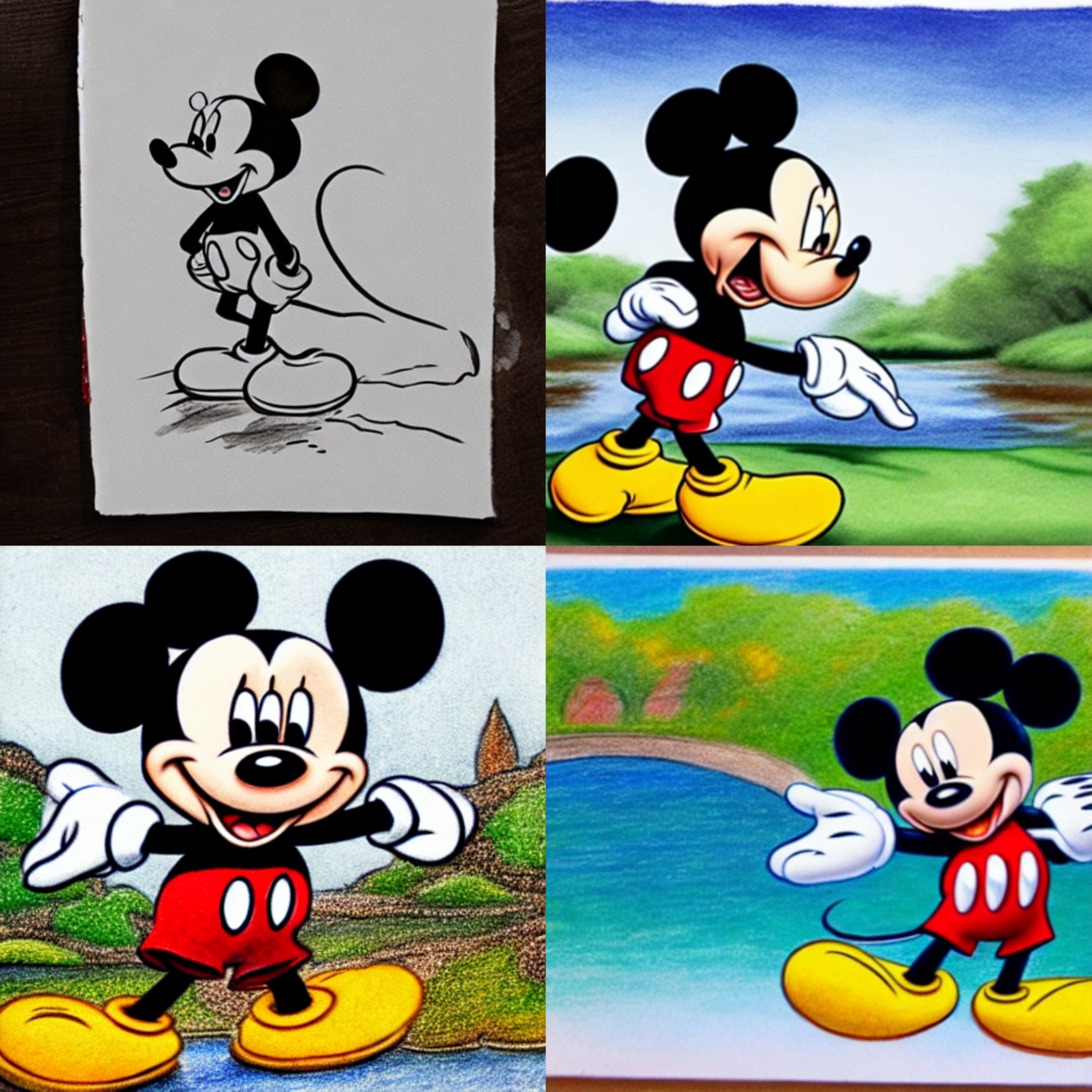}
\subcaption{Original SD}
\end{minipage}
\begin{minipage}[b]{0.19\linewidth}
\centering
\includegraphics[width=0.8\linewidth]{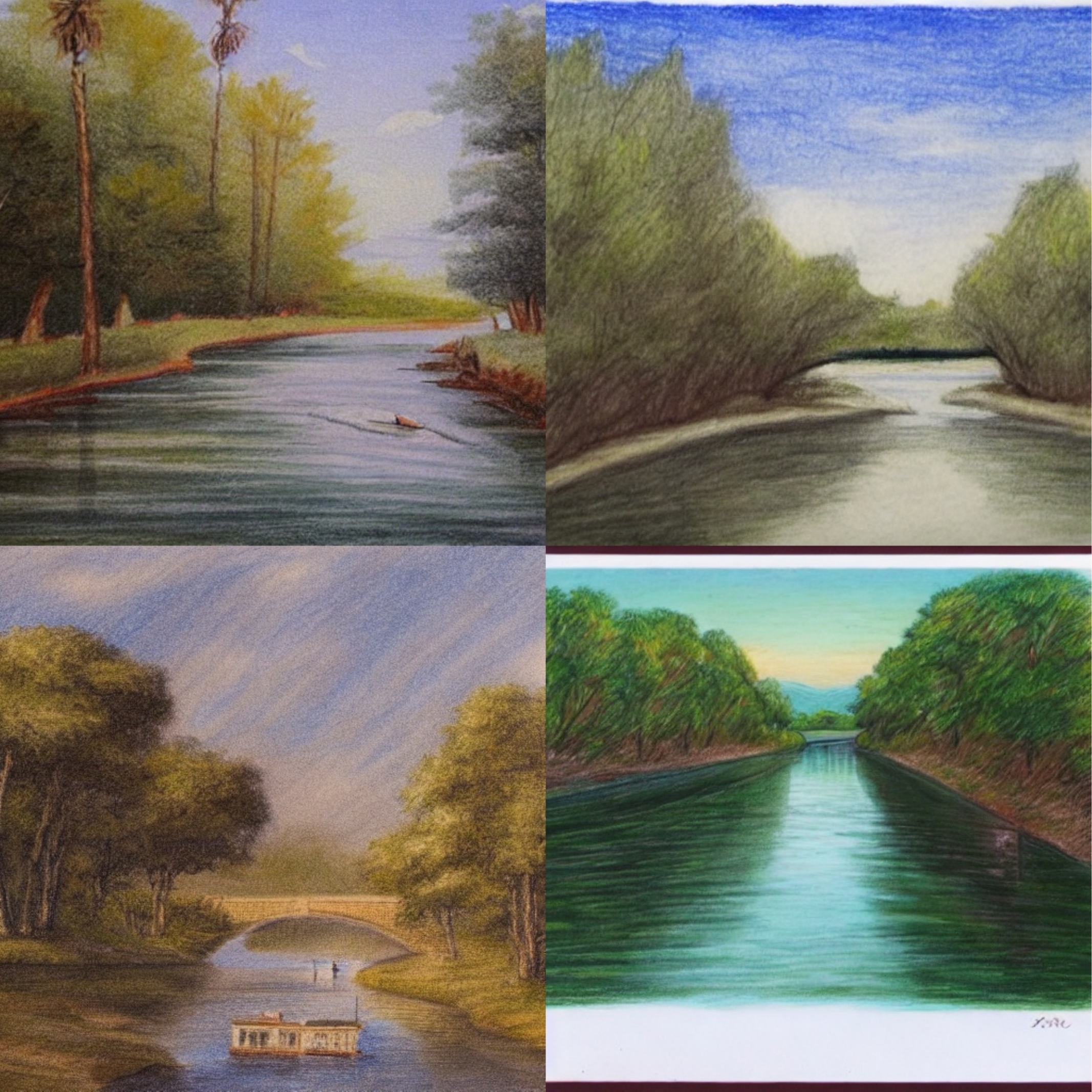}
\subcaption{ESD-x-1}
\end{minipage}
\begin{minipage}[b]{0.19\linewidth}
\centering
\includegraphics[width=0.8\linewidth]{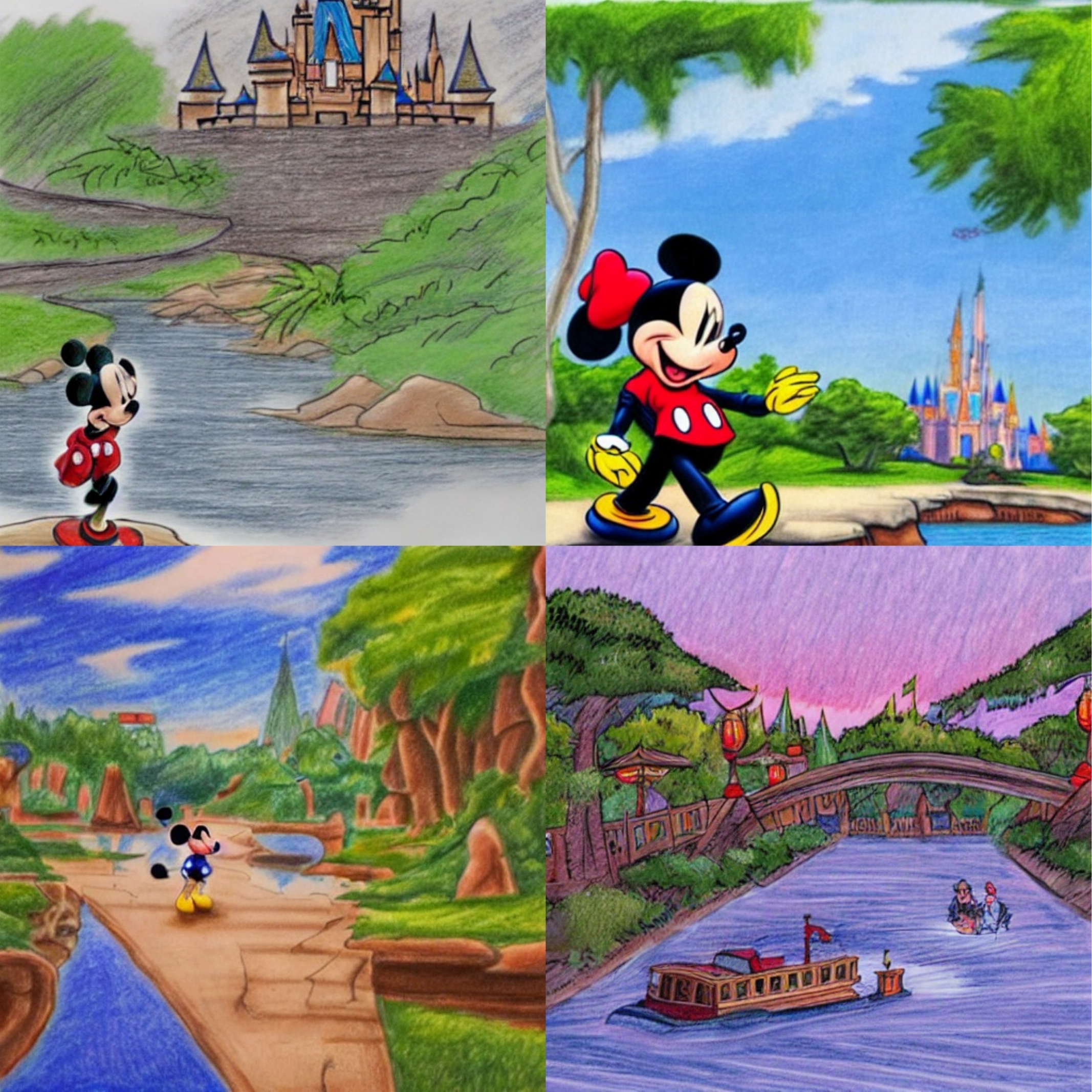}
\subcaption{UCE}
\end{minipage}
\begin{minipage}[b]{0.19\linewidth}
\centering
\includegraphics[width=0.8\linewidth]{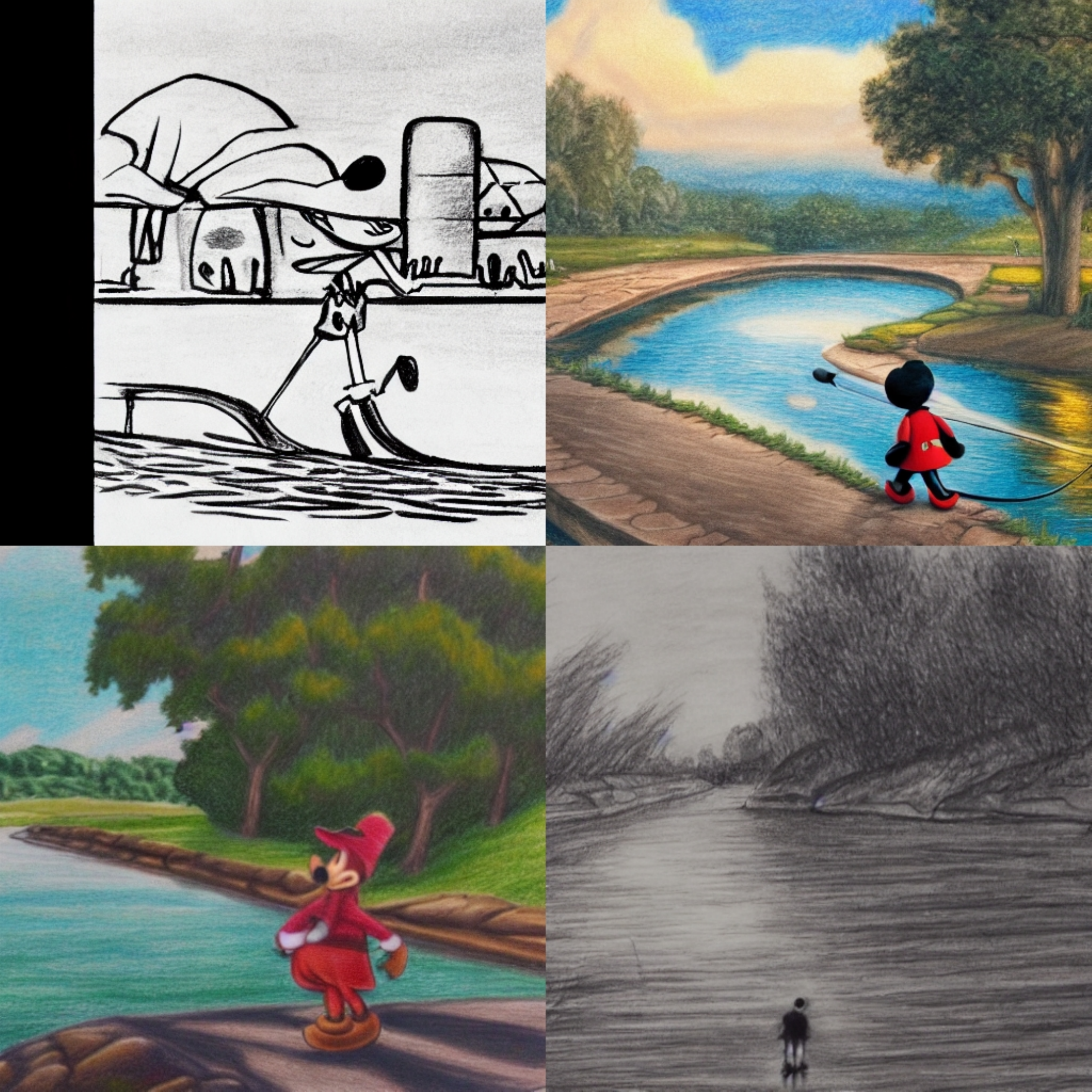}
\subcaption{SPM}
\end{minipage}
\begin{minipage}[b]{0.19\linewidth}
\centering
\includegraphics[width=0.8\linewidth]{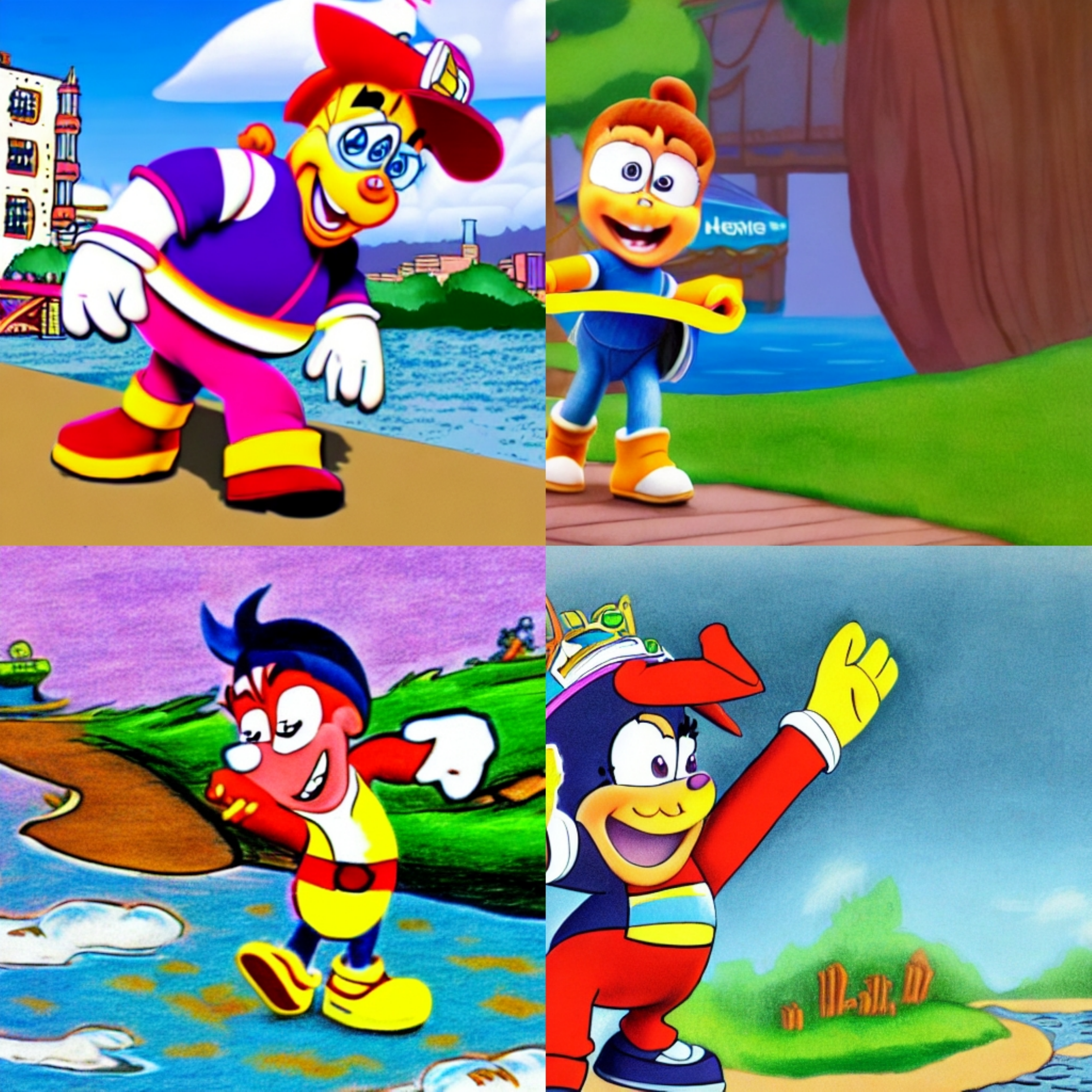}
\subcaption{Ours}
\end{minipage}   
\caption{Comparison of generated images when erasing ``Mickey Mouse''}
\label{fig:erasing-mickey}
\end{figure}

\cref{fig:erasing-mickey} presents the results of erasing ``Mickey Mouse''. ESD-x-1 seemed to disregard the target concept, resulting in images more aligned with "drawing" and "river" present in the prompt. Intuitively, something other than Mickey Mouse should be ``walking along the river''. Additionally, both SPM and UCE occasionally produced images reminiscent of ``Mickey Mouse''.

\begin{figure}[!htbp]
\begin{minipage}[b]{0.19\linewidth}
\centering
\includegraphics[width=0.8\linewidth]{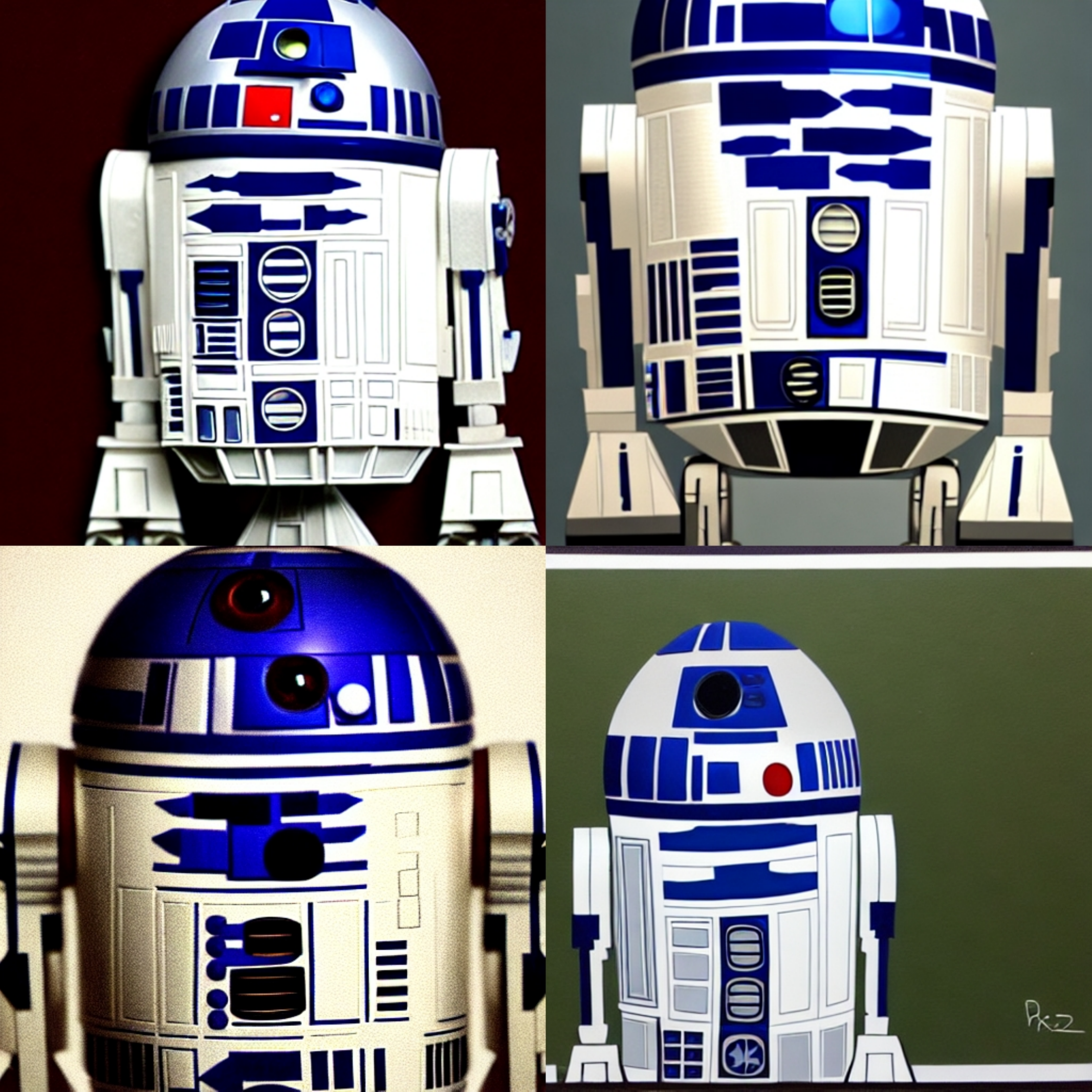}
\subcaption{Original SD}
\end{minipage}
\begin{minipage}[b]{0.19\linewidth}
\centering
\includegraphics[width=0.8\linewidth]{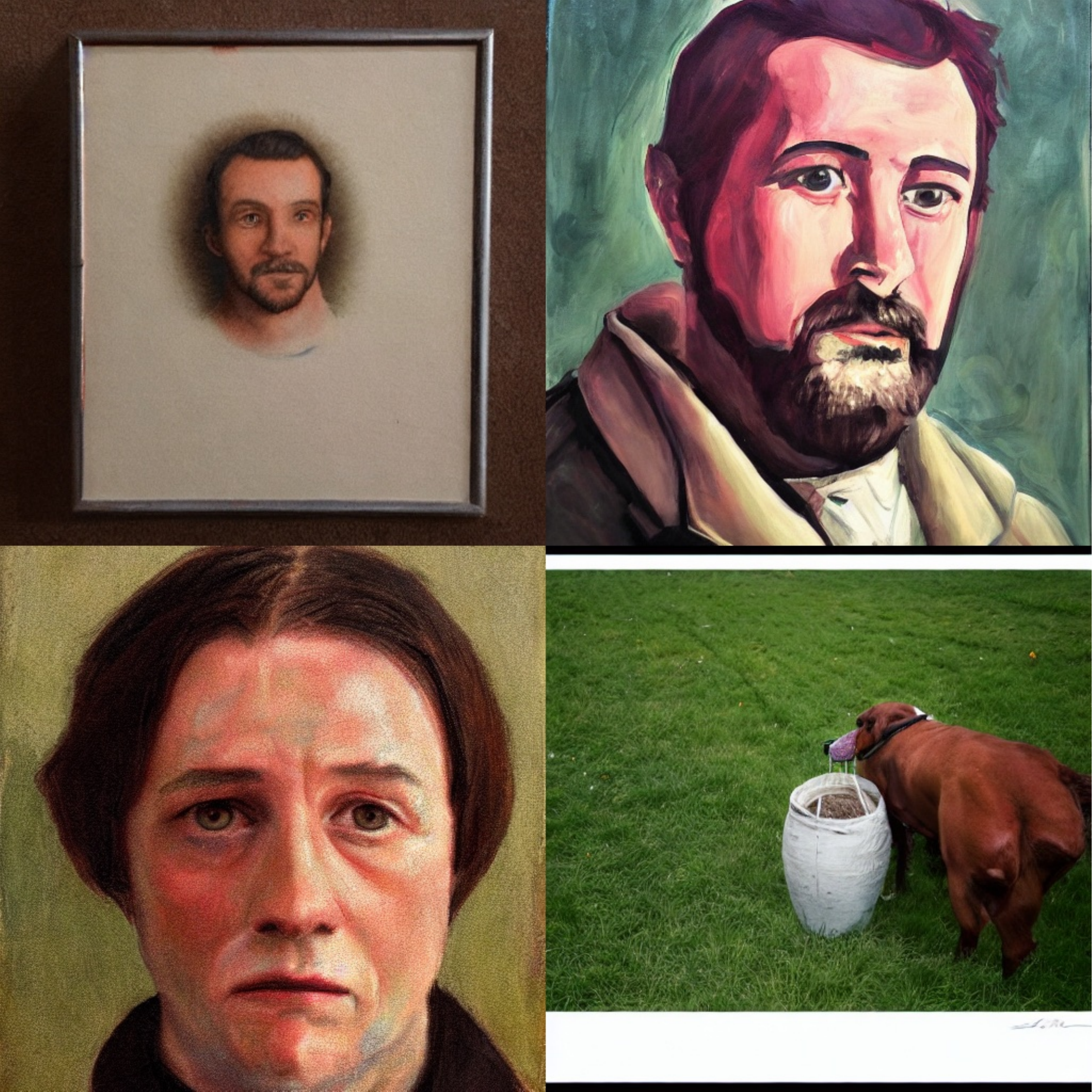}
\subcaption{ESD-x-1}
\end{minipage}
\begin{minipage}[b]{0.19\linewidth}
\centering
\includegraphics[width=0.8\linewidth]{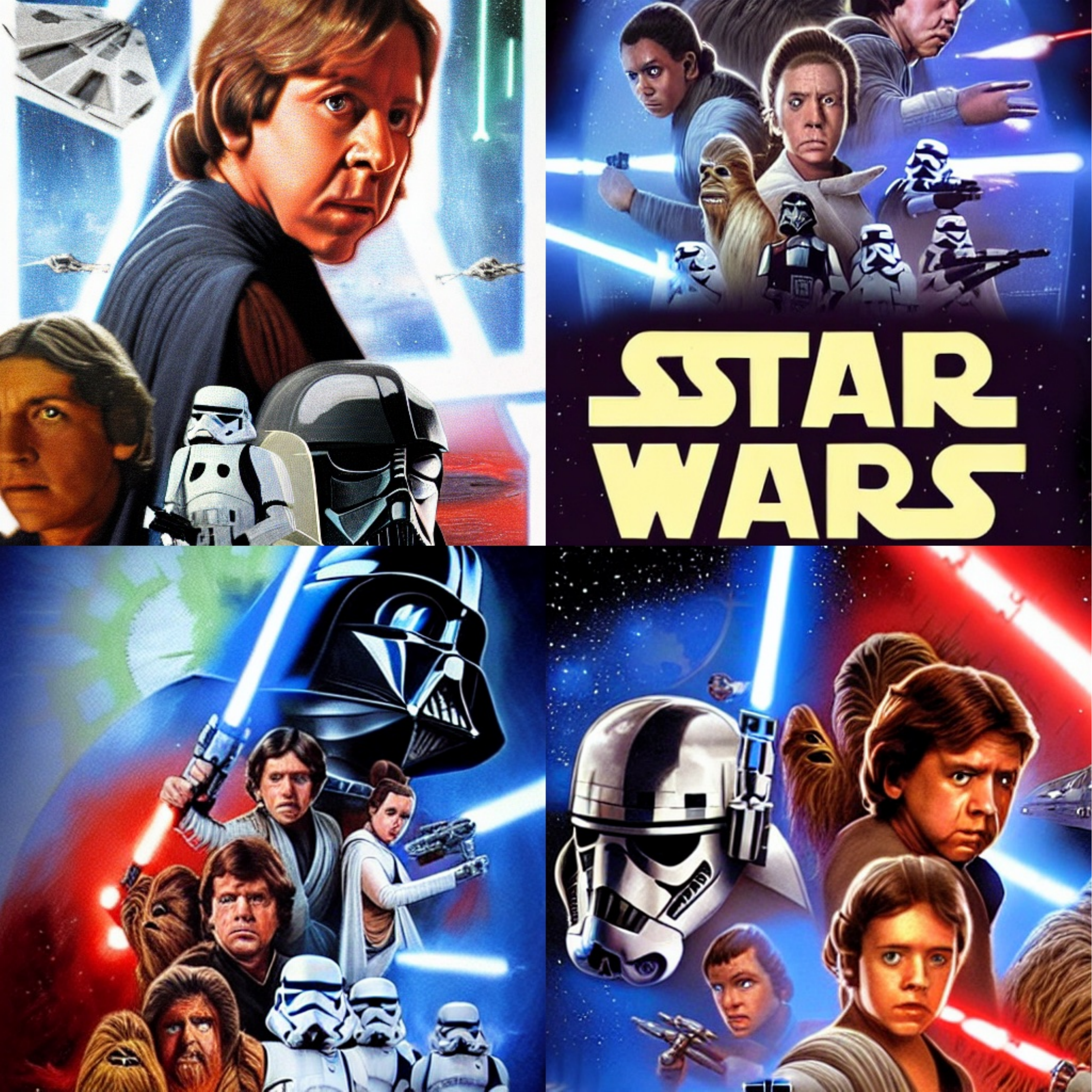}
\subcaption{UCE}
\end{minipage}
\begin{minipage}[b]{0.19\linewidth}
\centering
\includegraphics[width=0.8\linewidth]{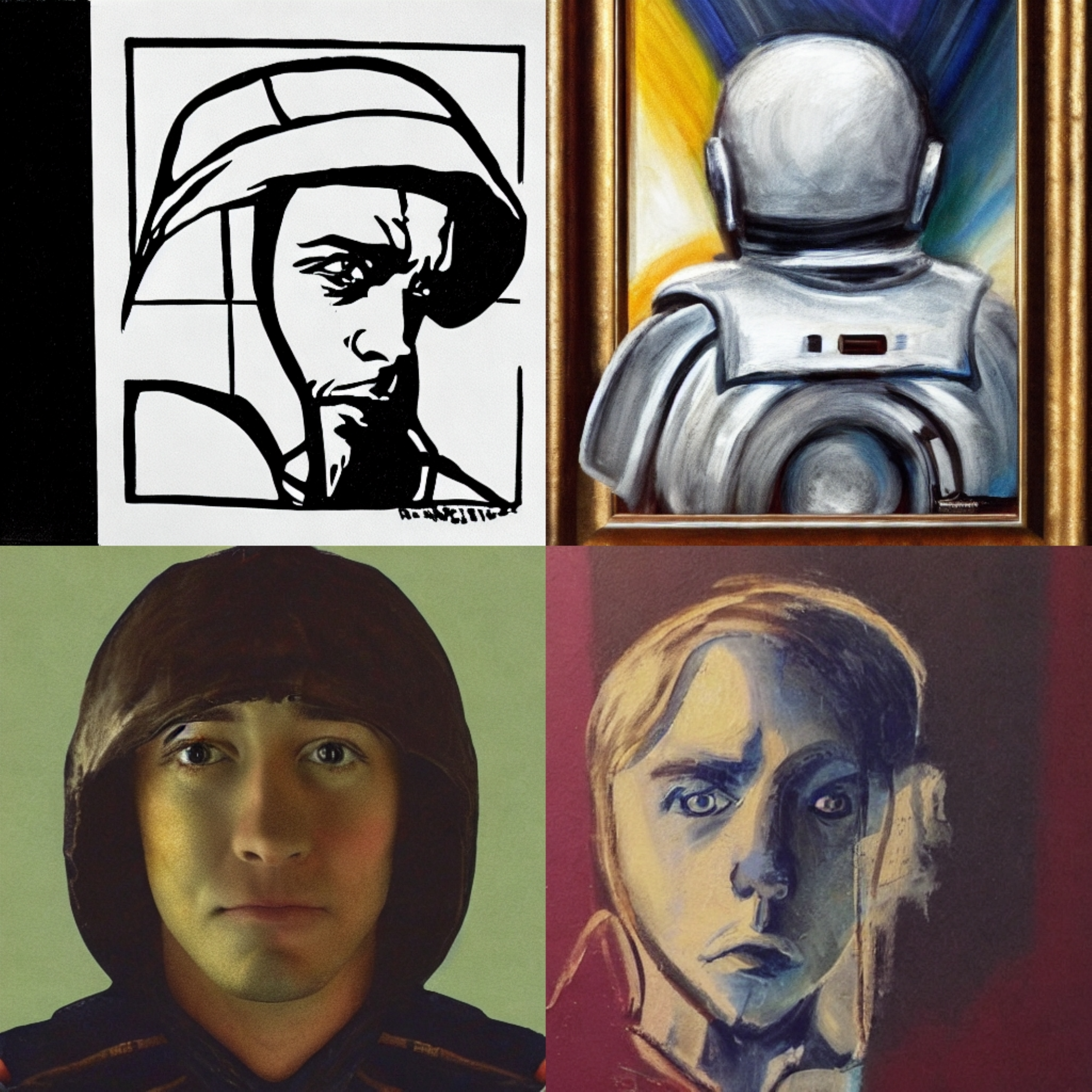}
\subcaption{SPM}
\end{minipage}
\begin{minipage}[b]{0.19\linewidth}
\centering
\includegraphics[width=0.8\linewidth]{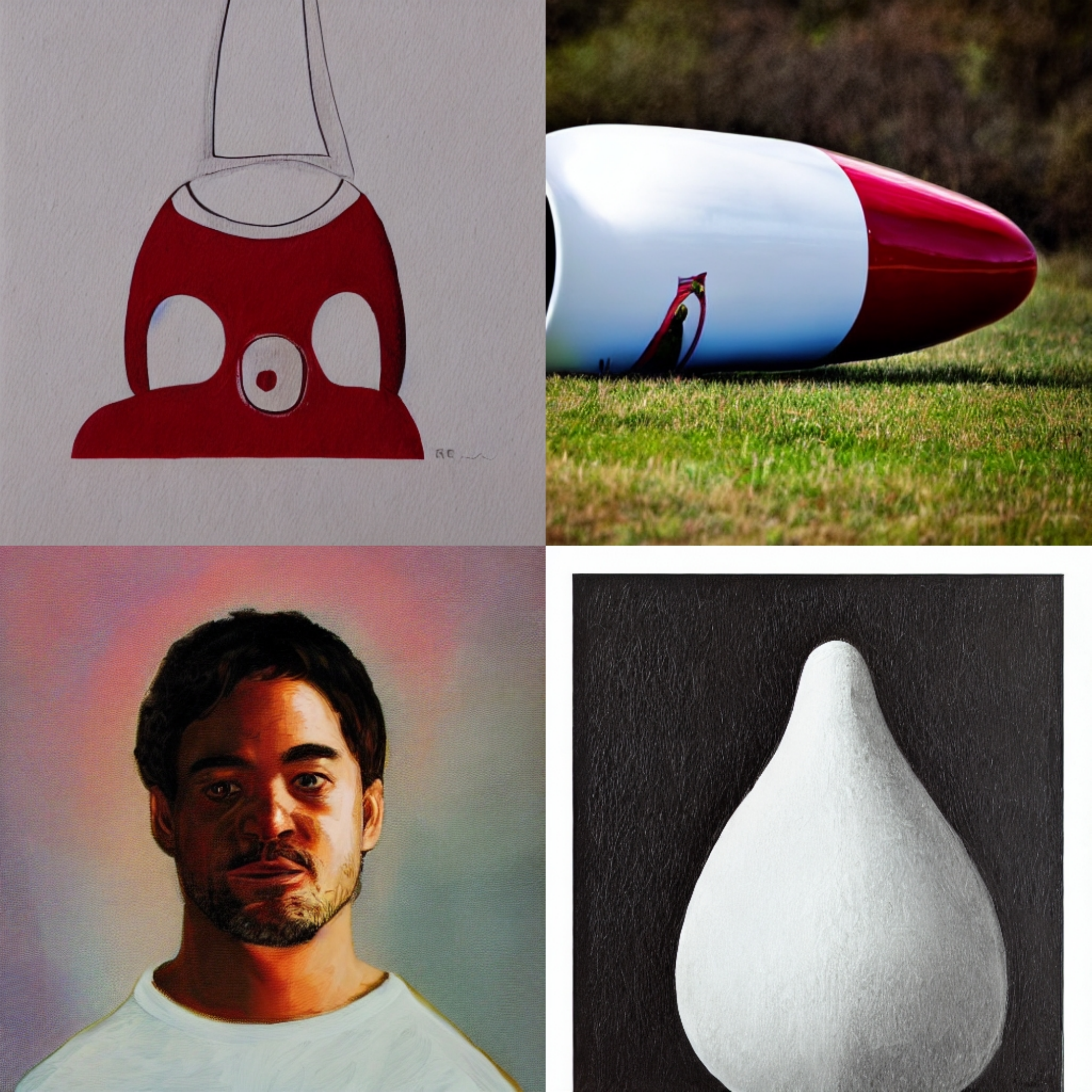}
\subcaption{Ours}
\end{minipage}   
\caption{Comparison of generated images when erasing ``R2D2''}
\label{fig:erasing-r2d2}
\end{figure}

\cref{fig:erasing-r2d2} illustrates the results of erasing ``R2D2''. While UCE produced images related to ``Star Wars'', it is worth noting that specifying proper nouns as anchor concepts becomes inappropriate when attempting to erase multiple concepts. As the number of concepts to erase increases, it becomes challenging for humans to designate anchor concepts while considering their relationships. This difficulty also applies when using large language models.

\begin{figure}[!htbp]
\begin{minipage}[b]{0.19\linewidth}
\centering
\includegraphics[width=0.8\linewidth]{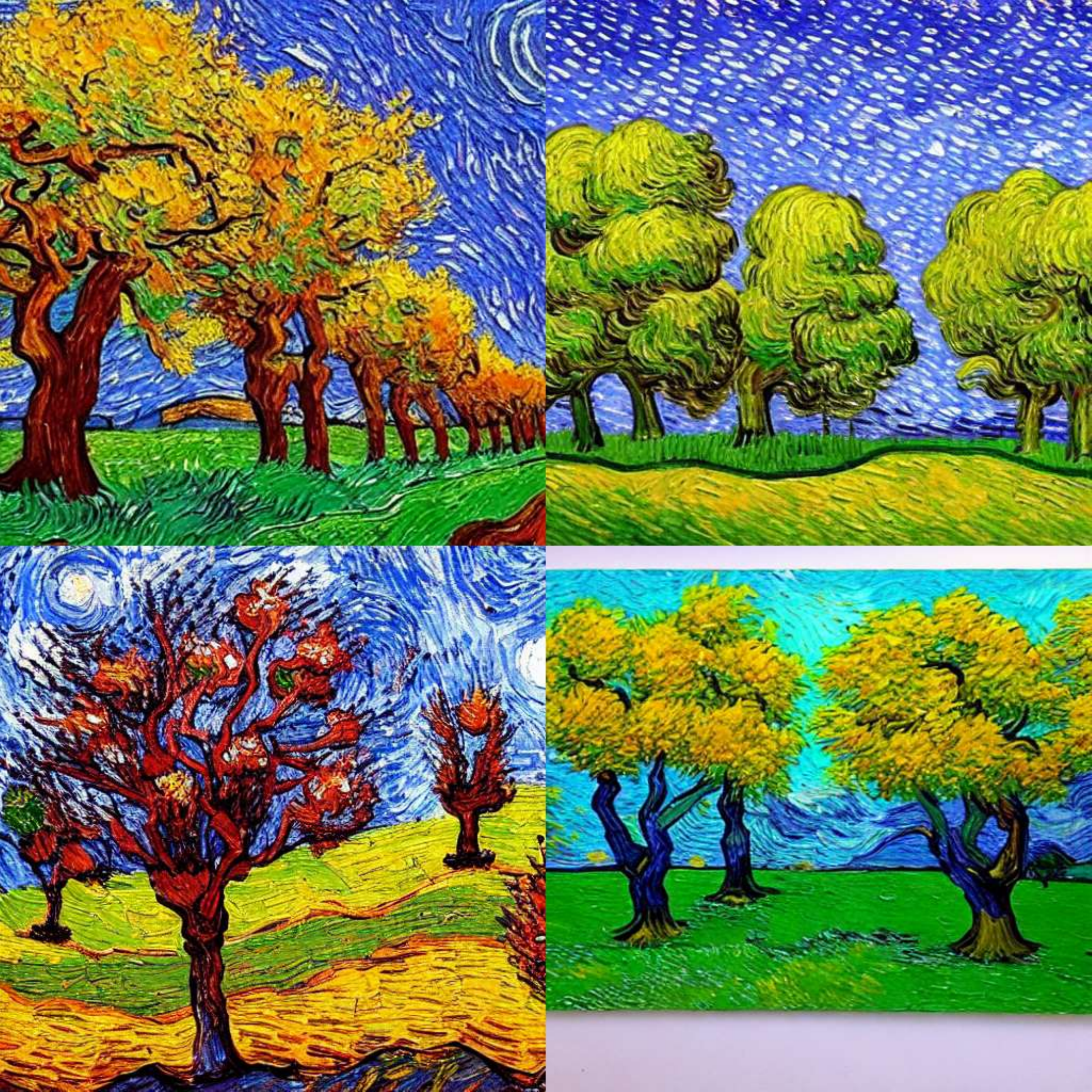}
\subcaption{Original SD}
\end{minipage}
\begin{minipage}[b]{0.19\linewidth}
\centering
\includegraphics[width=0.8\linewidth]{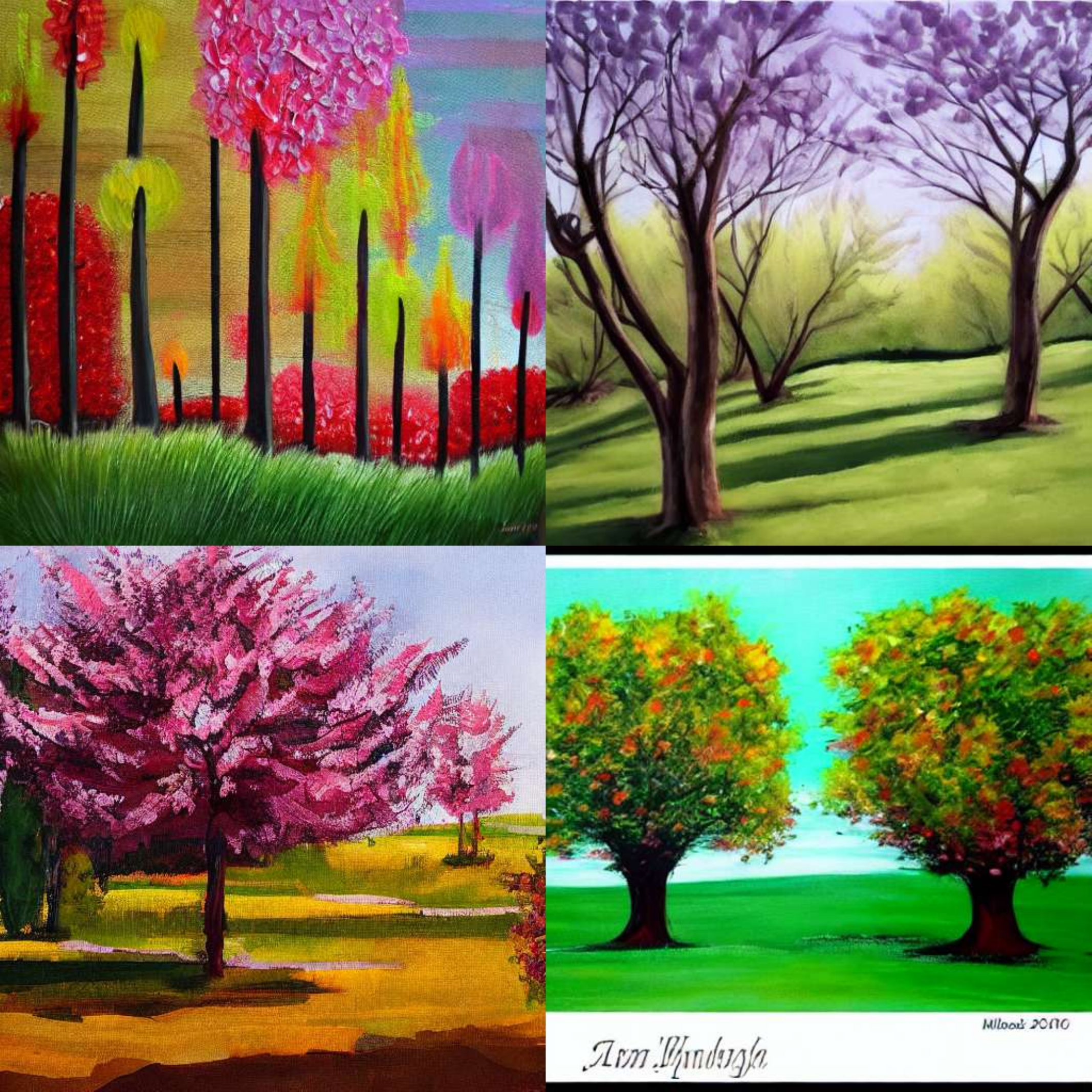}
\subcaption{ESD-x-1}
\end{minipage}
\begin{minipage}[b]{0.19\linewidth}
\centering
\includegraphics[width=0.8\linewidth]{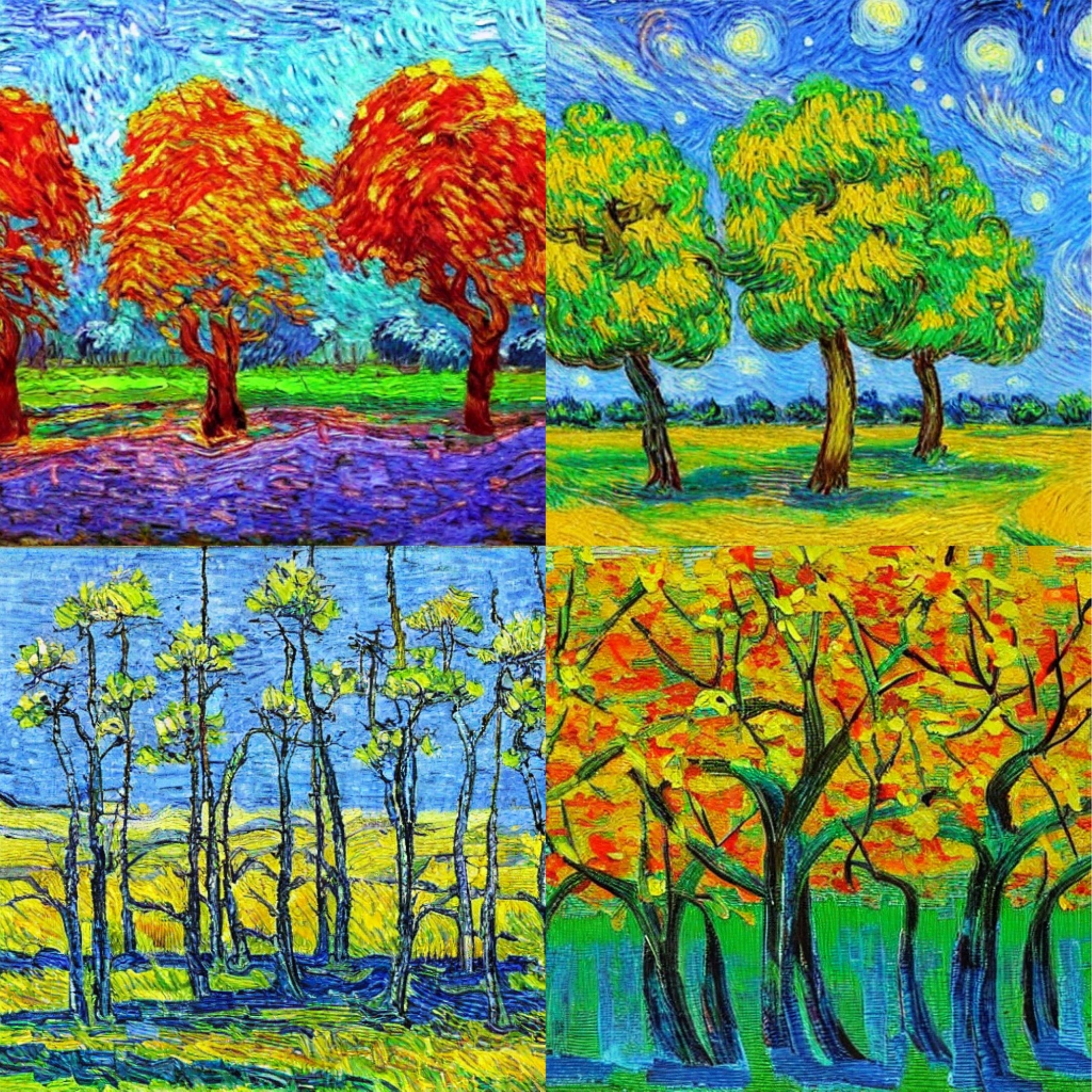}
\subcaption{UCE}
\end{minipage}
\begin{minipage}[b]{0.19\linewidth}
\centering
\includegraphics[width=0.8\linewidth]{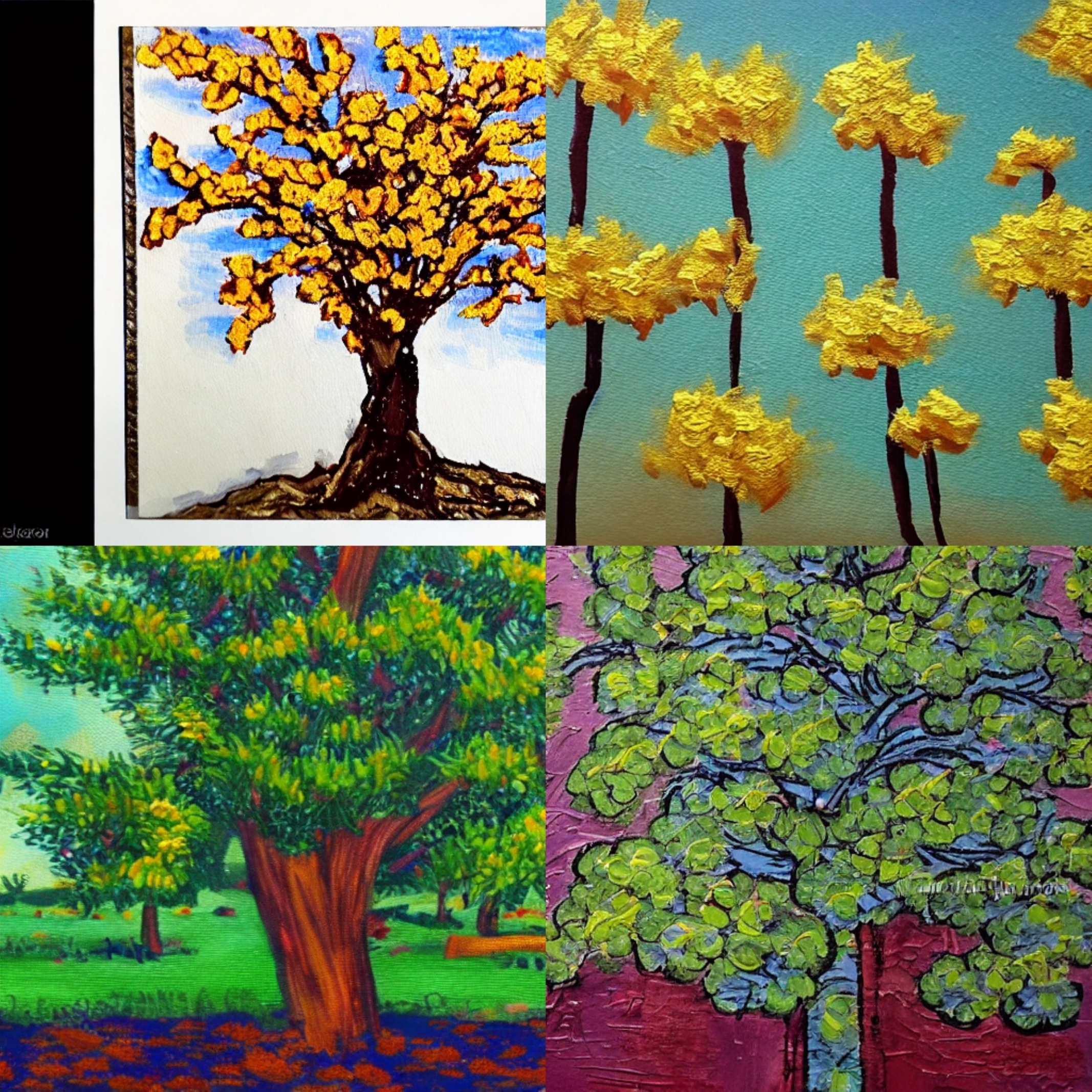}
\subcaption{SPM}
\end{minipage}
\begin{minipage}[b]{0.19\linewidth}
\centering
\includegraphics[width=0.8\linewidth]{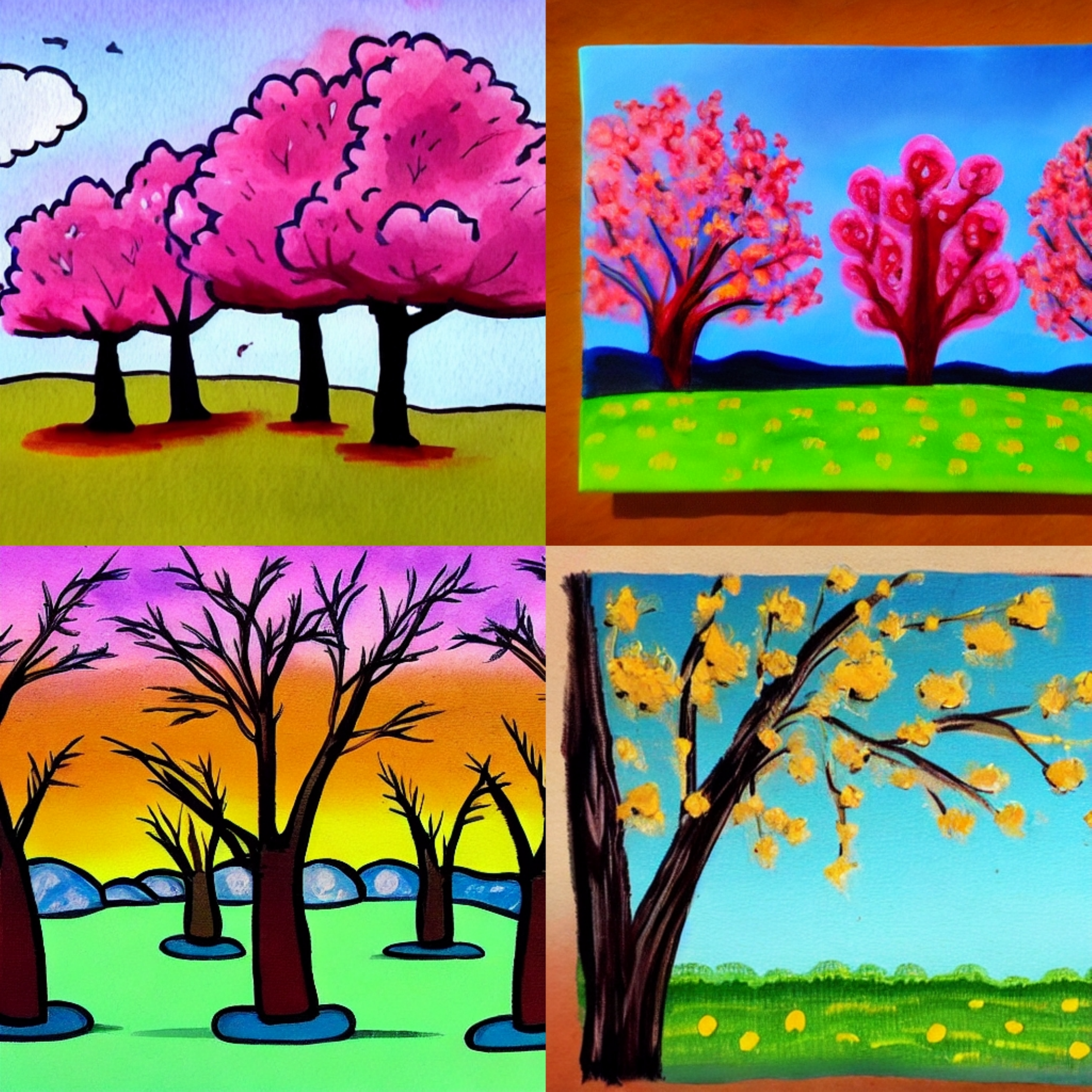}
\subcaption{Ours}
\end{minipage}   
\caption{Comparison of generated images when erasing ``Van Gogh style''}
\label{fig:erasing-gogh}
\end{figure}

\cref{fig:erasing-gogh} shows the result of erasing the ``Gogh style''. It is difficult to say that the Gogh style was completely erased because UCE output starry night, which is strongly associated with Gogh.

\subsection{Additional Results of Our Method}
\label{appendix:imagenette}
\begin{figure}[!htbp]
\centering
\begin{tabular}{@{}cc@{}}
\begin{tabular}{@{}c@{}}
Pre-trained \\
\end{tabular} &
\includegraphics[width=0.7\linewidth,valign=c]{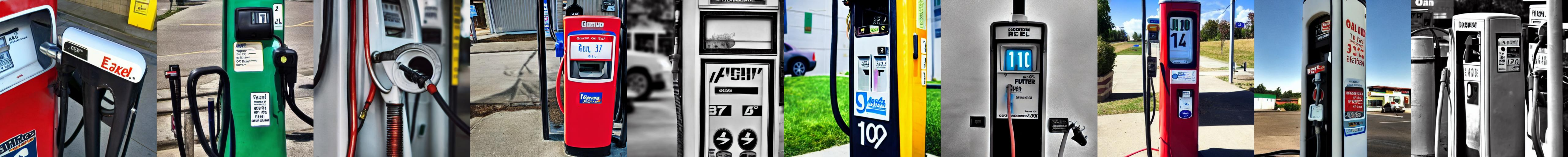} \\ \\
\begin{tabular}{@{}c@{}}
Erased \\
\end{tabular} &
\includegraphics[width=0.7\linewidth,valign=c]{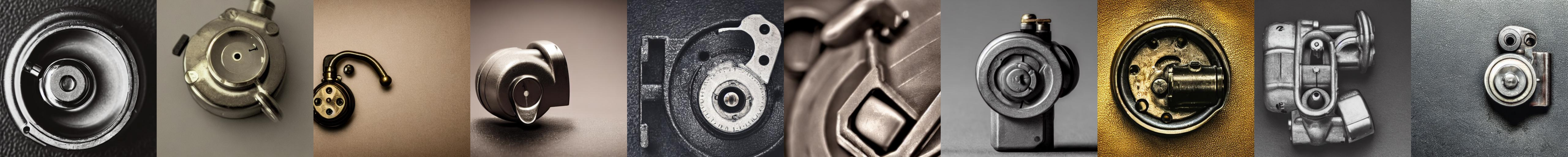}
\end{tabular}
\caption{Generated images of gas pump}
\label{fig:gas}
\end{figure}

\begin{figure}[!htbp]
\centering
\begin{tabular}{@{}cc@{}}
\begin{tabular}{@{}c@{}}
Pre-trained \\
\end{tabular} &
\includegraphics[width=0.7\linewidth,valign=c]{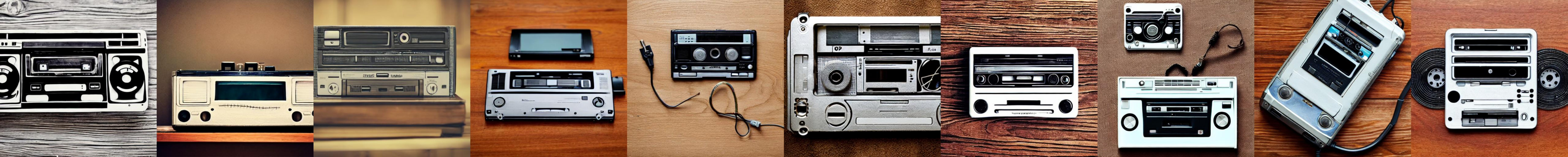} \\ \\
\begin{tabular}{@{}c@{}}
Erased \\
\end{tabular} &
\includegraphics[width=0.7\linewidth,valign=c]{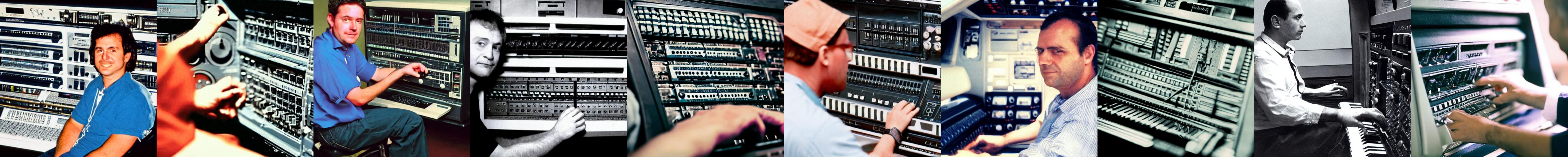}
\end{tabular}
\caption{Generated images of cassette player}
\label{fig:cassette}
\end{figure}

\begin{figure}[!htbp]
\centering
\begin{tabular}{@{}cc@{}}
\begin{tabular}{@{}c@{}}
Pre-trained \\
\end{tabular} &
\includegraphics[width=0.7\linewidth,valign=c]{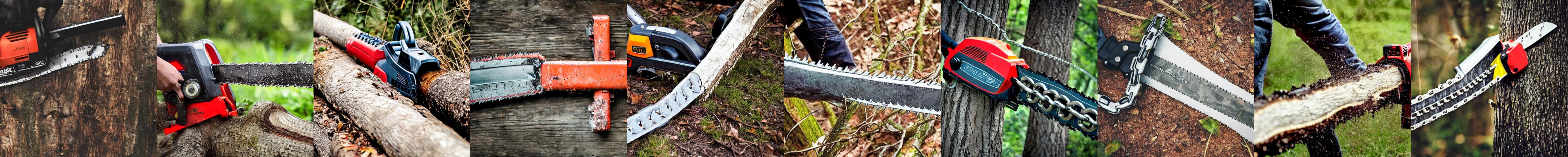} \\ \\
\begin{tabular}{@{}c@{}}
Erased \\
\end{tabular} &
\includegraphics[width=0.7\linewidth,valign=c]{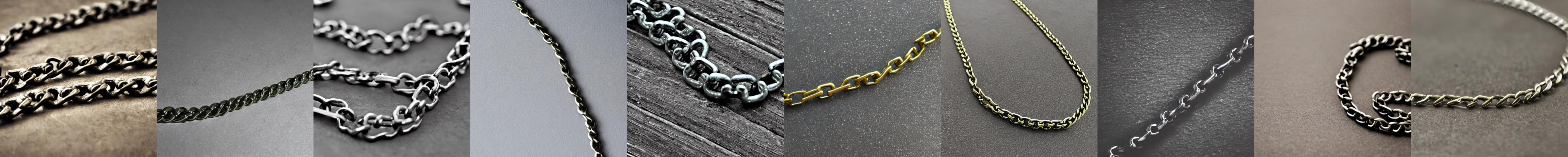}
\end{tabular}
\caption{Generated images of chainsaw}
\label{fig:chainsaw}
\end{figure}

\begin{figure}[!htbp]
\centering
\begin{tabular}{@{}cc@{}}
\begin{tabular}{@{}c@{}}
Pre-trained \\
\end{tabular} &
\includegraphics[width=0.7\linewidth,valign=c]{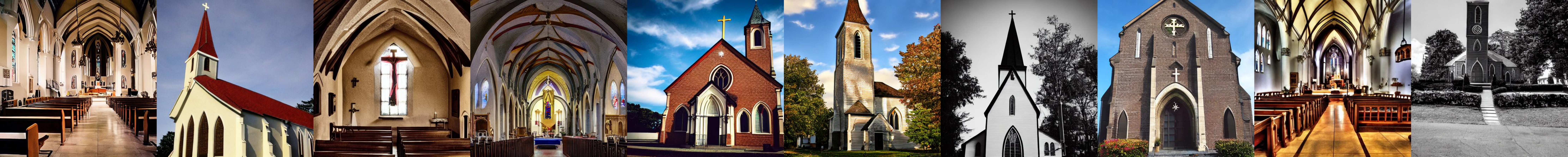} \\ \\
\begin{tabular}{@{}c@{}}
Erased \\
\end{tabular} &
\includegraphics[width=0.7\linewidth,valign=c]{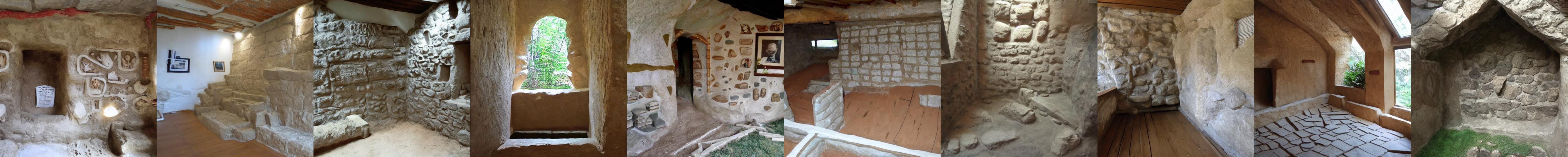}
\end{tabular}
\caption{Generated images of church}
\label{fig:church}
\end{figure}

\begin{figure}[!htbp]
\centering
\begin{tabular}{@{}cc@{}}
\begin{tabular}{@{}c@{}}
Pre-trained \\
\end{tabular} &
\includegraphics[width=0.7\linewidth,valign=c]{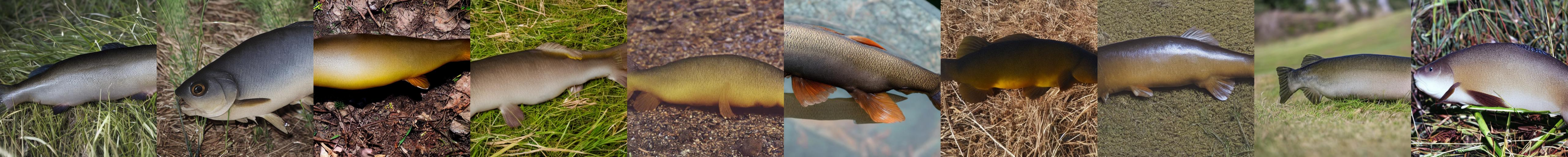} \\ \\
\begin{tabular}{@{}c@{}}
Erased \\
\end{tabular} &
\includegraphics[width=0.7\linewidth,valign=c]{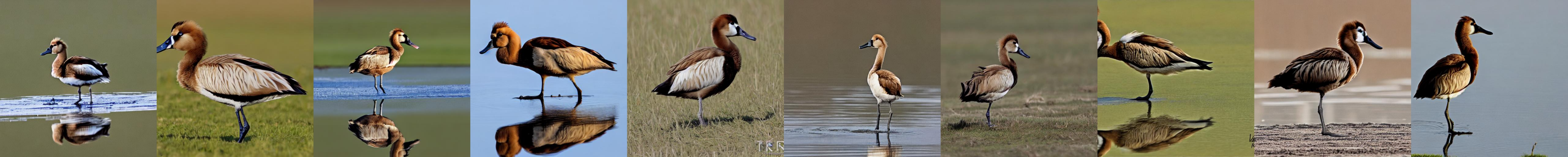}
\end{tabular}
\caption{Generated images of tench (a type of fish)}
\label{fig:tench}
\end{figure}

\begin{figure}[!htbp]
\centering
\begin{tabular}{@{}cc@{}}
\begin{tabular}{@{}c@{}}
Pre-trained \\
\end{tabular} &
\includegraphics[width=0.7\linewidth,valign=c]{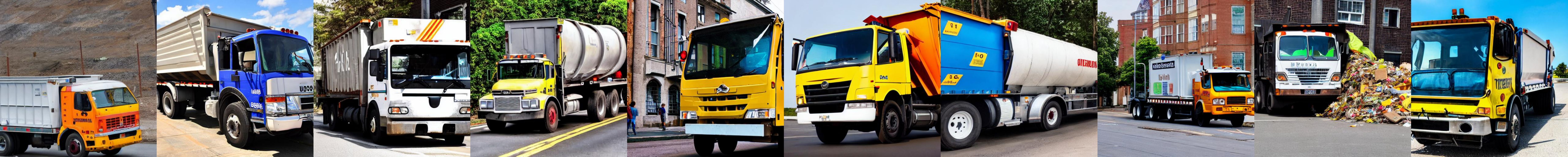} \\ \\
\begin{tabular}{@{}c@{}}
Erased \\
\end{tabular} &
\includegraphics[width=0.7\linewidth,valign=c]{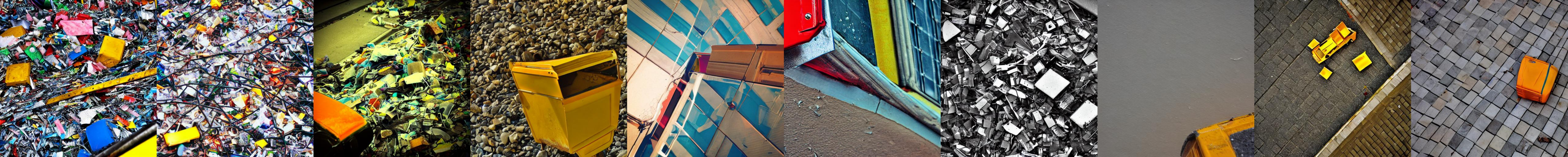}
\end{tabular}
\caption{Generated images of garbage truck}
\label{fig:garbage}
\end{figure}

\begin{figure}[!htbp]
\centering
\begin{tabular}{@{}cc@{}}
\begin{tabular}{@{}c@{}}
Pre-trained \\
\end{tabular} &
\includegraphics[width=0.7\linewidth,valign=c]{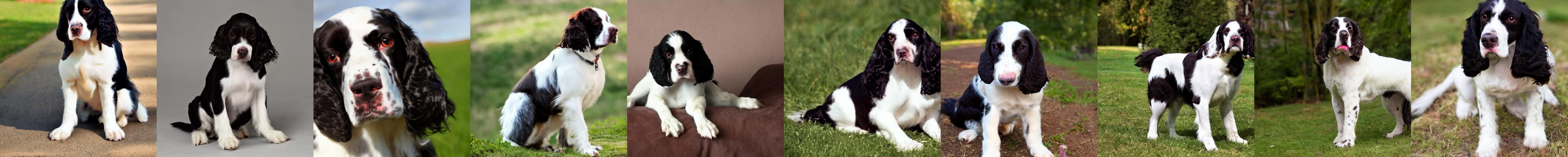} \\ \\
\begin{tabular}{@{}c@{}}
Erased \\
\end{tabular} &
\includegraphics[width=0.7\linewidth,valign=c]{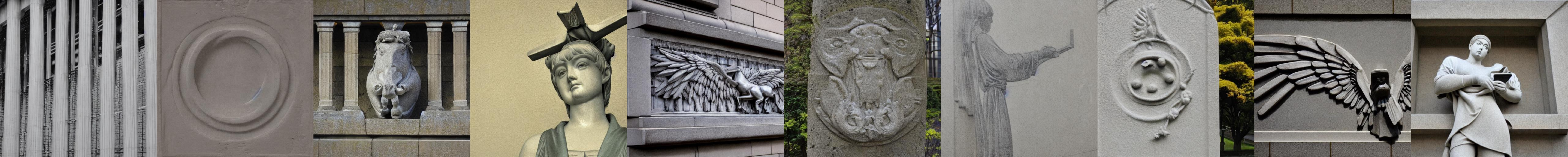}
\end{tabular}
\caption{Generated images of English springer}
\label{fig:english}
\end{figure}

\begin{figure}[!htbp]
\centering
\begin{tabular}{@{}cc@{}}
\begin{tabular}{@{}c@{}}
Pre-trained \\
\end{tabular} &
\includegraphics[width=0.7\linewidth,valign=c]{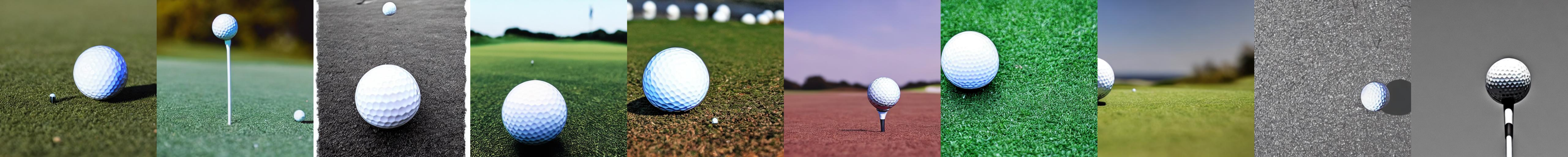} \\ \\
\begin{tabular}{@{}c@{}}
Erased \\
\end{tabular} &
\includegraphics[width=0.7\linewidth,valign=c]{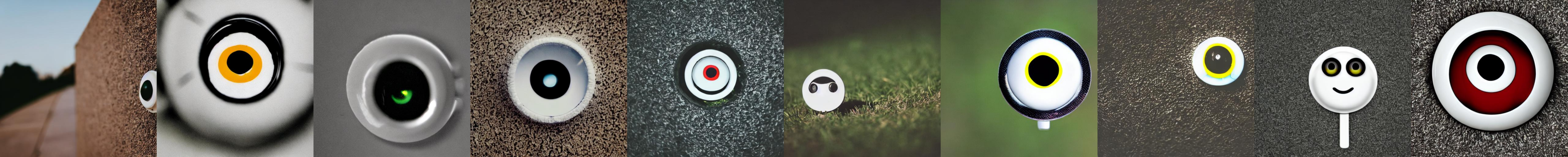}
\end{tabular}
\caption{Generated images of golf ball}
\label{fig:golf}
\end{figure}

\begin{figure}[!htbp]
\centering
\begin{tabular}{@{}cc@{}}
\begin{tabular}{@{}c@{}}
Pre-trained \\
\end{tabular} &
\includegraphics[width=0.7\linewidth,valign=c]{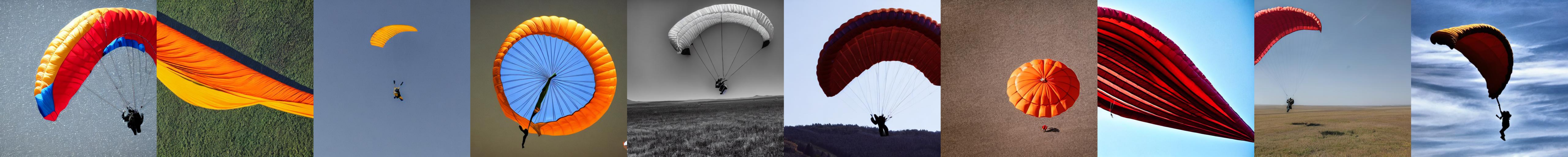} \\ \\
\begin{tabular}{@{}c@{}}
Erased \\
\end{tabular} &
\includegraphics[width=0.7\linewidth,valign=c]{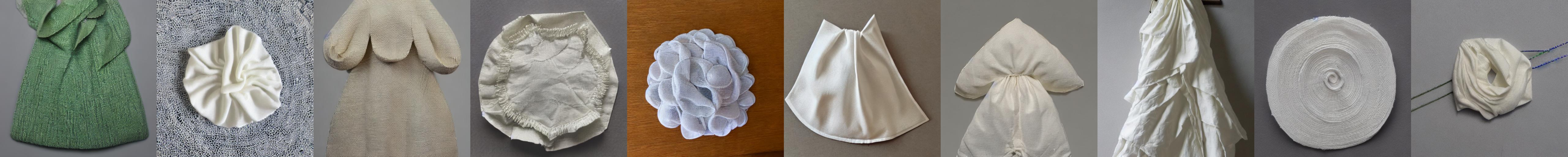}
\end{tabular}
\caption{Generated images of parachute}
\label{fig:parachute}
\end{figure}

\begin{figure}[!htbp]
\centering
\begin{tabular}{@{}cc@{}}
\begin{tabular}{@{}c@{}}
Pre-trained \\
\end{tabular} &
\includegraphics[width=0.7\linewidth,valign=c]{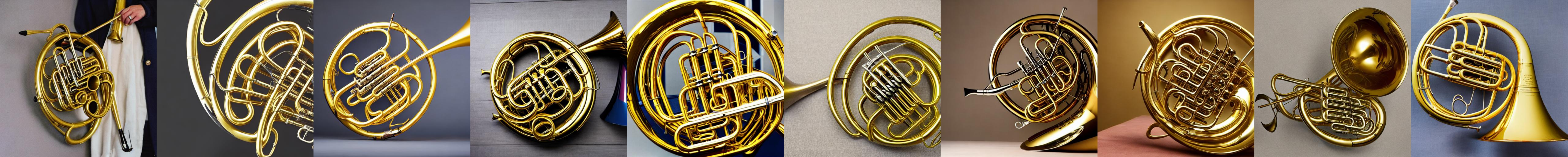} \\ \\
\begin{tabular}{@{}c@{}}
Erased \\
\end{tabular} &
\includegraphics[width=0.7\linewidth,valign=c]{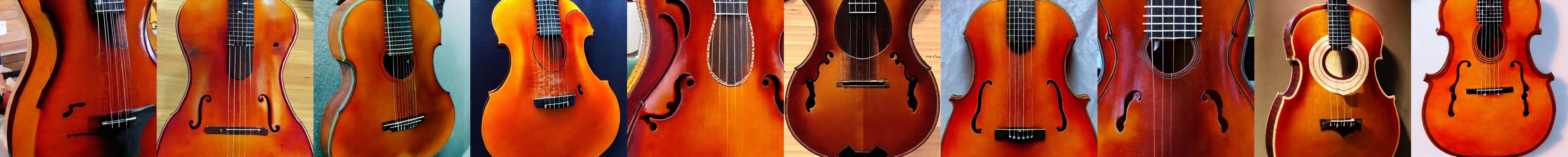}
\end{tabular}
\caption{Generated images of French horn}
\label{fig:horn}
\end{figure}

We present the results without comparing them with other methods. The purpose is solely to demonstrate the performance of our proposed method. We used all classes of Imagenette~\cite{info11020108}, which is a subset of ImageNet. We generated ten images for each class. The caption we used was ``a photo of a \{\textit{class name}\}''. In Figures~\ref{fig:gas}-\ref{fig:horn} show the comparisons between before and after erasing. 

Our proposed method demonstrates highly effective erasure performance even for objects containing common nouns such as those in Imagenette.

\subsection{Transferability}
\begin{figure}[!htbp]
\begin{minipage}[b]{0.49\linewidth}
\centering
\includegraphics[width=0.8\linewidth]{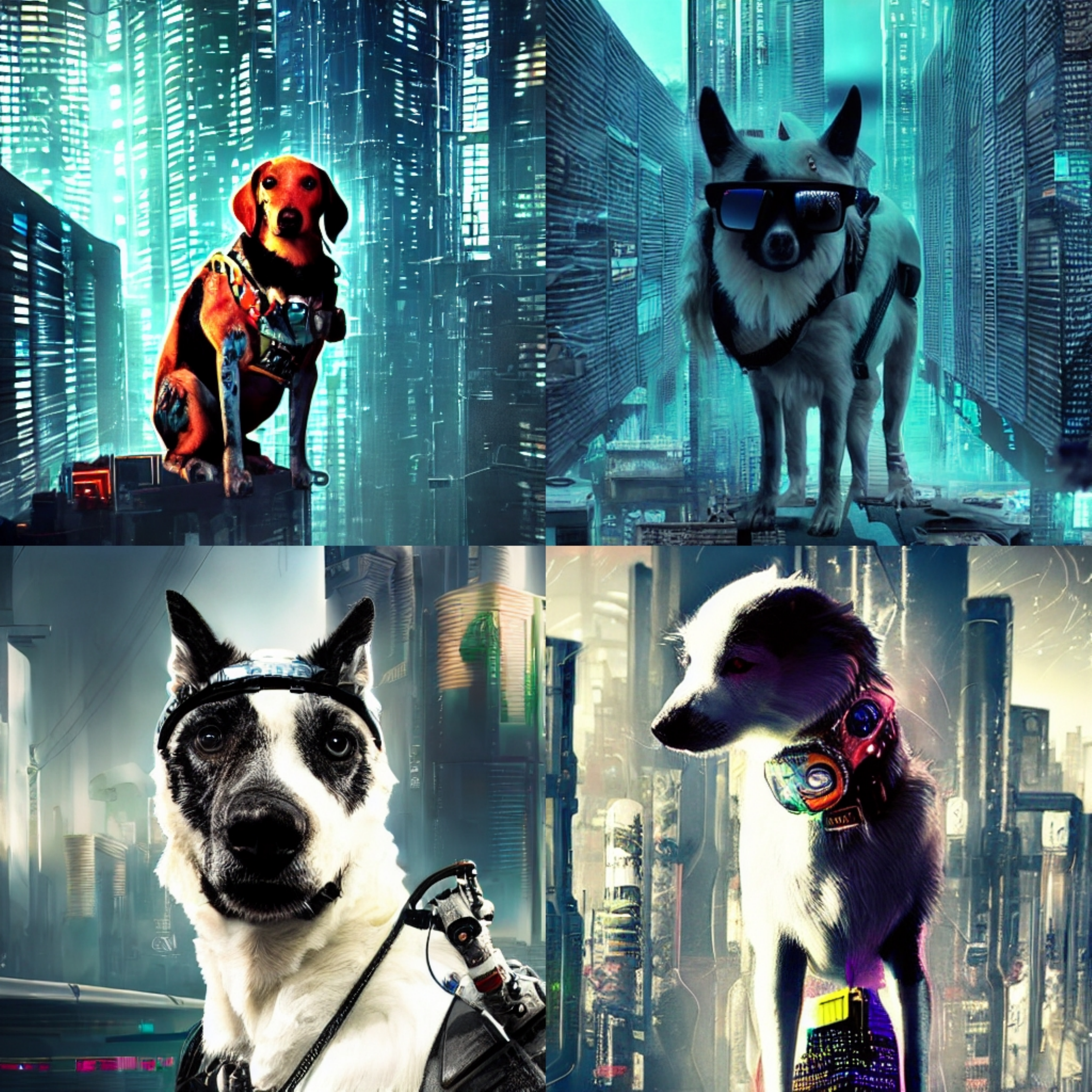}
\subcaption{Snoopy in cyberpunk style.}
\end{minipage}
\begin{minipage}[b]{0.49\linewidth}
\centering
\includegraphics[width=0.8\linewidth]{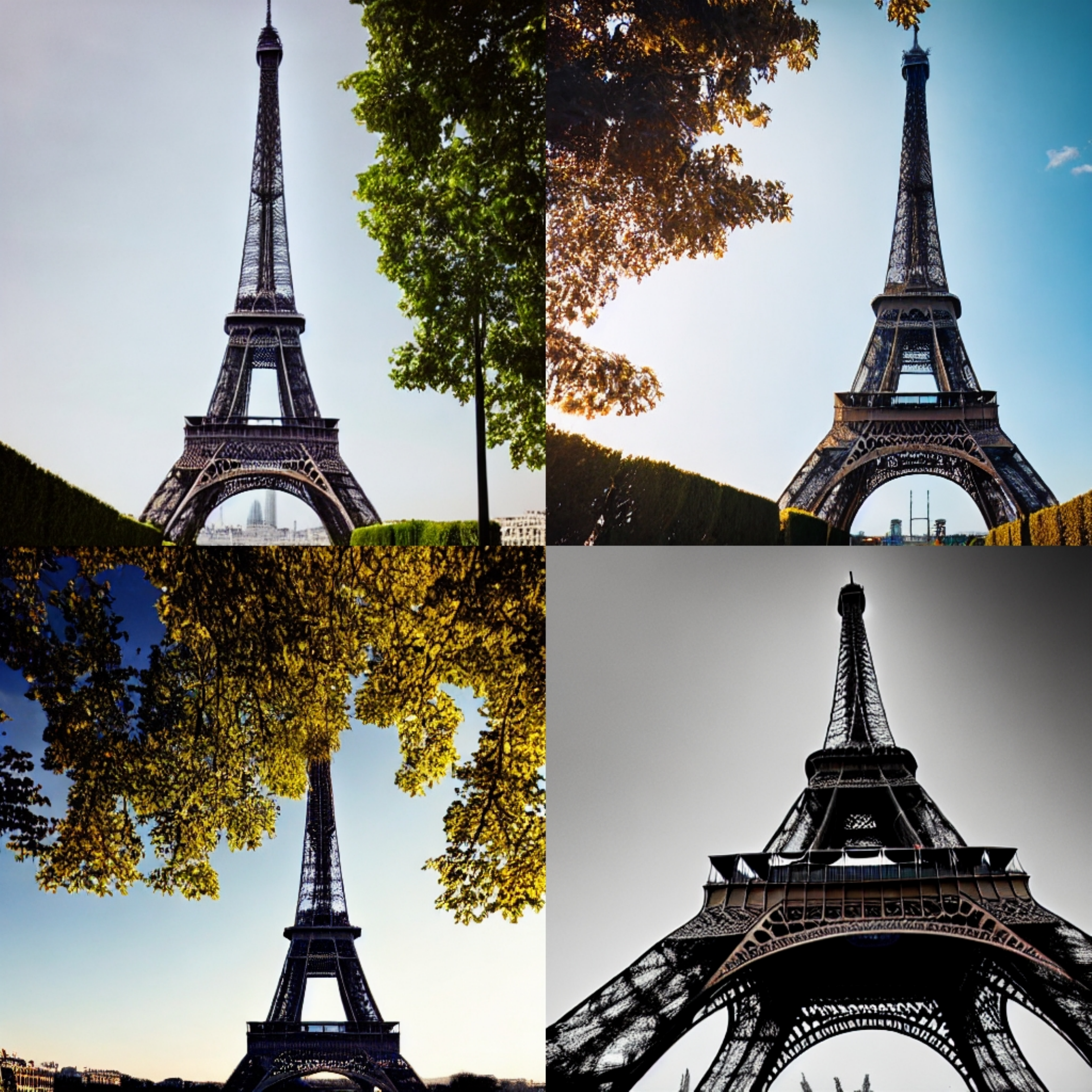}
\subcaption{a photo of Eiffel Tower.}
\end{minipage}   
\caption{Generated images using the text encoder from which ``Snoopy'' was erased with Stable Diffusion 1.4.}
\label{fig:sd14-snoopy-eiffel}
\end{figure}

Our proposed method updates the text encoder. Therefore, it can transfer the erased model to other text-to-image diffusion models that use the same text encoder. We confirm the transferability to other text-to-image diffusion models. We used Stable Diffusion 1.4\footnote{\url{https://huggingface.co/CompVis/stable-diffusion-v1-4}} with VAE and U-Net. The text encoder is the model from which ``Snoopy'' was erased. \cref{fig:sd14-snoopy-eiffel} shows the results. Once a concept is erased from the text encoder, we can be sure that the concept is also erased from the model using the same text encoder.

\begin{figure}[!htbp]
\begin{minipage}[b]{0.49\linewidth}
\centering
\includegraphics[width=0.8\linewidth]{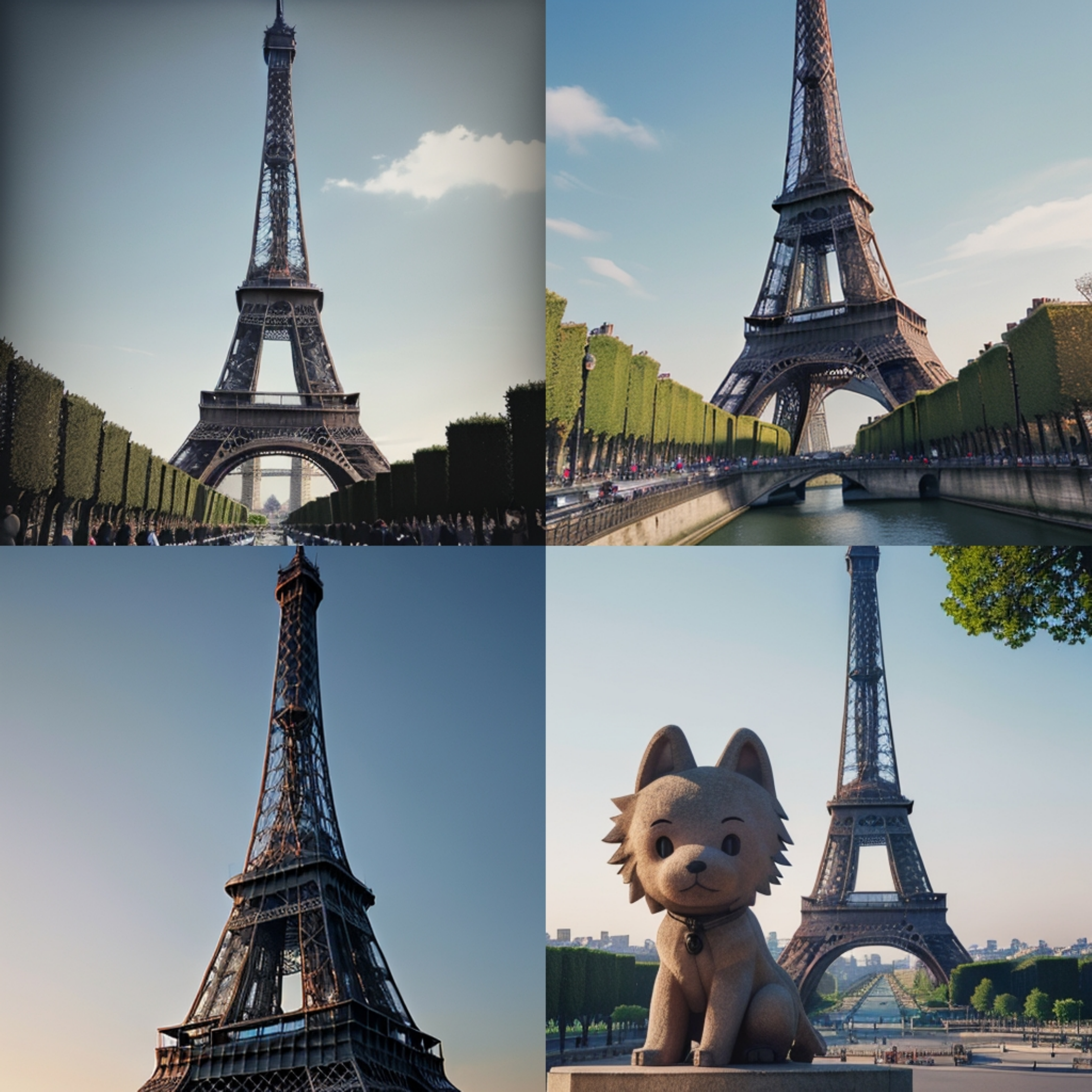}
\subcaption{Using text encoder from which ``Eiffel Tower'' was not erased}
\end{minipage}
\begin{minipage}[b]{0.49\linewidth}
\centering
\includegraphics[width=0.8\linewidth]{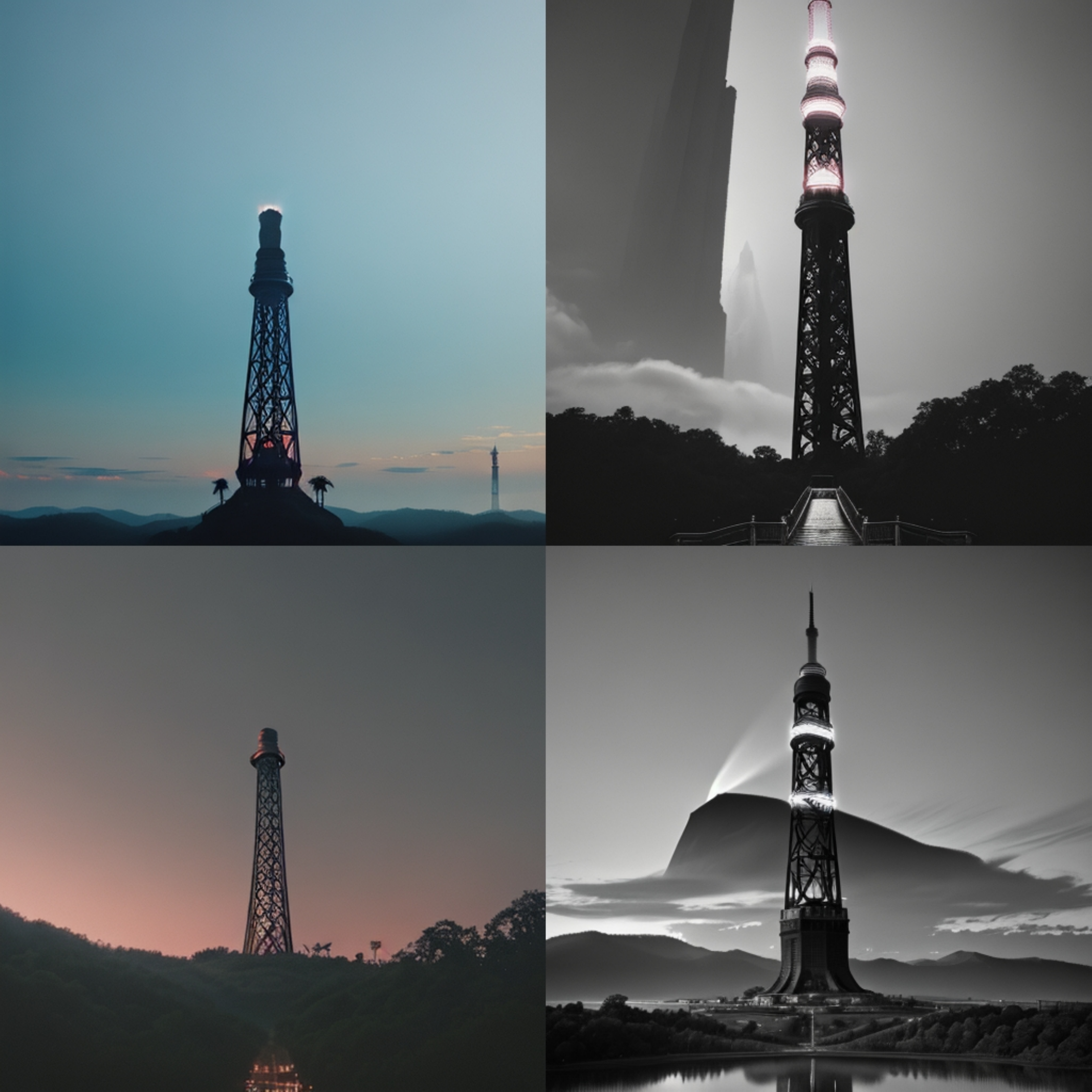}
\subcaption{Using text encoder from which ``Eiffel Tower'' was erased}
\end{minipage}   
\caption{Generated images before and after erasing ``Eiffel Tower'' with DreamShaper}
\label{fig:dream-eiffel}
\end{figure}

We used DreamShaper\footnote{\url{https://huggingface.co/Lykon/DreamShaper}} for community models. We used DEIS Scheduler~\cite{zhang2023fast} with 25 inference steps. \cref{fig:dream-eiffel} shows the results before and after erasing ``Eiffel Tower''. Because of the erasure of the Eiffel Tower from the text encoder, it was erased from the generated image.

\subsection{Additional Quantitative Results}
\label{subsec:additional-quantitative}
We provide the additional quantitative results in this subsection. First, we show the detection rate. We erased Imagenette classes and generated 100 images for each class. Then, we evaluated the top-1 accuracy using EVA02~\cite{FANG2024105171}, which is the highest top-1 accuracy on the ImageNet according to timm~\cite{rw2019timm}\footnote{\url{https://github.com/huggingface/pytorch-image-models/blob/main/results/results-imagenet.csv}}. We used ``a photo of a [\textit{class name}]'' as the prompt. The results are shown in \cref{tab:detection-rate-imagenette}. Our proposed method succeeds in erasing concepts in many cases even if we consider the misdetection of the classifier used in this experiment.

\begin{table}[htbp]
\caption{Detection rate (\%) of Imagenette. Lower score indicates that the target concept is erased correctly.}
\label{tab:detection-rate-imagenette}
\centering
\begin{tabular}{lcc}
\toprule
Class & Erased & Original SD \\
\midrule
church & 0 & 89 \\
parachute & 0 & 100 \\
golf ball & 15 & 100 \\
gas pump & 0 & 98 \\
garbage truck & 8 & 86 \\
tench & 0 & 95 \\
French horn & 0 & 100 \\
chain saw & 0 & 76 \\
English springer & 0 & 94 \\
cassette player & 3 & 12 \\
\midrule
Average & 2.6 & 85 \\
\bottomrule
\end{tabular}
\end{table}

Second, we confirmed the detection rate due to change the epochs. We use four classes: French horn, golf ball, garbage truck, and tench. \cref{tab:detection-rate-imagenette-epochs} shows the results. This results indicate the target concept is gradually erased. In addition, as well as described in \cref{subsec:num-epochs}, the difficulty to erase for each concept is different. The difficult concept to erase requires the large number of epochs to erase.

\begin{table}[htbp]
\caption{Detection rate (\%) of Imagenette. Lower score indicates that the target concept is erased correctly.}
\label{tab:detection-rate-imagenette-epochs}
\centering
\begin{tabular}{lccccc}
\toprule
Class & Epoch 1 & Epoch 2 & Epoch 3 & Epoch 4 & Epoch 5 \\
\midrule
French horn & 0 & 0 & 0 & 0 & 0 \\
golf ball & 100 & 92 & 45 & 26 & 15 \\
garbage truck & 86 & 85 & 72 & 46 & 8 \\
tench & 0 & 0 & 0 & 0 & 0 \\
\bottomrule
\end{tabular}
\end{table}

Third, we calculated the FID scores in addition to CLIP scores shown in \cref{tab:clip-score}. Following Lyu et al. \yrcite{Lyu_2024_CVPR}, we used CLIP ImageNet Template small. It has 27 prompts for objects and 19 prompts for styles. We generated 20 images each prompt. We used clean-fid~\cite{parmar2021cleanfid} for computing FID. It is natural that the baselines are lower FID because their ground truth for unrelated concept is the same images of the original SD. From the perspective of the difference of ground truth as well, it is not appropriate to use FID to compare our method and the baselines. It is not expected to improve the evaluation metrics after erasing because most erasure methods are not designed to improve the evaluation metrics. Therefore, it is expected that the methods whose ground truth is the same of the original SD get better score when comparing the existing metrics. The reason that our proposed method gets worse scores is aforementioned. However, as shown in \cref{tab:clip-score}, our proposed method is competitive in terms of CLIP Score, indicates text-image alignment. This metric does not rely on the generated images from original SD. Therefore, we consider the CLIP Score is better metric.

\begin{table}[htbp]
\caption{FID between each erasure method and original SD. Lower score indicates the images generated by erased model are the same that of original SD.}
\label{tab:fid-score}
\centering
\begin{tabular}{lccc|ccc}
\toprule
& \multicolumn{3}{c}{Erasing \textit{Eiffel Tower}} & \multicolumn{3}{c}{Erasing \textit{Monet Style}} \\
\midrule
& Tokyo Tower & Triumphal arch & cat & Gogh style & Picasso Style & Hokusai style \\
\midrule
ESD-x-1 & 173.313 & 18.199 & 27.612 & 231.889 & 149.214 & 121.994 \\
UCE & 114.863 & 5.663 & 25.892 & 48.512 & 37.102 & 56.135 \\
SPM  & 47.780 & 6.236 & 34.701 & 57.024 & 62.081 & 57.530 \\
Ours & 81.158 & 28.801 & 26.738 & 252.364 & 90.956 & 206.979 \\
\bottomrule
\end{tabular}
\end{table}

Fourth, we evaluated the generative ability of the unrelated concepts using MSCOCO-30k FID and CLIP Score. We evaluated only our proposed method. The concepts to be erased are Eiffel Tower and Monet Style. The results are shown in \cref{tab:mscoco-scores}. These results indicate that our proposed method has minimal impact on the generative ability when erasing a single concept. Moreover, when erasing Eiffel Tower, FID and CLIP Score get better. The fact that performance improvements are observed despite not making any enhancements to the model suggests that thiese evaluation metrics are inherently inappropriate.

\begin{table}[htbp]
\caption{FID and CLIP score of our proposed method on MSCOCO-30k}
\label{tab:mscoco-scores}
\centering
\begin{tabular}{lcc}
\toprule
& FID $\downarrow$ & CLIP Score $\uparrow$ \\
\midrule
Original SD & 13.8996 & 0.2667 \\
Erasing \textit{Eiffel Tower} & 13.4874 & 0.2672 \\
Erasing \textit{Monet Style} & 13.9456 & 0.2669 \\
\bottomrule
\end{tabular}
\end{table}

Fifth, we calculated detection rate when erasing multiple concepts. We erased tench, English springer, chain saw in that order. The results are shown in \cref{tab:detection-rate-multi}. We used the model at the end of third epoch. Similar to the results in \cref{fig:multi}, the more concepts that are erased, the more difficult it becomes to maintain other concepts. Notably, using the model at the end of the fourth epoch resulted in a significant degradation of other concepts. This indicates that our method is sensitive to the number of epochs. However, it is easy to erase one concept while maintaining the others according to \cref{tab:mscoco-scores}. Therefore, task vectors~\cite{ilharco2023editing} or other methods may be used to overcome this challenge.

\begin{table}[htbp]
\caption{Detection rate (\%)}
\label{tab:detection-rate-multi}
\centering
\begin{tabular}{l|cccc}
\toprule
\multirow{2}{*}{Erased Class} & \multicolumn{4}{c}{Rate} \\
& tench & English springer & chainsaw & church (unerased concept) \\
\midrule
tench & 0 & 95 & 100 & 94 \\
$+$ English springer & 0 & 0 & 94 & 93 \\
$+$ English springer $+$ chainsaw & 0 & 0 & 2 & 85 \\
\bottomrule
\end{tabular}
\end{table}

\end{document}